\def\PREPRINT{}
\documentclass[10pt]{article}
\ifdefined\PREPRINT\usepackage[preprint]{tmlr}\else\usepackage{tmlr}\fi
\makeatletter\newif\ifanon\if@accepted\anonfalse\else\anontrue\fi\makeatother

\usepackage{amsmath,amssymb}
\usepackage{booktabs}
\usepackage{tabularx}
\usepackage{array}
\usepackage{enumitem}
\usepackage{microtype}
\usepackage{graphicx}
\usepackage{placeins}
\usepackage{xcolor}
\usepackage{tikz}
\usetikzlibrary{positioning,arrows.meta,fit}
\usepackage{xurl}
\usepackage[hidelinks]{hyperref}

\setlist{itemsep=2pt,topsep=4pt}
\newcolumntype{L}{>{\raggedright\arraybackslash}X}
\newcolumntype{P}[1]{>{\raggedright\arraybackslash}p{#1}}

\newcommand{\anonelse}[2]{\ifanon #1\else #2\fi}

\newcommand{\prbias}{\anonelse{a vLLM PR}{vLLM PR \#63081}}
\newcommand{\citepbias}{\ifanon\citep{vllm63081anon}\else\citep{vllm63081}\fi}

\title{Fast 4-bit KV-Cache Decoding on Hopper GPUs\\
Without FP4 Hardware}
\author{\name Jian Zhang\\
      \addr Northeastern University}
\def\month{MM}
\def\year{YYYY}
\def\openreview{\url{https://openreview.net/forum?id=XXXX}}

\begin{document}

\maketitle
{\renewcommand{\thefootnote}{}\footnotetext{\textbf{AI disclosure statement.} The research questions, design decisions and all claims in this paper are the \ifanon authors'\else author's\fi{}. Generative AI tools (Claude, Claude Code) were used for writing code, running and orchestrating experiments and drafting text. Every reported result comes from logged runs that the \ifanon authors\else author\fi{} checked, and the \ifanon authors take\else author takes\fi{} full responsibility for the content.}}

\begin{abstract}
On Hopper GPUs, which have no FP4 hardware, a 4-bit KV cache is slower than FP8 in today's serving stacks: on H200 the FlashInfer kernel behind vLLM's NVFP4 KV cache reaches only 10--19\% of memory bandwidth. We show that this is a property of current implementations, not of the format. NVBit instruction counts against Hopper's issue budget, about 7 thread instructions per byte on H200, give two causes: the kernels behind vLLM's 4-bit paths execute 2--6$\times$ that budget, and the kernels we counted serialize loading, decoding and matrix multiplies. A decode kernel that stays within the budget and overlaps decoding with loading (a TMA producer warp, a cache layout that feeds decoded values to the tensor cores from registers, swapped operands for small query groups) is, as a bare kernel, 1.10--1.48$\times$ faster than the faster of TensorRT-LLM's XQA and FlashAttention-4's native FP8 kernels at all 11 shapes we measured, and 1.22--1.76$\times$ when the 4-bit K and V run on FP8 tensor cores. In vLLM, with K mean removal, K rotation, four BF16 sinks and layer 0 in BF16, a decode step on Llama-3.1-8B is 1.12--1.64$\times$ faster than with FlashInfer FP8 at batch 8 and above (on par at batch 1; 1.04--1.18$\times$ on 32B and 72B models), the cache holds 1.63$\times$ as many tokens, and on five models the 95\% intervals exclude accuracy losses larger than about 3.5 points against BF16 KV. At 120K tokens on Llama-3.1-8B our cache loses 15 points against BF16 KV, about as much as vLLM's default FP8 path, while offline FP8 storage alone loses 3 and our cache 10. One H200 absorbed 1.23--2.20$\times$ FP8's request rate in 600-s bursts from an empty cache (mostly one seed); a higher sustained rate we examined but did not establish. In Qwen2-family models that collapse under FP8 KV, keeping layer 0 (whose key mean comes from the key-projection bias) in BF16 restores them, one only partly.
\end{abstract}

\section{Introduction}
\label{sec:intro}

In long-context and batched serving, every decoding step reads the whole KV cache, so decode attention is limited by memory bandwidth. For Llama-3.1-8B (32 query heads, 8 KV heads, head dimension 128), each KV row of 256 bytes in BF16 feeds about 1,024 floating-point operations, roughly 4 per byte, while an H200 can do about 206 dense BF16 tensor-core operations per byte (989 TFLOPS $\div$ 4.8 TB/s, spec-sheet values). Time per step is therefore close to bytes read divided by bandwidth, and fewer bytes means faster decoding and more requests per GPU.

NVIDIA's Blackwell GPUs decode 4-bit formats such as NVFP4 in hardware. The installed base is still largely Hopper (H100, H200, H20), which supports BF16 and FP8 but not FP4. There, 4-bit values must be unpacked and scaled in software inside the attention kernel. We measured what that costs today. At batch 32 and 16K context on H200, one call of the NVFP4 decode kernel of FlashInfer~\citep{ye2025flashinfer}, which computes one attention layer for one decode step of all 32 sequences, takes 0.669 ms, more than its BF16 kernel in the same job (0.495 ms), although it reads less than a third of the bytes: the BF16 kernel runs at 90\% of peak bandwidth, the NVFP4 kernel at 19\%. In another job, its FP8 kernel XQA, TensorRT-LLM's decode kernel for grouped-query attention, which vLLM uses, takes 0.255 ms, at 88\% (Table~\ref{tab:xqa}). Across the 11 shapes of Figure~\ref{fig:kernel}, FlashInfer's NVFP4 kernel reaches 10--19\% of peak bandwidth. A vLLM~\citep{kwon2023vllm} user who enables an NVFP4 KV cache on Hopper saves memory but decodes more slowly.

The usual reading is that on such GPUs a 4-bit cache buys capacity, not speed. PyTorch's INT4 decode kernels for grouped-query attention reach 17--30\% of bandwidth on H100~\citep{pytorch2024int4decoding}; the UltraQuant paper finds that 4-bit KV ``buys capacity, not kernel speed'' on AMD MI355X~\citep{chakrabarti2026ultraquant}; and QServe~\citep{lin2025qserve} argued for INT4 on A100 that software decoding makes low-bit attention compute-bound. H200 makes the problem harder: its memory bandwidth grew faster than its instruction issue rate, so a kernel at full bandwidth may spend about 7 instructions per byte, against about 10 on H100 SXM and A100 (Section~\ref{sec:stall}). FOSCAR~\citep{zhuang2026foscar} reports INT4 decoding faster than TensorRT-LLM's FP8 on H100. Our measurements show the same for the FP4 formats that vLLM ships, against the fastest FP8 kernels we found, and explain it: the slowness of 4-bit KV caches on Hopper is a property of current implementations, not of the format. A kernel that keeps decoding within the issue budget and overlaps it with loading beats XQA and FlashAttention-4's native FP8 at every shape we measured, and the gain carries into vLLM at batch 8 and above with no accuracy loss larger than about 3.5 points on the models we ran repeatedly. The paper gives the evidence in that order: why current kernels are slow, how a kernel avoids it, what it gains inside vLLM, and what it costs in accuracy.

The evidence comes in four parts:

\begin{enumerate}
  \item \textbf{Why current kernels are slow.} Using NVBit~\citep{villa2019nvbit} instruction counts against Hopper's issue budget, we show that the 4-bit decode kernels we counted stall for two separate reasons, too many instructions and serialized waiting: the Triton and FlashInfer kernels execute 2--6$\times$ the budget yet use only 40--52\% of the issue rate, so they also wait, while BitDecoding stays within the budget (without its spin-waits) and only waits. XQA's FP8 kernel, for reference, is within the budget and near the bandwidth limit. The same FlashInfer NVFP4 kernel that is 1.3--1.7$\times$ slower than FlashInfer's FlashAttention-2-style FP8 kernel on H200 (not XQA, its faster FP8 kernel there) is 1.06--1.15$\times$ faster on an RTX A5000 and 1.56--1.69$\times$ faster on an L40S, GPUs with 2.6$\times$ and 7.6$\times$ H200's instructions per byte, consistent with the budget argument (Section~\ref{sec:stall}).
  \item \textbf{A kernel that avoids both.} Our Hopper-native decode kernel for NVFP4, MXFP4 and INT4 KV caches (paged for the FP4 formats) combines a producer warp that loads with TMA and overlaps loading with two compute warpgroups, a cache layout chosen when the cache is written so that decoded values feed wgmma (Hopper's warpgroup matrix multiply) from registers, and swapped operands so that small query groups are not padded to wgmma's 64 rows. None of these is new on its own (Sections~\ref{sec:background} and~\ref{sec:design}); together they keep 4-bit decoding within the issue budget. In the paged form our vLLM plugin uses, the bare kernel is 1.10--1.48$\times$ faster than the fastest FP8 kernel at each shape (XQA or FlashAttention-4's native FP8; Table~\ref{tab:fa4}); with the recipe's q rotation and sink merge it stays ahead at batch 4 and above and falls to about 0.85$\times$ at batch 1 and 16K (Section~\ref{sec:kernel}). It is 1.26--2.62$\times$ faster than BitDecoding's Hopper INT4 kernel. Converting the 4-bit K, and then also V, to FP8 by bit placement so that they run on FP8 tensor cores (C1 and C2, Sections~\ref{sec:c1} and~\ref{sec:c2}) raises the margin over FP8 to 1.22--1.76$\times$.
  \item \textbf{The gain inside vLLM.} Against vLLM's FlashInfer FP8 path (the faster of vLLM 0.33's two FP8 backends, with XQA decode), a decode step on Llama-3.1-8B is 1.12--1.64$\times$ faster at batch 8 and above and the cache holds 1.63$\times$ as many tokens. In steady state, the brackets put our sustainable rate at no less than about 1.31$\times$ FP8's on Qwen3-32B (bracket 1.31--1.60$\times$) and 1.14$\times$ on Qwen2.5-72B (1.14--1.50$\times$), and at 1.00--1.25$\times$ on Llama-3.1-8B, but they rest on rates added after the first results, mostly with one seed, and the test fixed in advance gave no valid verdict, so we do not count a higher sustained rate as established (Section~\ref{sec:e2e}, Table~\ref{tab:capacity}).
  \item \textbf{What it costs in accuracy.} With the recipe (K mean removal, K rotation, four BF16 sinks, layer 0 in BF16), the 95\% intervals exclude accuracy losses larger than about 3.5 points against BF16 KV on five models with repeated runs. In Qwen2-family models, several of which collapse under FP8 KV in vLLM 0.33, layer 0's K mean is the known key-projection bias~\citep{liu2025quantreasoning} seen through RoPE (rotary position embedding; 99.7--100\% of its energy, computed from the weights alone), and keeping only layer 0 in BF16 repairs FP8 (on one model only partly) and, with K mean removal, 4-bit KV; layer-0 statistics computed from the weights do not, however, predict which of these models collapse. On Qwen2.5-7B-Instruct, SmoothAttention does not prevent the collapse unless layer 0 is also kept, while per-channel K (KIVI) avoided it, there and on R1-Distill-Qwen-7B, which points to the per-token K formats that serving caches use (one run per model; Section~\ref{sec:accuracy}).
\end{enumerate}

\anonelse{}{\paragraph{Changes from version 1.} Version 1 reported, in its Table 9 and abstract, the highest request rate one H200 sustains within the latency targets as 1.23--1.32$\times$, 1.45--1.80$\times$ and 1.67--2.20$\times$ FP8's on Llama-3.1-8B, Qwen3-32B and Qwen2.5-72B. Those brackets came from 600-s arrival windows with one seed at most points. This version measures them again with steady-state runs and rules fixed before the results (Section~\ref{sec:e2e} and Appendix~\ref{sec:serving}); its results supersede version 1's capacity numbers as measures of the sustained rate. Version 1's brackets are kept as what they measured, the tolerance of a 600-s burst from an empty cache (Section~\ref{sec:e2e}). Under the pre-set test no model shows a gain at the design rates, and five of version 1's verdicts at the bracket ends do not hold up, four of them against our configuration (Appendix~\ref{sec:serving}). The other changes: the subtitle is now ``Without FP4 Hardware''. The kernel speed-ups are those of the paged NVFP4 kernel our plugin uses, timed in one job with the final settings: 1.10--1.48$\times$ the faster of XQA and FlashAttention-4's native FP8, where version 1 gave 1.12--1.49$\times$ for the contiguous kernel. Section~\ref{sec:c1} is new; it converts the 4-bit K, and then K and V, to FP8 by bit placement so that they run on FP8 tensor cores (1.15--1.70$\times$ and 1.22--1.76$\times$). The accuracy result is stated as the largest loss the 95\% intervals exclude, about 3.5 points, rather than as the absence of a significant difference. The main text is shorter: the layer-0 analysis, the serving details, the findings reported to vLLM and the detailed related work are now appendices, and a closing section collects lessons for kernel writers. New in this version: FlashInfer's NVFP4 and FP8 kernels on an RTX A5000 and an L40S, which check the issue-budget argument on GPUs with larger budgets (Section~\ref{sec:stall}), a comparison with SmoothAttention and KIVI (Section~\ref{sec:accuracy}), and the same layer-0 collapse in llama.cpp.}

\anonelse{We also list four findings about vLLM's KV-cache paths}{We also reported four findings to vLLM} (Appendix~\ref{sec:vllm}) and collect practical notes for vLLM users on Hopper in Appendix~\ref{sec:practical}.

\section{Background and related work}
\label{sec:background}

\paragraph{Notation.} $B$ is the number of sequences in a decode batch and $S$ the context length in tokens. A \emph{row} is one token's K (or V) for one KV head: 128 values. $H_{kv}$ is the number of KV heads (8 for Llama-3.1-8B) and $G$ the number of query heads per KV head (4 for Llama-3.1-8B). Decode kernels split each sequence into chunks processed in parallel and combine the partial results in a \emph{split merge}, as in Flash-Decoding~\citep{dao2023flashdecoding}. \emph{Sink tokens} are the first four tokens of each sequence, which we keep in BF16 outside the 4-bit cache~\citep{xiao2023streamingllm,liu2024intactkv}. The \emph{recipe} is NVFP4 with K mean removal, Hadamard rotation of K, four BF16 sinks and layer 0 in BF16. KL is the Kullback--Leibler divergence of the next-token distribution under a quantized cache from that under BF16, averaged over tokens.

\paragraph{Formats.} Table~\ref{tab:formats} lists the formats per row (MX formats, \citealp{rouhani2023mx}; NVFP4, \citealp{nvidia2025nvfp4}).

\begin{table}[!tb]
\centering
\caption{Formats, per 128-value head row.}
\label{tab:formats}
\small
\begin{tabularx}{\linewidth}{@{}llLr@{}}
\toprule
Format & Element & Scale & Bytes per row \\
\midrule
BF16 & 16-bit float & none & 256 \\
FP8 (E4M3) & 8-bit float & per tensor & 128 \\
INT4 (per token, asymmetric) & 4-bit integer & FP16 scale and zero point per row & 68 \\
NVFP4 & E2M1 & E4M3 per 16 values, plus a per-tensor FP32 scale & 72 \\
MXFP4 / UltraQuant & E2M1 & E8M0 (power of two) per 32 values & 68 \\
\bottomrule
\end{tabularx}
\end{table}

\paragraph{Fast low-bit decode kernels.} On H100, PyTorch's INT4 decode kernels for grouped-query attention (GQA) reach 17--30\% of bandwidth~\citep{pytorch2024int4decoding}; on AMD MI355X, the UltraQuant paper~\citep{chakrabarti2026ultraquant} finds that 4-bit KV ``buys capacity, not kernel speed''. On Hopper, FOSCAR~\citep[affine folding;][]{zhuang2026foscar} and BitDecoding~\citep[TMA, warp specialization, wgmma;][]{du2025bitdecoding} decode INT4 KV caches, FServe~\citep{baek2026fserve} decodes INT4 KV to FP8 in registers inside vLLM, and FuTransformer~\citep{wang2026futransformer} keeps an INT4 copy of K in fragment order; BitDecoding-FP4~\citep{du2026bitdecodingfp4} and NVFP4-MLA~\citep{ni2026nvfp4mla} target native FP4 tensor cores (Blackwell, Rubin). Our kernel combines warp specialization (as in FlashAttention-3, \citealp{shah2024flashattention3}, and LiquidGEMM's W4A8 GEMM, \citealp{hu2025liquidgemm}), write-time layouts (as in Marlin, \citealp{frantar2024marlin}, and TensorBridge's NVFP4 weights on FP8 wgmma, \citealp{zou2026tensorbridge}) and folding (as in FOSCAR). It differs from FServe, the closest work, in the formats (E2M1 with block scales as well as INT4), in feeding the decoded values to FP16 wgmma as the register operand with the operands swapped, and in the write-time layout. Appendix~\ref{sec:relwork} describes each of these works and what it was compared with.

\paragraph{Accuracy of low-bit KV caches.} KIVI~\citep{liu2024kivi} and KVQuant~\citep{hooper2024kvquant} quantize keys per channel because of channel outliers; QuaRot~\citep{ashkboos2024quarot} and SAW-INT4~\citep{jia2026sawint4} rotate K with a Hadamard transform, and SAW-INT4 shows that per-token INT4 needs it. The ``smooth K'' of SageAttention~\citep{zhang2025sageattention} subtracts the per-channel mean of K. CarrierKV~\citep{luo2026carrierkv}, an edge accelerator, removes a per-head K mean before RoPE and finds that a Hadamard rotation alone lowers Qwen3-4B's 32K RULER score with NVFP4 keys from 79.4 to 57.7; our mean is static, calibrated and taken after RoPE, and our rotation follows it. For INT4 keys, FServe~\citep{baek2026fserve} subtracts an offline-calibrated fraction of the per-channel mean after RoPE (``shrunk centering'') and aligns per-channel ranges with a scale folded into q, and FireQ~\citep{baek2025fireq} combines pre- and post-RoPE scaling. FlagRot~\citep{shao2026flagrot} reports that whole-head Hadamard rotation raises the KL of NVFP4 caches on Qwen3-8B (by 215\%) and that choosing per sub-vector whether to rotate avoids this; BitDecoding-FP4's default recipe on Blackwell includes static key smoothing (a calibrated per-head mean), a Hadamard rotation and the first 16 tokens in BF16, close to ours. Jet-QSparse~\citep{tang2026jetqsparse} likewise combines an NVFP4 cache with calibrated key centering and at most two BF16 pages per sequence and layer on Blackwell (GB300), with FP8 prefill. KVQuant and IntactKV~\citep{liu2024intactkv} keep the first token in high precision. Concurrent work under review at ICLR 2027, \mbox{Easy-Kuant}~\citep{easykuant2026}, reports the same collapse of the Qwen2.5 family inside vLLM under FP8 and NVFP4 KV caches, attributes it to a handful of post-RoPE key channels that are large at every token, and keeps the one to four layers responsible in BF16 (details in Appendix~\ref{sec:layer0}). We reach the same layers from the other side: for this family the offending channels are a constant key bias, and removing the calibrated per-channel mean, which leaves attention unchanged and costs no capacity, is what repairs it (Table~\ref{tab:ladder}), with layer 0 in BF16 adding the rest. On Hopper inside vLLM, our accuracy results confirm the benefit of mean removal, rotation after it, sinks and layer-0 protection; what we add is stated in Section~\ref{sec:accuracy}.

\paragraph{Serving.} vLLM 0.33 offers FP8 KV through FlashAttention-3 (default) or FlashInfer (faster than FlashAttention-3 for FP8 KV on H100 in vLLM \#66590~\citep{vllm66590}, and on H200 in our runs), an NVFP4 KV cache decoded by FlashInfer (which provides the BF16, FP8 and NVFP4 decode paths of vLLM's FlashInfer backend on Hopper), and UltraQuant~\citep{chakrabarti2026ultraquant}, a 4-bit format (MXFP4 with K rotation). With FlashInfer and FP8 KV on Hopper, vLLM decodes with TensorRT-LLM's XQA kernel~\citep{trtllm}, shipped in FlashInfer, but only if a network check to NVIDIA's artifact server succeeds; otherwise it falls back to FlashInfer's own decode kernel without an error, so we log the resolved kernel (Section~\ref{sec:setup}). On NVIDIA, UltraQuant falls back to Triton~\citep{tillet2019triton} decode kernels and FlashAttention-2~\citep{dao2023flashattention2} prefill (the latter addressed by vLLM PR \#47531, \citealp{vllm47531}). The FP8 baseline is moving: open PRs bring the native FP8 of FlashAttention-4~\citep{zadouri2026flashattention4} to vLLM's SM90 FP8-KV path~\citep{fa206,vllm63139}; we measure FlashAttention-4's native FP8 in Section~\ref{sec:kernel} (Table~\ref{tab:fa4}).

\section{Why the 4-bit kernels we examined stall}
\label{sec:stall}

The 4-bit kernels we examined either execute two to six times Hopper's instruction budget per byte, or wait on serialized loads and barriers, or both; reaching bandwidth needs both problems fixed.

\paragraph{Issue budget.} An H200 issues at most about 33 trillion thread instructions per second (132 SMs $\times$ 4 schedulers $\times$ 32 lanes $\times$ 1.98 GHz) and reads 4.8 TB/s, so a kernel running at full bandwidth can spend about 7 thread instructions per byte, or about 3.9 per value at 72 bytes per 128-value row. This follows FOSCAR's Eq.~2. With the same formula and spec-sheet values, the budget is about 10 on H100 SXM and A100 (80 GB SXM) and about 53 on L40S, which is why software 4-bit decoding that pays off on A100 or L40S~\citep{lin2025qserve} can still be issue-bound on H200.

\paragraph{Measured instruction counts.} GPU performance counters were not available on our cluster, so we counted executed instructions with NVBit, NVIDIA's binary instrumentation tool ($B = 32$, $S = 16$K; Table~\ref{tab:nvbit}).

\begin{table}[!tb]
\centering
\caption{Measured instruction counts (NVBit, H200, $B = 32$, $S = 16$K). Instructions per byte = instructions per value $\times$ 128 $\div$ bytes per row; implied issue utilization = instructions per byte $\times$ measured bandwidth fraction $\div$ 7. Problem ``both'': over the budget and at most about half of the issue rate, so the kernel also waits; ``waiting'': within the budget and still far below the issue rate. XQA (FP8, no decoding) is a reference: within the budget and near the bandwidth limit (Section~\ref{sec:c1}). $^{a}$Without spin-waits. Per-byte counts and utilizations come from the unrounded counts and bandwidths. UltraQuant: vLLM 0.33's decode kernel for \texttt{-{}-kv-cache-dtype ultraquant\_4bit} (MXFP4, 68 bytes per row), called directly on a paged cache written by vLLM's own store kernel, H200 SXM.}
\label{tab:nvbit}
\small
\begin{tabularx}{\linewidth}{@{}Lllllll@{}}
\toprule
Kernel & Bytes/row & Instr./value & Instr./byte & Bandwidth & Implied issue util. & Problem \\
\midrule
Our first Triton NVFP4 kernel & 72 & 11.7 & 20.7 & 17\% & 52\% & both \\
FlashInfer NVFP4 & 72 & 8.3 & 14.7 & 19\% & 40\% & both \\
vLLM UltraQuant (Triton, as shipped) & 68 & 22.5 & 42.3 & 8\% & 50\% & both \\
BitDecoding INT4 (Hopper branch) & 68 & 3.6$^{a}$ & 6.8$^{a}$ & 25\% & about 24\% & waiting \\
\midrule
XQA FP8 (reference) & 128 & 3.9 & 3.9 & 88\% & 49\% & none \\
\bottomrule
\end{tabularx}
\end{table}

vLLM's UltraQuant kernel exceeds the budget about six times (42 instructions per byte, 79\% of them integer, bit and select operations, with byte-wide global loads and shared-memory stores), and our first Triton kernel and FlashInfer's two to three times; most extra instructions are integer and bit operations (IMAD, LOP3, PRMT), multiplies and type conversions, plus, in Triton, round trips of decoded values through shared memory (three to five per tile, depending on the Triton version). BitDecoding decodes cheaply (about 1.3 instructions per value) and stays within the budget without its spin-waits, but serializes decoding and the matrix multiply, spends 38\% of executed instructions spin-waiting on barriers, and puts the 4 query heads on wgmma's 64-row dimension, so softmax runs on 64 rows for 4 real ones: it executes 16.5 exponentials per real score, and its softmax instructions (about 1.3 per value) are over a third of its instructions outside the spin-waits. Our kernel addresses both.

\paragraph{The same kernel where the budget is larger.} The budget also says where software 4-bit decoding should stop being issue-bound. We ran FlashInfer 0.7.3's decode on NVFP4 and FP8 caches with one benchmark script (the timing of Section~\ref{sec:setup}, 16-token pages, the Llama-3.1-8B head shape, the fastest of FlashInfer's decode configurations per format and shape) on H200 and on two GPUs with larger budgets, an RTX A5000 (Ampere) and an L40S (Ada) (Table~\ref{tab:crossgpu}). On all three, both formats resolve to the same FlashAttention-2-style kernel source, compiled for each architecture, so the comparison holds the code fixed; on H200, FlashInfer's XQA FP8 kernel is faster still (Table~\ref{tab:xqa}), so the gap to the best FP8 kernel there is larger than this one. At batch 8 and above, the NVFP4 path is 1.28--1.69$\times$ slower than this FP8 kernel on H200, 1.06--1.15$\times$ faster on the A5000 and 1.56--1.69$\times$ faster on L40S. If the 14.7 instructions per byte and the 40\% issue utilization counted on H200 (Table~\ref{tab:nvbit}) carried over, the budget would cap the NVFP4 path at about 19\% of peak bandwidth on H200 ($0.40 \times 7 / 14.7$; equal to the measurement by construction), about 49\% on the A5000 and above the bandwidth limit on L40S; it reaches 13--20\%, 52--57\% and 55--80\% (on H200 these runs use 16-token pages and other shapes than Figure~\ref{fig:kernel}, whose 11 shapes give the 10--19\% of Section~\ref{sec:intro}). We did not count instructions on the two other GPUs, whose code is compiled for another architecture, so this agreement is consistent with the budget argument but does not test the counts; before the A5000 run we had predicted only the direction (faster than FP8, at a lower fraction of bandwidth than on L40S). The L40S margin does not depend on the quality of the FP8 baseline: an FP8 kernel at full bandwidth would still be 1.17--1.42$\times$ slower than the measured NVFP4 times at four of the five shapes, and level (0.97$\times$) at $8 \times 8$K. Inside vLLM 0.33 (Llama-3.1-8B, decode step at $8 \times 16$K), stock NVFP4 KV gives a 13\% shorter step than FP8 on L40S and 3.4\% shorter on an RTX A6000 (Ampere, the A5000's chip with more SMs), in the same order. The same source is thus slow where the budget is small and fast where it is ample: on H200, what holds 4-bit decoding back is its instruction count against the budget, not the format.

\begin{table}[!tb]
\centering
\caption{FlashInfer 0.7.3's decode on NVFP4 and FP8 caches across GPUs: one benchmark script, ms per call at four shapes with batch 8 and above per GPU (the ranges in the text cover every such shape of each run: 9 on H200, 5 on each other GPU), percentage of the spec-sheet peak bandwidth in parentheses. Budget: thread instructions per byte at full bandwidth, from spec-sheet SM count, boost clock and bandwidth (Section~\ref{sec:stall}). Every time is the resolved FlashAttention-2-style kernel (tensor-core or CUDA-core variant, whichever was faster). Ratio above 1: NVFP4 faster.}
\label{tab:crossgpu}
\small
\begin{tabular}{@{}llrrr@{}}
\toprule
GPU (budget) & $B \times S$ & FP8 & NVFP4 & FP8 / NVFP4 \\
\midrule
H200 SXM (7) & $8 \times 16$K & 0.119 (47\%) & 0.186 (17\%) & 0.64 \\
 & $8 \times 64$K & 0.401 (56\%) & 0.676 (19\%) & 0.59 \\
 & $32 \times 16$K & 0.400 (56\%) & 0.667 (19\%) & 0.60 \\
 & $64 \times 64$K & 3.057 (59\%) & 5.079 (20\%) & 0.60 \\
RTX A5000 (18) & $8 \times 8$K & 0.214 (82\%) & 0.189 (52\%) & 1.13 \\
 & $8 \times 32$K & 0.782 (89\%) & 0.688 (57\%) & 1.14 \\
 & $8 \times 128$K & 3.048 (92\%) & 2.888 (55\%) & 1.06 \\
 & $32 \times 32$K & 3.092 (90\%) & 2.910 (54\%) & 1.06 \\
L40S (53) & $8 \times 8$K & 0.261 (60\%) & 0.160 (55\%) & 1.64 \\
 & $8 \times 32$K & 0.830 (75\%) & 0.531 (66\%) & 1.56 \\
 & $8 \times 128$K & 2.958 (84\%) & 1.754 (80\%) & 1.69 \\
 & $32 \times 32$K & 2.954 (84\%) & 1.748 (80\%) & 1.69 \\
\bottomrule
\end{tabular}
\end{table}

\section{Design}
\label{sec:design}

\begin{figure}[!tb]
\centering
\begin{tikzpicture}[font=\small, box/.style={draw, rounded corners=2pt, align=center, inner sep=4pt}, >=Stealth]
\node[box, text width=2.7cm] (hbm) at (0,0) {HBM: paged 4-bit\\K, V codes + scales};
\node[box, text width=2.7cm] (prod) at (4.2,0) {Producer warp\\TMA bulk copies};
\node[box, text width=3.2cm] (smem) at (8.6,0) {Shared memory: 2 stages $\times$ 256 tokens (mbarriers)};
\node[box, text width=4.4cm] (wg0) at (2.6,-2.3) {Warpgroup 0 (tokens 0--127 of a stage): decode in registers, wgmma $S^\top{=}Kq^\top$, softmax, wgmma $O^\top{=}V^\top P^\top$};
\node[box, text width=4.4cm] (wg1) at (8.0,-2.3) {Warpgroup 1 (tokens 128--255): the same};
\node[box, text width=6.4cm] (merge) at (5.3,-4.3) {Merge the two softmax states; split merge with the BF16 sinks};
\draw[->] (hbm) -- (prod);
\draw[->] (prod) -- (smem);
\draw[->] (smem.south) -- ++(0,-0.45) -| (wg0.north);
\draw[->] (smem.south) -- ++(0,-0.45) -| (wg1.north);
\draw[->] (wg0.south) -- (wg0.south |- merge.north);
\draw[->] (wg1.south) -- (wg1.south |- merge.north);
\end{tikzpicture}
\caption{Kernel structure. Loads, decoding and the two matrix multiplies overlap: while the warpgroups decode and multiply one stage, the producer fills the other. Decoded K and $V^\top$ are wgmma's A operand, read from registers; query heads are the N = 8 dimension.}
\label{fig:pipeline}
\end{figure}
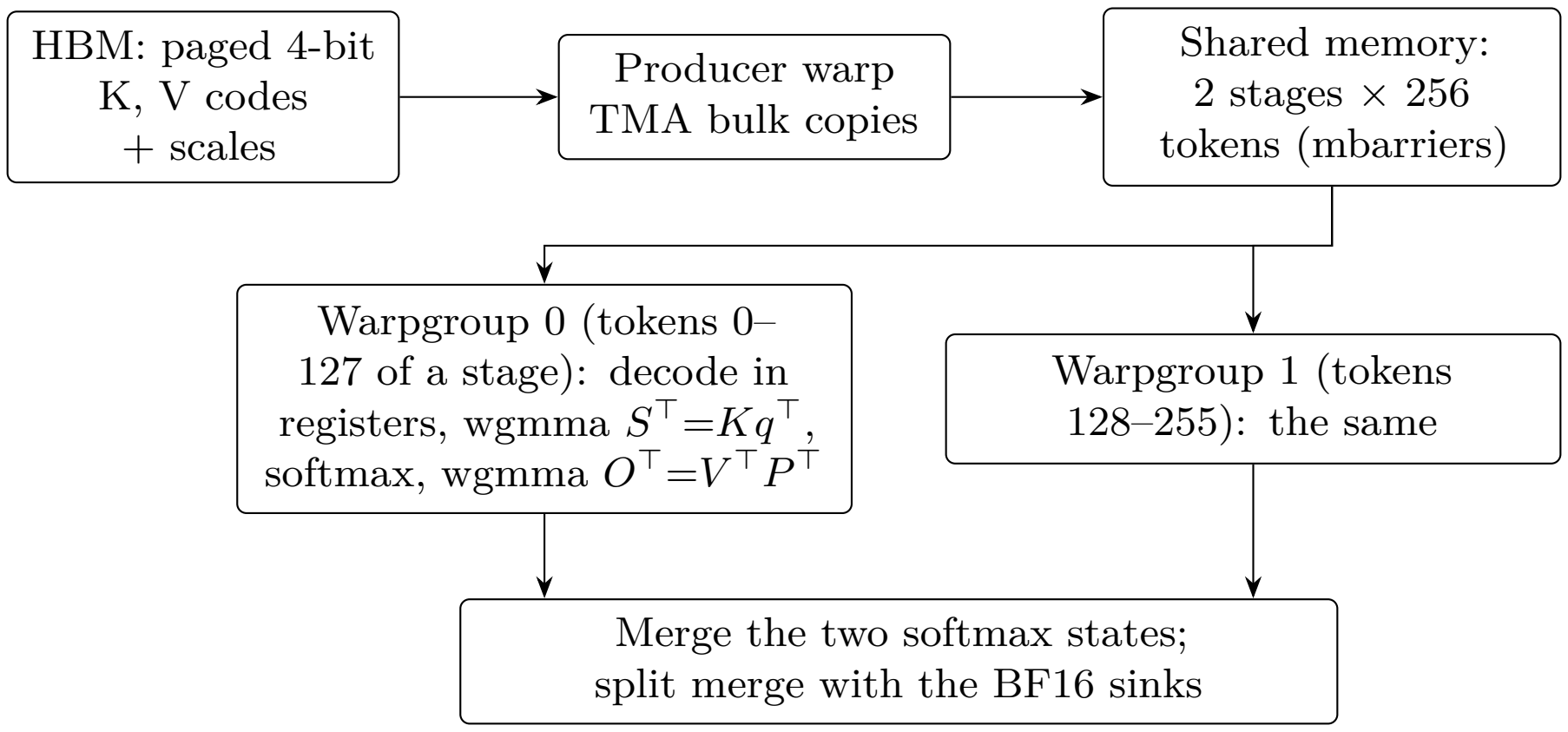

The kernel fixes waiting with warp specialization and fixes the instruction count by choosing the cache layout at write time; INT4, NVFP4 and MXFP4 share one structure (Figure~\ref{fig:pipeline}).

\begin{enumerate}
  \item \textbf{Overlap: a TMA producer and two compute warpgroups.} A producer warp streams 256-token stages of K and V (codes and scales) into shared memory with TMA bulk copies, double-buffered with mbarriers (Hopper's shared-memory barriers that track the completion of asynchronous copies). Each compute warpgroup (4 warps) owns half of every stage and keeps its own online-softmax state; the two merge once at the end.
  \item \textbf{Operand swap, decoded values straight from registers.} QK is computed as $S^{\top} = K \cdot q^{\top}$ and PV as $O^{\top} = V^{\top} \cdot P^{\top}$ with \texttt{wgmma.m64n8k16}. The decoded K or V$^{\top}$ is the A operand, which wgmma reads from registers, so decoded values never return to shared memory. Query heads take the N dimension ($N = 8$; 4 or 8 real heads), so softmax does no work on padding rows. Only P goes through shared memory (2 KB per warpgroup). GEMMs with a small M swap operands in the same way \citep[e.g.,][]{vllm20396,zou2026tensorbridge}, and BitDecoding-FP4 swaps both attention matrix multiplies on Blackwell~\citep{du2026bitdecodingfp4}; here the swap puts the small query group in N and lets software-decoded K and V$^{\top}$ be wgmma's register operand.
  \item \textbf{Write-time layout.} We control the cache format, so each K row and each V block (256 tokens in the dense kernel, one 64-token page in the paged one) is permuted when written. Each thread's register fragment then sits in contiguous 16-byte chunks: one 128-bit shared-memory load per thread per row, no bank conflicts, no shuffles. Scales are permuted the same way. Marlin~\citep{frantar2024marlin} and TensorBridge~\citep{zou2026tensorbridge} do the same for 4-bit weights, offline, FuTransformer~\citep{wang2026futransformer} for an INT4 K copy, and BitDecoding~\citep{du2025bitdecoding} packs KV in a tensor-core-aware layout; here the permutation happens in the cache-write kernel.
  \item \textbf{Cheap reconstruction.} INT4 uses FP16 magic numbers~\citep{kim2022elephants}: \texttt{(w \& 0x000F000F)}\allowbreak\ \texttt{| 0x64006400} gives $1024 + c$ for two codes, and masking with \texttt{0x00F000F0} gives $1024 + 16c$ with no shift; the factor 16 is folded into q (0.625 instructions per value). For E2M1, placing the exponent and mantissa bits at FP16 bits 9--11 and the sign at bit 15 gives the value times $2^{-14}$ exactly, including the subnormal 0.5.
  \item \textbf{Scales.} INT4's per-token scale and zero point are folded around the multiplies (FOSCAR's affine folding). NVFP4's per-16 E4M3 scales and MXFP4's per-32 E8M0 scales are applied per register with paired half-precision multiplies, because Hopper has no block-scaled tensor cores. In an earlier build of the kernel, NVBit counted 3.8 instructions per value for FP4 against 2.6 for INT4; the scales account for about 0.7 of that gap and the E2M1 reconstruction for most of the rest.
\end{enumerate}

\paragraph{Numerical pitfalls found by tests.} (1) The INT4 magic-number offset of 1024, subtracted after the multiply, grew to $2\times10^{-2}$ relative error over thousands of tokens, consistent with Hopper's tensor-core accumulation not being IEEE round-to-nearest~\citep{khattak2025tensorcores}; subtracting it before the multiply (one HADD2 per register) brings the error to $1.3$--$2.3\times10^{-4}$. Our random-data tests caught this one, but only once a split accumulated thousands of tokens. (2) On real Llama-3.1-8B KV, thousands of small attention weights next to a strong sink fell into FP16 subnormals; scaling P by $2^{10}$ fixes it at no cost. Random test data does not show this problem.

\paragraph{Paging and sinks.} The FP4 kernels read vLLM-style paged caches (64-token pages, a block table, per-sequence lengths) at the same speed as contiguous caches (0.99--1.035$\times$, pages shuffled); the INT4 kernel is dense only. Four BF16 sink tokens are kept outside the 4-bit cache and joined in the split merge (the \emph{sink merge}, which combines the split partials with the BF16 sinks; in its first, serial version every thread walks all splits in turn, and in the current, \emph{parallel} one the splits are spread over several warps); with K mean removal they store $k - \mu$ like every other key, so $q \cdot \mu$ still cancels in the softmax. The kernel runs 4 or 8 query heads per KV head; other $G \leq 8$ run by zero-padding q.

\paragraph{FP8 tensor-core variants.} Two variants of the paged NVFP4 kernel turn 4-bit codes into FP8 by bit placement so that they feed FP8 tensor cores: C1 does this for K, so $qK^\top$ runs on FP8 wgmma, and C2 for K and V, so $PV$ does too (Section~\ref{sec:c1}).

\section{Experimental setup}
\label{sec:setup}

\paragraph{Hardware.} H200 SXM (141 GB, driver 570.86.15) unless noted; some accuracy runs and two marked speed numbers used an H200 NVL node (driver 595.58.03), which is slower; no speed ratio mixes the two. H100 SXM5 (80 GB, driver 545.23.08) for the H100 kernel numbers (driver compatibility in Appendix~\ref{sec:kerneldetail}). The cross-GPU comparison of Section~\ref{sec:stall} also uses an RTX A5000 (24 GB, driver 570.86.15, peak 768 GB/s) and an L40S (48 GB, driver 595.58.03, peak 864 GB/s), and the SmoothAttention and KIVI runs of Section~\ref{sec:accuracy} an L40S. Bandwidth fractions are K and V bytes read ($B \times H_{kv} \times S \times 2 \times$ bytes per row; q, outputs, block tables, partial results and sinks not counted) divided by the spec-sheet peak (4.8 TB/s for H200, 3.35 TB/s for H100); a read-only kernel over 8 GB reaches 93--95\% of the H200 figure, and a device copy 87--90\% counting its read and write bytes (Appendix~\ref{sec:kerneldetail}).

\paragraph{Kernel benchmarks.} CUDA 12.8 (nvcc \texttt{-O3 -{}-use\_fast\_math}, \texttt{sm\_90a}), PyTorch 2.14.1 (CUDA 12.6 build), Triton 3.8.0, FlashInfer 0.7.3, FOSCAR at commit 9089e1f, BitDecoding's \texttt{hopper} branch (584d0d3), and the sources of vLLM PR \#53060 unmodified. FlashAttention-4 (Table~\ref{tab:fa4}) runs in its own CUDA 13 environment (PyTorch 2.14.0, flashinfer-python 0.7.4, which also provides that table's XQA). Random K, V and q in the Llama-3.1-8B head shape (8 KV heads, $G = 4$, head dimension 128). Each time is the median of 50 calls (eager or CUDA-graph replays) after 5 warm-ups, timed with CUDA events, with L2 flushed before every call. Such an isolated call includes fixed per-call costs (kernel ramp-up, the separate merge launch, writing back the flush's dirty L2 lines) that weigh most at batch 1: at $1 \times 16$K, the plugin's calls for one layer (q rotation, decode kernel at 16 splits, serial sink merge) take 22.8~\textmu s when replayed back to back over separate caches, as inside vLLM, and 29.0~\textmu s when timed in isolation (one job; Appendix~\ref{sec:prefill}). Our split count is the best of 1--64 per shape under the same timer, while FlashInfer and the other baselines use their own split heuristics, so the kernel tables slightly favor our kernel. Inside vLLM (Section~\ref{sec:e2e}) our split count is not tuned: it comes from a fixed rule, $\min(16, \lfloor 264 / (B \cdot H_{kv}) \rfloor)$, at least 1 (Appendix~\ref{sec:kerneldetail} compares it with the search); with the parallel sink merge, the current default, the cap at $B = 1$ is 33 instead of 16, and the runs we label ``current defaults'' use it. Tables with the recipe time a CUDA-graph replay. All columns of a kernel table come from one job on one GPU.

\paragraph{vLLM.} vLLM 0.33.0 (CUDA 13 wheel; flashinfer-python 0.7.2.post1 in this environment, one patch release older than the kernel-level FlashInfer), our plugin built with nvcc 13.2 for \texttt{sm\_90a} (it runs under the same compatibility package). The recipe uses a static per-head K mean calibrated offline and 64-token pages; the FP8 baselines use vLLM's defaults (16-token pages) with FlashInfer (\texttt{-{}-attention-backend}\allowbreak\ \texttt{FLASHINFER}) or the default FlashAttention-3 path. The page sizes therefore differ; at kernel level, FP8 XQA with 64-token pages is 0--4\% faster than with 16-token pages at our 11 shapes (Table~\ref{tab:xqa} takes the best of 16, 32, 64 and 128), so FP8's default pages cost it at most that share of its attention time. Inside vLLM, a whole FP8 decode step with 64-token pages takes 0.988--1.003$\times$ the time with 16-token pages on Qwen3-32B and Qwen2.5-72B at batch sizes near FP8's serving load (8--80 requests, 4K--8K tokens, \texttt{vllm bench forward}, XQA decode in both), so the page sizes do not explain the serving comparison in Appendix~\ref{sec:serving}.
GPU memory utilization is 0.90 for speed and serving runs and 0.85 for perplexity, accuracy and prefill-profile runs. FP8 numbers inside vLLM come from runs whose log shows FlashInfer's resolved decode kernel as XQA (\texttt{decode\_backend=xqa}; a few earlier runs predate this logging, Appendix~\ref{sec:kerneldetail}). The serving runs set \texttt{VLLM\_HAS\_FLASHINFER\_CUBIN=1} (Appendix~\ref{sec:serving}). Pure decode time uses \texttt{vllm bench forward} (dummy weights, CUDA graphs, median per shape, all configurations in one job). End-to-end numbers use \texttt{vllm bench latency} (time per output token $= (t_{256} - t_1)/255$, one warm-up and three timed iterations) and \texttt{vllm bench throughput} (random prompts, 256 requests).

\paragraph{Accuracy.} Greedy decoding; MATH-500 answers are compared after normalizing the last boxed expression. Paired comparisons use a permutation test on per-problem correctness, averaged over runs when there are several.

\section{Kernel results}
\label{sec:kernel}

\begin{figure}[!tb]
\centering
\includegraphics[width=\linewidth]{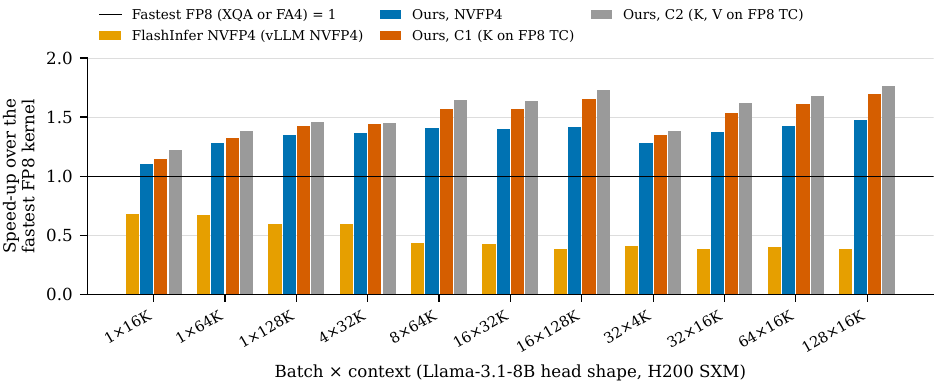}
\caption{Kernel speed relative to the fastest FP8 decode kernel at each shape (the faster of XQA and FlashAttention-4's native FP8; H200 SXM, Llama-3.1-8B head shape, all 11 shapes; higher is faster). Ours (paged NVFP4, C1, C2): the times of Table~\ref{tab:fa4} (one job, CUDA-graph replays). FlashInfer NVFP4, vLLM's NVFP4 KV path, comes from a second job on the same node that times it eagerly, so its bar is that job's fastest eager FP8 time over its eager time. The comparison with FlashInfer's FA2 FP8 kernel is in Table~\ref{tab:kernel}.}
\label{fig:kernel}
\end{figure}

On H200 the paged NVFP4 kernel our plugin uses is 1.10--1.48$\times$ faster than the faster of TensorRT-LLM's XQA and FlashAttention-4's native FP8 kernels at each of 11 shapes, C1 1.15--1.70$\times$ and C2 1.22--1.76$\times$ (Figure~\ref{fig:kernel}, Table~\ref{tab:fa4}). These are bare-kernel times, without the recipe's q rotation and sink merge (Section~\ref{sec:accuracy}). Measured in two other jobs at the same shapes, those add 0--7\% at batch 4 and above (9--13\% at $16 \times 128$K) and 12--31\% at batch 1 (Appendix~\ref{sec:kerneldetail}); scaling each shape's margin by that overhead, the recipe path stays ahead of the faster FP8 kernel at batch 4 and above (about 1.20--1.47$\times$) and at $1 \times 64$K and $1 \times 128$K (1.09--1.21$\times$), and is about 0.85$\times$ at $1 \times 16$K. This is an estimate across jobs (the bare-kernel times agree within 1\% between them), and it fits the whole decode step in vLLM being on par with FP8 at batch 1 (Section~\ref{sec:e2e}).
FlashInfer's NVFP4 kernel, which serves vLLM's NVFP4 KV cache, is 1.47--2.62$\times$ slower than that fastest FP8 kernel (Figure~\ref{fig:kernel}). Against FlashInfer's own FP8 kernels, our NVFP4 kernel is 1.4--2.3$\times$ faster than its FlashAttention-2-style kernel (FA2 below; selected shapes in Table~\ref{tab:kernel}) and 1.2--2.1$\times$ faster than its XQA kernel, the faster of those two at 9 of the 11 shapes (Table~\ref{tab:xqa}); FA2 is never the fastest FP8 kernel (Appendix~\ref{sec:kerneldetail}). Our INT4 is faster than FOSCAR's INT4 at equal bytes in all 15 shapes measured (Appendix~\ref{sec:other4bit}). BitDecoding's Hopper INT4 kernel, the closest earlier design, reaches 22--26\% of peak bandwidth at batch 4 and above and is 1.12--1.83$\times$ slower than the faster FP8 kernel at every shape; our paged NVFP4 kernel is 1.26--2.62$\times$ faster than it (Appendix~\ref{sec:other4bit}).

\paragraph{Against FlashAttention-4's native FP8.} An open pull request adds the native FP8 (E4M3 query, key and value) of FlashAttention-4~\citep{zadouri2026flashattention4}, FA4 in the tables, to Hopper and a second one makes vLLM select it for every SM90 FP8-KV model~\citep{fa206,vllm63139}; the reviewers state that it would then become vLLM's FP8 path on Hopper, so we measured it although neither had merged (versions in Appendix~\ref{sec:kerneldetail}). It is faster than XQA at 8 of the 11 shapes, by up to 1.5$\times$, and reaches 85--92\% of H200's peak bandwidth at batch 8 and above (71\% at $32 \times 4$K). Table~\ref{tab:fa4} sets it against the kernels the plugin runs: the paged NVFP4 kernel, C1 and C2. All three stay ahead of the faster FP8 kernel at every shape, and wherever FlashAttention-4 is the faster one our margin over it is smaller than over XQA. Times are CUDA-graph replays, as vLLM decodes, which is the conservative choice for comparisons with FlashAttention-4. Its error against an FP32 reference is smaller than the error that rounding the query to E4M3 introduces. The FP8 times reproduce those of an earlier job within 2.4\% at every shape but $16 \times 128$K (Appendix~\ref{sec:kerneldetail}).

\begin{table}[!tb]
\centering
\caption{Against the fastest FP8 decode kernel at each shape, once FlashAttention-4's native FP8 is included (H200 SXM, one job, CUDA-graph replays, best page size and split count per kernel, Llama-3.1-8B head shape). Ours: the paged kernels the plugin runs, with default settings except the split count (64-token pages in shuffled order; decode kernel plus the plain split merge, without sinks): NVFP4, C1 (K on FP8 tensor cores) and C2 (K and V; Section~\ref{sec:c1}). ``Best FP8'' is the smaller of the XQA and FlashAttention-4 times (in bold). Further notes in Appendix~\ref{sec:kerneldetail}.}
\label{tab:fa4}
\small
\begin{tabular}{@{}lrrrrrrrr@{}}
\toprule
& \multicolumn{3}{c}{Ours, paged (ms)} & \multicolumn{2}{c}{FP8 (ms)} & \multicolumn{3}{c}{Best FP8 / ours} \\
\cmidrule(lr){2-4}\cmidrule(lr){5-6}\cmidrule(lr){7-9}
$B \times S$ & NVFP4 & C1 & C2 & XQA & FA4 & NVFP4 & C1 & C2 \\
\midrule
$1 \times 16$K   & 0.0217 & 0.0208 & 0.0196 & 0.0275 & \textbf{0.0239} & 1.10$\times$ & 1.15$\times$ & 1.22$\times$ \\
$1 \times 64$K   & 0.0410 & 0.0396 & 0.0380 & 0.0705 & \textbf{0.0525} & 1.28$\times$ & 1.33$\times$ & 1.38$\times$ \\
$1 \times 128$K  & 0.0622 & 0.0591 & 0.0575 & 0.1254 & \textbf{0.0840} & 1.35$\times$ & 1.42$\times$ & 1.46$\times$ \\
$4 \times 32$K   & 0.0602 & 0.0571 & 0.0568 & 0.1210 & \textbf{0.0823} & 1.37$\times$ & 1.44$\times$ & 1.45$\times$ \\
$8 \times 64$K   & 0.1849 & 0.1653 & 0.1577 & 0.3443 & \textbf{0.2596} & 1.40$\times$ & 1.57$\times$ & 1.65$\times$ \\
$16 \times 32$K  & 0.1851 & 0.1655 & 0.1584 & 0.3434 & \textbf{0.2597} & 1.40$\times$ & 1.57$\times$ & 1.64$\times$ \\
$16 \times 128$K & 0.6882 & 0.5911 & 0.5657 & 1.3228 & \textbf{0.9757} & 1.42$\times$ & 1.65$\times$ & 1.72$\times$ \\
$32 \times 4$K   & 0.0605 & 0.0573 & 0.0559 & \textbf{0.0773} & 0.0788 & 1.28$\times$ & 1.35$\times$ & 1.38$\times$ \\
$32 \times 16$K  & 0.1862 & 0.1660 & 0.1576 & \textbf{0.2552} & 0.2556 & 1.37$\times$ & 1.54$\times$ & 1.62$\times$ \\
$64 \times 16$K  & 0.3494 & 0.3088 & 0.2963 & 0.6141 & \textbf{0.4978} & 1.42$\times$ & 1.61$\times$ & 1.68$\times$ \\
$128 \times 16$K & 0.6805 & 0.5918 & 0.5711 & \textbf{1.0051} & 1.0541 & 1.48$\times$ & 1.70$\times$ & 1.76$\times$ \\
\bottomrule
\end{tabular}
\end{table}

\paragraph{Bandwidth.} The contiguous NVFP4 kernel reaches 67\% ($B = 32$) to 75\% ($B = 128$, both at 16K) of H200's peak and INT4 73--83\% (computed from the times in Table~\ref{tab:kernel}); at batch 1 it reaches far less (Section~\ref{sec:limits}). On H100 (3.35 TB/s), at batch 8 and above, NVFP4 reaches 79--86\% and INT4 82--90\%. MXFP4 reads 6\% fewer bytes than NVFP4 but takes the same time, so FP4 on H200 is limited by its instruction stream, not by bytes (Appendix~\ref{sec:ablationdetail}). Appendix~\ref{sec:kerneldetail} gives the kernel-level cost of the recipe's sink merge and q rotation.

\paragraph{What each design choice buys (cumulative ablation).} Starting from a kernel that keeps only the pipeline (the TMA producer warp, the mbarrier double buffer, the two compute warpgroups, split-KV and the merge), we add the design choices of Section~\ref{sec:design} one at a time (Table~\ref{tab:cumul}). Every step lowers the time, by 1.19--1.49$\times$; together they make the kernel 2.76$\times$ faster at $32 \times 16$K and 1.97$\times$ at $1 \times 64$K. The bit-level decode is the largest step at $32 \times 16$K, where 1.27$\times$ the instructions take 1.49$\times$ the time, which points to instruction latency and dependency chains rather than total issue throughput alone. Every variant keeps tokens on wgmma's M and the query heads on N, so the register step is a lower bound on what the operand swap buys; an unswapped kernel would also pay 8$\times$ the matrix-multiply work. Every variant's output is bit-identical to the shipped kernel's, whose compiled code the ablation switch leaves unchanged; Appendix~\ref{sec:ablationdetail} gives the remaining details.

\begin{table}[!tb]
\centering
\caption{Cumulative ablation (H200 SXM, one job, dense NVFP4 kernel, $G = 4$, Llama-3.1-8B head shape; split count fixed at the full kernel's best, 1 at $32 \times 16$K and 32 at $1 \times 64$K; each variant's own best split changes its time by at most 3\%). Time (fraction of peak bandwidth), speed-up over the previous row, and NVBit thread instructions per byte (including the merge kernel, 0.03--0.05). Pipeline only: the quantizer's natural layout (row-major nibbles), decoded values staged through shared memory, generic E2M1 decode (a 16-entry PRMT table, 15 instructions per 8 values; the bit-level decode takes 11). Last row: write-time layout and bit-level decode with staged operands. A device copy of the same bytes reaches 87\% and 76\% of peak, counting read and write bytes.}
\label{tab:cumul}
\small\setlength{\tabcolsep}{4pt}
\begin{tabular}{@{}lrrrrrr@{}}
\toprule
& \multicolumn{3}{c}{$32 \times 16$K} & \multicolumn{3}{c}{$1 \times 64$K} \\
\cmidrule(lr){2-4}\cmidrule(lr){5-7}
Variant & ms (peak) & Step & Instr./B & ms (peak) & Step & Instr./B \\
\midrule
Pipeline only                     & 0.5187 (24\%) & --           & 9.97 & 0.0820 (19\%) & --           & 10.08 \\
+ write-time layout               & 0.3578 (35\%) & 1.45$\times$ & 9.21 & 0.0626 (25\%) & 1.31$\times$ & 9.33 \\
+ decoded operand from registers  & 0.2799 (45\%) & 1.28$\times$ & 8.21 & 0.0525 (30\%) & 1.19$\times$ & 8.25 \\
+ bit-level decode (shipped)      & 0.1882 (67\%) & 1.49$\times$ & 6.44 & 0.0416 (38\%) & 1.26$\times$ & 6.54 \\
\midrule
Layout, staging, bit-level decode & 0.2911 (43\%) & --           & 7.55 & 0.0546 (29\%) & --           & 7.71 \\
\bottomrule
\end{tabular}
\end{table}

\paragraph{Where the time goes.} Dropping one part of the work at a time (Table~\ref{tab:ablation}, Appendix~\ref{sec:ablationdetail}) shows that decoding is a real but minority cost: without E2M1 reconstruction the kernel takes 0.87$\times$ the time at $B \geq 8$: the shipped decode costs about 13\% of the time, although with a generic decode the kernel takes 1.49$\times$ the time (Table~\ref{tab:cumul}). No single pipe is near its limit either: with per-instruction rates we measured on H200, the busiest pipe at $32 \times 16$K (integer and logic instructions) is about 41\% busy (Appendix~\ref{sec:kerneldetail}). Neither bytes nor the throughput of one pipe sets the kernel's time; instruction latency and dependency chains are the likely remaining limit, which we could not confirm without stall counters.

\section{Running 4-bit K and V on FP8 tensor cores}
\label{sec:c1}

This section changes the kernel; its in-vLLM numbers use the setup of Section~\ref{sec:setup}, and Section~\ref{sec:e2e} gives the end-to-end comparison they build on. Instruction counts explain why our margin over XQA stays below the 1.78$\times$ byte ratio where XQA runs near bandwidth, and suggest the next step. At $32 \times 16$K, XQA executes 3.9 thread instructions per value and our NVFP4 kernel 3.6. Per value we are not more expensive, but a 4-bit cache holds 1.78$\times$ as many values per byte, so per byte we execute 6.4 instructions against XQA's 3.9, close to the issue budget of about 7; 30\% of XQA's instructions are barrier waits and their spin loops. In our kernel, 57\% of the instructions reconstruct and scale the 4-bit values.

\paragraph{K by bit placement (C1).} The reconstruction of K can be removed almost entirely. Every E2M1 value is exactly representable in E4M3: placing the sign, the two exponent bits and the mantissa bit of an E2M1 code into an E4M3 byte (sign at bit 7, the three remaining bits at bits 4--2) yields the same value times $2^{-6}$. With a write-time layout in which one 32-bit word holds the codes of four E4M3 bytes at the right offsets, five bitwise instructions turn eight codes into two FP8 A-operand registers, and $qK^\top$ runs on FP8 wgmma ($m64n8k32$, K from registers, q in E4M3 with one scale per sequence and KV head). NVFP4's E4M3 block scale is not a power of two and so cannot be folded into the exponent (an MXFP4 scale could be). Since one FP8 wgmma sums 32 products, K keeps one E4M3 scale per 32 dimensions instead of 16, every $k32$ step starts from a zero accumulator, and its result is scaled in FP32 after the multiply, as DeepSeek-V3 does for block-scaled FP8 GEMMs~\citep{deepseekv3}; Hopper's FP8 accumulation is not IEEE round-to-nearest~\citep{khattak2025tensorcores}.

\paragraph{Prior art.} FP4-to-FP8 bit conversion has been used for FP4 weights with power-of-two scales~\citep{wu2023zeroquantfp} and for MXFP4 activations in MoE training on Hopper~\citep{zhang2026fp4moe}; CUTLASS's Hopper mixed-input GEMMs pass the narrow operand through registers~\citep{cutlass55}; TensorRT-LLM's XQA feeds FP8 K from shared memory into FP8 matrix multiplies; and FServe~\citep{baek2026fserve} converts INT4 KV to FP8 in registers, as TensorBridge~\citep{zou2026tensorbridge} builds FP8 operands from NVFP4 weights. What is specific here is the handling of NVFP4's block scale. FServe's 4-bit values are INT4 integers with per-token group scales; an NVFP4 code carries an E4M3 scale every 16 values along the head dimension, which is the reduction dimension of $qK^\top$, and that scale is not a power of two, so it can be neither folded into the FP8 exponent nor applied once per row. C1 therefore regroups K's scales to 32 dimensions, one per FP8 wgmma step, starts each step from a zero accumulator and scales its result in FP32, and C2 regroups V's scales along the token axis (Section~\ref{sec:c2}).

\subsection{V on FP8 tensor cores (C2)}
\label{sec:c2}

C1 leaves V unchanged: it still reconstructs every 4-bit value to FP16 and scales it with half-precision multiplies before an FP16 matrix multiply. Hopper's FP8 wgmma takes only K-major operands and sums 32 products per step; in $O^\top = V^\top P^\top$ the summed dimension is the token axis, while NVFP4's V scales change every 16 channels of each token, so a scale would vary inside one sum. C2 therefore keeps V's E2M1 codes but gives them one E4M3 scale per channel and 32 consecutive tokens of a page (times the per-layer FP32 scale); a group never spans a page, and the page keeps its size. V's codes use C1's word layout read along tokens, so the same five bitwise instructions per eight values give the FP8 register operand of \texttt{wgmma m64n8k32} with $V^\top$ from registers, and nothing is transposed. P is rounded to E4M3 as $2^{s - m + 8} \in (0, 256]$, that is, scaled by 256 before rounding, while the softmax denominator is summed from the unrounded values; each $k32$ step starts from a zero accumulator and is scaled by its V scale in FP32, as K is in C1. Because decode appends one token at a time, C2's cache-write kernel re-encodes a group's earlier codes when a new token raises its scale, and C2 adds the BF16 sinks as one more partial in its own split merge; Appendix~\ref{sec:c2detail} gives the cache-write design, its cost and the correctness checks.

\subsection{Instruction counts, speed and accuracy}
\label{sec:c12results}

\paragraph{Instructions.} C1 removes 17\% of the instructions per value (2.99 against 3.61, dense kernels; 5.46 against 6.42 per byte, counting the 140 and 144 bytes per token and head that each reads). C2 removes a further 23\%: NVBit at $32 \times 16$K counts 2.27 instructions per value for C2's decode kernel against 2.96 for paged C1 (4.03 against 5.26 per byte of the 144 bytes per token and head that each paged kernel reads; one job, both built before the register restructuring of Appendix~\ref{sec:kerneldetail} was applied to four query heads; breakdown in Appendix~\ref{sec:c2detail}).

\paragraph{Kernel speed.} In the paged kernel with four query heads per KV head (Table~\ref{tab:fa4}; with the register restructuring of Appendix~\ref{sec:kerneldetail}, now the default), C1 is 1.12--1.16$\times$ faster than our NVFP4 kernel at batch 8 and above with 16K or more tokens per sequence and 1.04--1.06$\times$ at batch 1--4 and $32 \times 4$K, and 1.15--1.70$\times$ faster than the fastest FP8 kernel at each shape; Appendix~\ref{sec:kerneldetail} gives eight query heads. C2 is the fastest of our kernels at every shape: 1.04--1.05$\times$ faster than C1 at batch 8 and above with 16K or more tokens per sequence, 1.005--1.061$\times$ at batch 1--4 and $32 \times 4$K, and 1.22--1.76$\times$ faster than the fastest FP8 kernel. At batch 8 and above with 16K or more tokens it reaches 79--89\% of H200's peak bandwidth (C1 76--85\%, paged NVFP4 68--74\%), counting the 144 bytes per token and head that all three paged kernels copy (75--84\% and 74--83\% on the bytes C2 and C1 use; Appendix~\ref{sec:c2detail}). C2's 89\% is about 94\% of the 93--95\% that a read-only kernel reaches; the paged NVFP4 kernel's 68--74\% leaves clear headroom (Section~\ref{sec:limits}).

\paragraph{Accuracy.} The coarser K scale of C1 costs little: offline KL rises by 14--17\% on three models, and in vLLM the perplexity increase over BF16 KV grows by 0.03\% (Llama-3.1-8B) and 0.01\% (Qwen2.5-7B-Instruct) of BF16's perplexity. For C2, offline (simulated quantization, three models), the token-axis V scales raise the recipe's KL by 2--8\%, and the E4M3 P adds at most 5\% (13--30\% without the factor 256). In vLLM, C2's prefill perplexity (wikitext-2, $16 \times 8192$ tokens) rises over C1 by 0.026\% of BF16's perplexity on Llama-3.1-8B and 0.022\% on Qwen2.5-7B-Instruct. Prefill reads the cache through the dequantization kernel, so this measures the V format but neither the E4M3 P nor the re-encoding of decode-time groups. Generation exercises both: on R1-Distill-Llama-8B (MATH-500, three shuffled runs per configuration), C2 scores 1.60 points above C1 (95\% CI [$-$1.67, +4.87]), so a C2 loss larger than about 1.7 points is excluded at 95\% (Appendix~\ref{sec:c2detail}).

\paragraph{Inside vLLM.} On Llama-3.1-8B (q preparation fused into the existing rotation kernel, current defaults; FP8, the recipe and C1 in one job on H200 SXM), C1 makes a decode step 1.01$\times$ faster than our NVFP4 kernel at batch 1 and 1.03--1.09$\times$ at batch 8 and above, that is, 1.01--1.06$\times$ and 1.15--1.79$\times$ faster than vLLM's FlashInfer FP8 path. In 600-s serving runs under the reasoning-like stream of Section~\ref{sec:e2e} (the earlier design, not the steady-state runs of Table~\ref{tab:capacity}) it moves Llama-3.1-8B's sustainable rate from $[1.6, 1.65)$ to $[1.65, 1.7)$, 1.27--1.36$\times$ FP8's $[1.25, 1.3)$ in the same design, although the cache holds the same number of tokens (the larger models in Appendix~\ref{sec:serving}). C2's 4--5\% kernel gain over C1 at batch 8 and above with 16K or more tokens shrinks to at most 3.2\% per step, and only at large batch with long contexts, because attention is a minority of the step (Section~\ref{sec:e2e}; details in Appendix~\ref{sec:c2detail}). C2's main result is at the kernel level; it is opt-in in the plugin, and the other in-vLLM results in this paper do not use it.

\section{End-to-end in vLLM}
\label{sec:e2e}

Inside vLLM 0.33.0, with the recipe used for every accuracy number in Section~\ref{sec:accuracy}, our plugin decodes 1.12--1.64$\times$ faster than FlashInfer FP8 (XQA decode) at batch 8 and above, and is on par with it at batch 1 (1.01--1.05$\times$ per decode step, 0.96--1.03$\times$ in end-to-end time per output token). vLLM's own NVFP4 KV path is 1.4--2.3$\times$ slower than FP8 at batch 8 and above. Our cache holds 1.63$\times$ as many tokens, and time to first token is 1.5--6\% slower. In steady state the brackets put our sustainable rate at no less than about 1.31$\times$ FP8's on Qwen3-32B and 1.14$\times$ on Qwen2.5-72B and at 1.00--1.25$\times$ on Llama-3.1-8B, at rates that include ones added after the first results (``Sustained rate'' below; Table~\ref{tab:capacity}); we do not count this as established.

\begin{figure}[!tb]
\centering
\includegraphics[width=\linewidth]{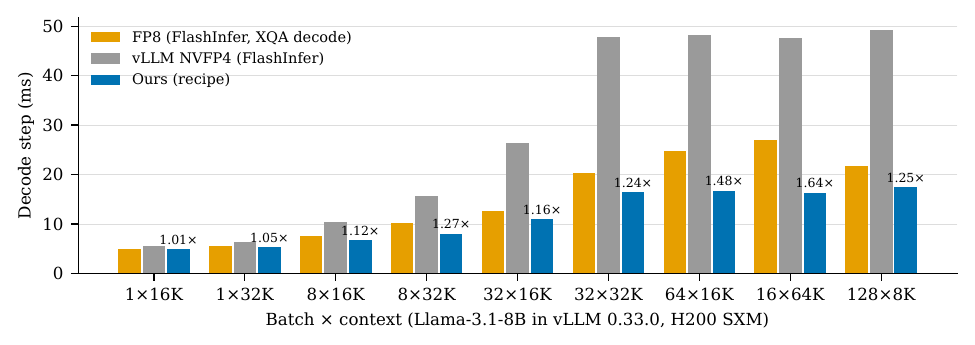}
\caption{Decode step time in vLLM (data of Table~\ref{tab:e2e}); labels give the speed-up.}
\label{fig:vllm}
\end{figure}

The plugin replaces the cache write, decode and prefill attention of vLLM's 4-bit backend with our kernels, and runs prefill attention on FlashAttention-3 in BF16. Table~\ref{tab:e2e} gives the pure decode time per step, and Figure~\ref{fig:vllm} plots it.

\begin{table}[!tb]
\centering
\caption{Pure decode time per step (\texttt{vllm bench forward}, Llama-3.1-8B, H200 SXM, all configurations in one job; median of 100 CUDA-graph replays). FP8: FlashInfer with XQA decode (resolved kernel logged); vLLM NVFP4: \texttt{-{}-kv-cache-dtype nvfp4}, FlashInfer's native decode; ours: the recipe, every sequence uses the sink merge. Speed-ups are over FP8.}
\label{tab:e2e}
\small
\begin{tabular}{@{}lrrrrr@{}}
\toprule
$B \times$ context & FP8 (ms) & vLLM NVFP4 (ms) & Ours (ms) & vLLM NVFP4 & Ours \\
\midrule
$1 \times 16$K   & 4.97  & 5.52  & 4.92  & 0.90$\times$ & 1.01$\times$ \\
$1 \times 32$K   & 5.57  & 6.42  & 5.31  & 0.87$\times$ & 1.05$\times$ \\
$8 \times 16$K   & 7.50  & 10.43 & 6.69  & 0.72$\times$ & 1.12$\times$ \\
$8 \times 32$K   & 10.25 & 15.69 & 8.06  & 0.65$\times$ & 1.27$\times$ \\
$32 \times 16$K  & 12.66 & 26.36 & 10.94 & 0.48$\times$ & 1.16$\times$ \\
$32 \times 32$K  & 20.42 & 47.80 & 16.43 & 0.43$\times$ & 1.24$\times$ \\
$64 \times 16$K  & 24.80 & 48.23 & 16.74 & 0.51$\times$ & 1.48$\times$ \\
$16 \times 64$K  & 26.90 & 47.59 & 16.35 & 0.57$\times$ & 1.64$\times$ \\
$128 \times 8$K  & 21.81 & 49.18 & 17.43 & 0.44$\times$ & 1.25$\times$ \\
\bottomrule
\end{tabular}
\end{table}

A rerun with the current defaults gives the same speed-ups within 0.013 (Appendix~\ref{sec:otherspeed}). On a vLLM development build in which FP8 KV runs on FlashAttention-4's native FP8, the recipe is 1.04--1.24$\times$ faster per step at batch 8 and above and on par at batch 1 (Appendix~\ref{sec:fa4vllm}). The per-step gain is smaller than at kernel level because attention is a minority of the step: each step also reads the 16 GB of BF16 weights, and the gain grows with context and batch (Appendix~\ref{sec:prefill}). We compare with FlashInfer FP8 because vLLM's default FP8 path (FlashAttention-3) was slower in our runs; against the default our ratios are larger (Appendix~\ref{sec:fa4vllm}). Prefill is 7\% slower than with FlashInfer FP8 (Appendix~\ref{sec:prefill}). Appendix~\ref{sec:serving} gives end-to-end latency and throughput, and Appendix~\ref{sec:otherspeed} the cost of mean removal and sinks, UltraQuant's own format and other models.

\paragraph{Serving under load.} We ran \texttt{vllm serve} with Poisson arrivals at fixed request rates. A request meets the service-level objective (SLO) if its time to first token is at most 2 s and its time per output token at most 100 ms (the MLPerf Llama-3.1-8B server targets; setup and every run in Appendix~\ref{sec:serving}). Rate scans on Llama-3.1-8B with 600-s arrival windows (Table~\ref{tab:serve}) show the mechanism. Under a reasoning-like stream (1K input, 8K output), both configurations meet the SLO for every request up to 1.25 requests/s (two runs of ours that did not reproduce are discussed in Appendix~\ref{sec:serving}), ours with 23--27\% lower mean time per output token and half the cache use. From 1.35 on, FP8's cache is full and requests are preempted, while ours stays at or below 81\% cache use. With 16K inputs ours is worse once prefill is the limit (Appendix~\ref{sec:serving}). In these scans cache use is still rising when arrivals stop (FP8 at 1.25 requests/s: 74\% at 480 s, 91\% at 600 s, peak 95\% at 660 s), so they describe a 600-s burst, not a steady state. We therefore report two quantities and keep them apart. Burst tolerance is the highest rate a GPU absorbs for 600 s of arrivals starting from an empty cache; at the boundary rates of these scans it is $[1.6, 1.65)$ requests/s for ours against $[1.25, 1.3)$ for FP8 on Llama-3.1-8B, $[0.4, 0.45)$ against $[0.25, 0.275)$ on Qwen3-32B and $[0.25, 0.275)$ against $[0.125, 0.15)$ on Qwen2.5-72B, that is 1.23--1.32$\times$, 1.45--1.80$\times$ and 1.67--2.20$\times$ FP8's rate. The larger cache is what lets ours absorb a longer burst: at 1.35--1.5 requests/s on Llama-3.1-8B FP8's cache was full while ours stayed at or below 81\%. These brackets rest on one seed at most points, with as few as 150 requests per point on Qwen2.5-72B, and they depend on the burst length and on starting from an empty cache (Appendix~\ref{sec:serving}). The sustained rate, which Table~\ref{tab:capacity} reports, comes only from the steady-state runs below.

\paragraph{Sustained rate.} For serving cost the number that matters is the highest request rate one GPU sustains, not the burst it absorbs from an empty cache. We measured it with steady-state runs: arrivals for a warm-up and then a measured window of 1,800~s (2,700~s on Qwen2.5-72B), of which only the requests arriving in the window count, and a run holds a rate if at least 99\% of them meet the SLO. The first rates were the boundary rates of the 600-s scans, under rules fixed in advance; they proved too high for ours on every model, so we added lower rates, each with its prediction fixed before it ran (Table~\ref{tab:capacity}; the rules and every run are in Appendix~\ref{sec:serving}). Over all rates run, the brackets put our sustainable rate at no less than about 1.31$\times$ FP8's on Qwen3-32B (ratio range 1.31--1.60$\times$; the upper end lies at a rate between two that were run) and 1.14$\times$ on Qwen2.5-72B (1.14--1.50$\times$); on Llama-3.1-8B the range, 1.00--1.25$\times$, includes no gain. These brackets rest on the added rates, most with one seed, and the test fixed in advance gives no valid verdict on any model, so we report them as measurements, not as the outcome of that test. A longer window is stricter: FP8 on Qwen3-32B met the SLO for 94.3\% of requests in a 3,600-s window at a rate it held in 1,800~s, while ours at 0.33 met it for all 1,166 requests in a 3,600-s window (one post-hoc run), so at that rate the longer window did not change our verdict.

\begin{table}[!tb]
\centering
\caption{Sustained request rate of one GPU under steady-state load (reasoning-like workload: 1K input, 8K output, lengths uniform within $\pm$50\%; Poisson arrivals; H200 SXM; four API-server processes; FP8 with XQA decode). Arrivals cover a warm-up $W$ and a measured window $M$ ($W = 600$ s and $M = 1800$ s on Llama-3.1-8B and Qwen3-32B, 900 and 2700 s on Qwen2.5-72B); only requests arriving in $M$ count. Unmarked rates are the design rates of the test fixed in advance (the boundary rates of the earlier 600-s brackets); $^{\dagger}$rates added after the design results, each with its prediction fixed before it ran, mostly with one seed (post-hoc; Appendix~\ref{sec:serving}). Runs: share of measured requests that meet the SLO (time to first token $\leq$ 2 s, time per output token $\leq$ 100 ms), the number of measured requests and the steadiness verdict (st: steady, hf: fails the headroom rule, ns: not steady), one entry per seed in seed order; seed 2 is an extra seed added by the rule of Appendix~\ref{sec:serving}. A run holds if its share is at least 0.99. Steadiness does not enter the verdicts; a run that is not steady or fails the headroom rule calls for an extra seed (Appendix~\ref{sec:serving}). The three ns runs on Llama-3.1-8B (share 1.000, mean cache use 23--35\%, falling) got none: FP8's uncertain endpoints count as held, so no second seed could move its bracket, and for ours at 1.0 only a failed second seed could have lowered our bracket end. Endpoint: held if every run holds, failed if every run fails, uncertain otherwise; $^{1}$single run. Bracket: [highest held rate, lowest rate not held) over all rates run, with our uncertain endpoints counted as failed and FP8's as held. Ratio: [our held rate / FP8's first rate not held, our first rate not held / FP8's held rate]; a dagger on a bracket end or a ratio marks one that rests on a post-hoc rate. The test fixed in advance, which uses the design rates only, gives no valid verdict on any model (Section~\ref{sec:e2e}). FP8 and ours of a design rate pair ran in one job on one GPU, except that ours on Llama-3.1-8B ran in jobs of its own after the paired jobs reached their time limit, as fixed in advance; post-hoc runs ran one configuration per job. KV tokens: vLLM's reported cache size (FP8 / ours) in the earlier serving runs with the same engine settings. Per step: pure decode speed-up over FP8 at $B \geq 8$ (\texttt{vllm bench forward}). Qwen2.5-72B-Instruct uses FP8 weights (RedHatAI FP8-dynamic) to fit on one GPU.}
\label{tab:capacity}
\footnotesize\setlength{\tabcolsep}{3pt}
\resizebox{\linewidth}{!}{\begin{tabular}{@{}lllllll@{}}
\toprule
Model & Config. & Rate & Runs: share (measured requests) steadiness, by seed & Endpoint & Bracket & Ratio \\
\midrule
Llama-3.1-8B                & FP8  & 0.75$^{\dagger}$  & 1.000 (1330) ns                                & held$^{1}$   & [1.0$^{\dagger}$, 1.1$^{\dagger}$)   & 1.00--1.25$\times$$^{\dagger}$ \\
KV 1.78M / 2.90M            &      & 0.85$^{\dagger}$  & 1.000 (1513) ns                                & held$^{1}$   &                                      & \\
per step 1.12--1.64$\times$ &      & 1.0$^{\dagger}$   & 1.000 (1787) st                                & held$^{1}$   &                                      & \\
                            &      & 1.1$^{\dagger}$   & 0.201 (1972) sa                                & failed$^{1}$ &                                      & \\
                            &      & 1.25              & 0.000 (2232) sa                                & failed$^{1}$ &                                      & \\
                            &      & 1.3               & 0.000 (2319) st                                & failed$^{1}$ &                                      & \\
                            & Ours & 1.0$^{\dagger}$   & 1.000 (1787) ns                                & held$^{1}$   & [1.1$^{\dagger}$, 1.25$^{\dagger}$)  & \\
                            &      & 1.1$^{\dagger}$   & 0.999 (1972) st                                & held$^{1}$   &                                      & \\
                            &      & 1.25$^{\dagger}$  & 1.000 (2232) hf; 0.887 (2236) hf               & uncertain    &                                      & \\
                            &      & 1.4$^{\dagger}$   & 0.311 (2496) hf                                & failed$^{1}$ &                                      & \\
                            &      & 1.6               & 0.012 (2860) st                                & failed$^{1}$ &                                      & \\
                            &      & 1.65              & 0.000 (2960) st                                & failed$^{1}$ &                                      & \\
\midrule
Qwen3-32B                   & FP8  & 0.25              & 0.996 (449) st; 1.000 (443) st                 & held         & [0.25, 0.275)                        & 1.31--1.60$\times$$^{\dagger}$ \\
KV 0.50M / 0.82M            &      & 0.275             & 0.246 (495) hf; 0.793 (489) hf; 0.272 (475) hf & failed       &                                      & \\
per step 1.05--1.18$\times$ & Ours & 0.3$^{\dagger}$   & 1.000 (542) st                                 & held$^{1}$   & [0.36$^{\dagger}$, 0.4)              & \\
                            &      & 0.33$^{\dagger}$  & 1.000 (586) st                                 & held$^{1}$   &                                      & \\
                            &      & 0.36$^{\dagger}$  & 1.000 (643) hf; 1.000 (633) st                 & held         &                                      & \\
                            &      & 0.4               & 0.200 (711) hf; 0.226 (708) hf                 & failed       &                                      & \\
                            &      & 0.45              & 0.004 (807) st                                 & failed$^{1}$ &                                      & \\
\midrule
Qwen2.5-72B (FP8 w.)        & FP8  & 0.125             & 1.000 (333) st; 1.000 (338) st                 & held         & [0.15, 0.175$^{\dagger}$)            & 1.14--1.50$\times$$^{\dagger}$ \\
KV 0.337M / 0.543M          &      & 0.15              & 0.985 (400) st; 0.995 (402) st                 & uncertain    &                                      & \\
per step 1.04--1.18$\times$ &      & 0.175$^{\dagger}$ & 0.081 (470) hf                                 & failed$^{1}$ &                                      & \\
                            & Ours & 0.175$^{\dagger}$ & 1.000 (470) st                                 & held$^{1}$   & [0.2$^{\dagger}$, 0.225$^{\dagger}$) & \\
                            &      & 0.2$^{\dagger}$   & 1.000 (542) ns; 1.000 (521) st                 & held         &                                      & \\
                            &      & 0.225$^{\dagger}$ & 0.957 (600) st                                 & failed$^{1}$ &                                      & \\
                            &      & 0.25              & 0.283 (672) hf                                 & failed$^{1}$ &                                      & \\
                            &      & 0.275             & 0.008 (733) st                                 & failed$^{1}$ &                                      & \\
\bottomrule
\end{tabular}}
\par\smallskip{\footnotesize\raggedright sa: saturated.\par}
\end{table}

\paragraph{Where the throughput gain comes from.} The per-step gain shrinks from 1.12--1.64$\times$ on Llama-3.1-8B (1.15--1.70$\times$ with FP8 weights; Appendix~\ref{sec:otherspeed}) to 1.04--1.18$\times$ on the larger models, whose step is dominated by the weight read (about 71 GB per step for the 72B model). Under overload, the output token rate is the number of running requests divided by the step time. On Qwen3-32B, with both configurations overloaded (FP8 at 0.275, ours at 0.4), ours runs 1.71$\times$ as many requests and serves 1.17$\times$ the output tokens per second; on Llama-3.1-8B (FP8 at 1.3, ours at 1.65) the ratios are 1.56$\times$ and 1.39$\times$. The mean inter-token latency seen by the client overstates the step under preemption, because a preempted request's next gap contains its wait. With gaps longer than 1 s removed, the step is 1.46$\times$ longer for ours on Qwen3-32B and 1.12$\times$ on Llama-3.1-8B (a larger batch, despite faster attention), and running requests divided by this step reproduce the throughput ratio within 0.1\% on both models. Both the larger cache and the faster step can raise the sustainable rate; we did not separate the two.

\paragraph{Caveats.} The serving runs use random prompts and one workload: 1,024-token inputs and outputs of 8,192 tokens $\pm$50\%, about 5,100 tokens of context per request in flight. A longer context raises each request's cost per decode step and fills the cache with fewer requests, and our kernel's advantage over FP8 grows with context (Section~\ref{sec:kernel}), so a longer-context workload may behave differently; we did not measure one. Ours uses 64-token pages and FP8 vLLM's default of 16; FP8's step time is the same with either page size within 1.2\%, and the larger pages cost ours about 0.5\% of its cache (Appendix~\ref{sec:serving}).

\section{Accuracy}
\label{sec:accuracy}

Inside vLLM (models from 7B to 32B and an MoE model), the 95\% intervals of the recipe against BF16 KV exclude accuracy losses larger than about 3.5 points on the five models with repeated runs and a paired test; three rows are single runs, with differences of $-$0.2 to +0.6 points. This bounds the loss; it does not show that there is none. We measure perplexity and task accuracy only; KV-cache quantization can also remove safety refusals at almost unchanged perplexity~\citep{xu2026alignmentcollapse}, which we did not test. Offline, it stays within our preset bound (perplexity +0.1, KL 0.02) on Llama-3.1-8B, Mistral-7B and Qwen3-8B-Base. Without layer-0 protection, several Qwen2-family models collapse, even under FP8 (R1-Distill-Qwen-1.5B and 7B, Qwen2.5-7B-Instruct; R1-Distill-Qwen-14B does not).

\begin{table}[!tb]
\centering
\caption{Each step of the recipe, added one at a time: mean KL against BF16 (offline, FP32, wikitext-2,
$8\times2048$ tokens, the K mean taken over each window; (a)--(c) without sinks, (d)--(f) with four BF16 sinks; FP8 as vLLM ships it, in every layer and without sinks; Qwen2.5-7B-I is Qwen2.5-7B-Instruct). The table measures what each step
contributes on one setup.}
\label{tab:ladder}
\small\setlength{\tabcolsep}{5pt}
\begin{tabular}{@{}llrrr@{}}
\toprule
& Step & Llama-3.1-8B & Qwen2.5-7B-I & Qwen3-8B-Base \\
\midrule
(a) & 4-bit FP4, one scale per tensor           & 0.297  & 10.1  & 3.00 \\
(b) & NVFP4: an E4M3 scale per 16 values        & 0.0269 & 5.90  & 0.0291 \\
(c) & + Hadamard rotation of K~\citep{ashkboos2024quarot} & 0.0187 & 6.95 & 0.281 \\
(d) & + 4 BF16 sink tokens~\citep{xiao2023streamingllm,liu2024intactkv} & 0.0168 & 7.05 & 0.249 \\
(e) & + K mean removal~\citep{zhang2025sageattention,shao2026flagrot}  & 0.0111 & 0.0167 & 0.0125 \\
(f) & + layer 0 in BF16                         & 0.0103 & 0.0147 & 0.0097 \\
\midrule
    & FP8, one scale per tensor (reference)     & 0.0037 & 3.86  & 0.0040 \\
\bottomrule
\end{tabular}
\end{table}

\paragraph{The recipe, step by step} (Table~\ref{tab:ladder}; offline, simulated quantization after RoPE). Block scales confine an outlier to its own 16 values: KL falls 11$\times$ on Llama and 103$\times$ on Qwen3-8B-Base. Rotation lowers Llama's KL by 30\% but does the opposite on Qwen3-8B-Base (0.029 to 0.281), whose K has a large per-channel mean that the rotation spreads into every block. Before the mean is removed, sinks help little. Mean removal, which cancels in the softmax and so costs nothing at read time, is the step that rescues the Qwen models (Qwen2.5-7B-Instruct from 7.05 to 0.0167, Qwen3-8B-Base from 0.249 to 0.0125); it needs the sinks. Layer 0 in BF16 lowers the KL by a further 7\% (Llama) to 22\% (Qwen3-8B-Base). Appendix~\ref{sec:accdetail} explains each step, sets the recipe beside each KV-cache option in vLLM (Table~\ref{tab:baselines}) and gives offline perplexity and long-context results.

\paragraph{Inside vLLM} (vLLM 0.33.0, recipe against BF16 KV; Table~\ref{tab:invllm}). Greedy decoding in vLLM is not reproducible across runs, because batch composition changes the numerics: on R1-Distill-Llama-8B, three BF16 runs score 68.4\%, 67.0\% and 67.6\% on MATH-500~\citep{hendrycks2021math,lightman2023verify}, and any two of them disagree on about 47 of 500 problems. So the reasoning rows use several runs with shuffled submission order, and a paired permutation test on each problem's mean correctness; a single-run difference within $\pm$3 points is not evidence either way. Runs used both H200 variants; within a row, BF16 and ours ran on the same variant, except one of the three R1-Distill-Llama-8B run pairs.

\begin{table}[!tb]
\centering
\caption{Inside vLLM (vLLM 0.33.0, H200, recipe against BF16 KV). For single runs no $p$ is given: a single-run test assumes fixed per-problem outcomes, which the run-to-run variation above contradicts. Intervals: two-level bootstrap (problems, then runs within each configuration, 2000 draws); $^{*}$one run, so the interval resamples problems only and omits run-to-run variation.}
\label{tab:invllm}
\small
\begin{tabularx}{\linewidth}{@{}P{4.6cm}LP{1.4cm}rrl@{}}
\toprule
Model & Test & Runs & BF16 & Ours & Ours $-$ BF16 [95\% CI] ($p$) \\
\midrule
Qwen2.5-7B-Instruct & MATH-500 & 5 & 71.2 & 72.0 & +0.8 [$-$1.6, +3.3] (0.49) \\
DeepSeek-R1-Distill-Qwen-7B & MATH-500 & 5 & 70.8 & 71.2 & +0.4 [$-$2.0, +3.1] (0.71) \\
DeepSeek-R1-Distill-Llama-8B & MATH-500 & 3 & 67.7 & 67.7 & +0.07 [$-$3.1, +3.3] (1.0) \\
Phi-4 & MATH-500 & 3 & 75.3 & 74.1 & $-$1.2 [$-$3.5, +1.1] (0.33) \\
Qwen3-32B, thinking on & MATH-500, first 200, 16K tokens & 3 & 90.0 & 90.3 & +0.3 [$-$2.7, +3.2] (0.89) \\
Qwen3-32B, thinking off & MATH-500 & 1 & 78.6 & 79.2 & +0.6 [$-$1.6, +2.8]$^{*}$ (--) \\
Qwen3-30B-A3B (MoE), thinking off & MATH-500 & 1 & 79.2 & 79.0 & $-$0.2 [$-$2.6, +2.2]$^{*}$ (--) \\
Llama-3-8B-ProLong-512k-Instruct & RULER-style~\citep{hsieh2024ruler} multi-key / multi-value / variable tracking, 32K--120K & 1 (450 paired trials) & -- & -- & +0.6 points (--) \\
\bottomrule
\end{tabularx}
\end{table}

MATH-500 rows use all 500 problems unless noted. Qwen2.5-7B-Instruct and R1-Distill-Qwen-7B have $G = 7$ and ran through our kernel with q zero-padded to 8 heads. In-vLLM perplexity is within 0.15\% of BF16 KV for ours on seven models (FP8: 0.04--0.95\% on six of them, and it collapses Qwen2.5-7B-Instruct; Appendix~\ref{sec:accdetail}). On Qwen2.5-72B-Instruct with FP8 weights, the largest model of Table~\ref{tab:capacity}, ours is also the more accurate configuration: wikitext-2 perplexity ($16 \times 8192$ tokens) is 0.95\% above BF16 KV for FP8 (0.88\% with layer 0 in BF16) and 0.03\% for ours. Within the recipe, mean removal and sinks matter: on Qwen2.5-7B-Instruct, our NVFP4 without K mean removal and sinks (layer 0 in BF16) scores 69.3\% over 5 runs, 2.7 points below the recipe ($p = 0.035$).

\paragraph{Reasoning and layer 0.} Inside vLLM (MATH-500, first 200 problems, one run), FP8 KV gives 0.5\%, 1.5\% and 0.0\% on R1-Distill-Qwen-7B, Qwen2.5-7B-Instruct and R1-Distill-Qwen-1.5B, and skipping layer 0 restores 71.5\%, 72.0\% and 61.0\%. On a fourth, Qwen2.5-1.5B-Instruct, it restores only part of the loss (GSM8K, 250 problems, one run: 60.0\% with BF16 KV, 1.2\% with FP8 KV, 52.8\% with layer 0 in BF16; Appendix~\ref{sec:layer0}). The collapse was reported before by \citet{li2025rllmserving} and by Easy-Kuant~\citep{easykuant2026}, its cause, the key-projection bias inherited from the Qwen base models, by \citet{liu2025quantreasoning}, and protecting the first layer by Easy-Kuant, \prbias~\citepbias{} and \citet{xu2026alignmentcollapse}. We add (i) that layer 0's large post-RoPE key channels are this bias seen through RoPE (99.7--100\% of layer 0's mean energy, computed from the weights alone), (ii) a control: the same R1 distillation on a Llama-3.1 base, which has no key bias, does not collapse, and (iii) a repair that keeps only layer 0 in BF16 and otherwise relies on mean removal, which costs no capacity (our cache holds 1.63$\times$ FP8's tokens, against 1.78$\times$ with 4 bits in every layer). Layer-0 statistics computed from the weights do not predict which bias-carrying models collapse: the R1 distillations collapse with less layer-0 key error than the 14B models tolerate (Appendix~\ref{sec:layer0}). The same collapse and the same one-layer repair also appear in llama.cpp, whose FP8 KV-cache code is unrelated to vLLM's: on R1-Distill-Qwen-7B (MATH-500, first 200 problems, L40S, one run each) it scores 72.5\% with an FP16 K cache, 0.5\% with its \texttt{f8\_e4m3} K cache and 72.5\% with only layer 0's K kept in FP16 (a local patch), and on all 500 problems on a V100 71.8\%, 0.6\% and 71.4\%. Appendix~\ref{sec:layer0} gives the full results, the prior work in detail and the fixed centers we tested inside vLLM.

\paragraph{Against SmoothAttention and KIVI.} Two published methods could make the layer-0 rule unnecessary: QServe's SmoothAttention~\citep{lin2025qserve}, which divides K's channels by per-channel factors (folded into q) before per-token INT4, and KIVI's per-channel K quantization~\citep{liu2024kivi}. We ran both, with predictions written before the runs, in a harness that simulates the quantized cache during decoding (Qwen2.5-7B-Instruct, MATH-500, first 200 problems, greedy, 4,096 generated tokens, one run per configuration on L40S; Table~\ref{tab:smoothkivi}). Per-token asymmetric INT4 for K and V collapses to 0\%, and so does SmoothAttention, although its factors were taken from each prompt's own keys, which favours it over QServe's offline calibration. With layer 0 in BF16, SmoothAttention recovers to 71.0\%, against 73.5\% and 74.0\% for the recipe at the same INT4 format and in NVFP4 (75.0\% with BF16 KV); with one run of 200 problems, none of the three differs significantly from BF16, and we do not rank them. KIVI (K per channel over 128-token groups, V per token over 32 dimensions, the incomplete last group kept in BF16) did not collapse, with or without layer 0 in BF16 (72.0\% both), against our prediction. A plausible reason, which we did not test directly: a per-channel zero point absorbs an offset that is constant along tokens, and on this model ($\texttt{rope\_theta} = 10^6$) the channels that carry layer 0's mean rotate by far less than a radian within a 128-token group. On this model, then, the collapse came with the per-token K formats that the paged caches of serving engines use (FP8 per tensor, NVFP4 per 16 values along the head dimension, QServe per token), not with 4-bit KV as such; per-channel K avoided it, but needs a token-group layout and a BF16 residual that these caches do not have. On a second model, DeepSeek-R1-Distill-Qwen-7B ($\texttt{rope\_theta} = 10^4$, by our arithmetic still far less than a radian per group for those channels), KIVI did not collapse either: 73.0\% against 71.5\% with BF16 KV (+1.5 points [$-$3.0, +6.0], worse / better 8 / 11, $p = 0.65$; same harness, first 200 problems, one run, predicted before the run). Two models, 200 problems and one run per configuration do not show that this holds in general. KVTuner~\citep{li2025kvtuner} reports KIVI-4 collapsing on the same model (GSM8K with 0--16 shots: 0.3--1.8\%, against 72--83\% with BF16 KV). Task, prompts and implementation differ (ours is an emulation, not KIVI's code), and we cannot say which difference matters; we report both numbers.

\begin{table}[!tb]
\centering
\caption{SmoothAttention and KIVI against the recipe (Qwen2.5-7B-Instruct, MATH-500 first 200 problems, greedy, 4,096 tokens, simulated quantization during decoding, one run per configuration, L40S). Recipe here: K rotation, K mean from the prompt, four BF16 sinks, layer 0 in BF16. Differences against BF16 with paired-bootstrap 95\% intervals; worse / better: problems BF16 solves and the configuration does not, and the reverse; $p$: exact McNemar test. Predictions were written before the runs.}
\label{tab:smoothkivi}
\small
\begin{tabularx}{\linewidth}{@{}Lrrrrl@{}}
\toprule
Configuration & Acc. & $-$ BF16 [95\% CI] & Worse / better & $p$ & Prediction \\
\midrule
BF16 KV & 75.0 & & & & \\
Per-token INT4, K and V & 0.0 & $-$75.0 & 150 / 0 & $< 10^{-4}$ & collapse: held \\
SmoothAttention (per-token INT4) & 0.0 & $-$75.0 & 150 / 0 & $< 10^{-4}$ & collapse: held \\
\quad + layer 0 in BF16 & 71.0 & $-$4.0 [$-$8.0, +0.5] & 14 / 6 & 0.12 & within 5 points: held \\
KIVI-4 & 72.0 & $-$3.0 [$-$7.0, +1.0] & 13 / 7 & 0.26 & collapse: failed \\
\quad + layer 0 in BF16 & 72.0 & $-$3.0 [$-$7.0, +1.0] & 11 / 5 & 0.21 & $> 20$ points below: failed \\
Recipe, INT4 (QServe's format) & 73.5 & $-$1.5 [$-$5.5, +2.5] & 10 / 7 & 0.63 & within 3 points: held \\
Recipe, NVFP4 & 74.0 & $-$1.0 [$-$5.5, +3.5] & 11 / 9 & 0.82 & within 3 points: held \\
\bottomrule
\end{tabularx}
\end{table}

\section{Limitations and next steps}
\label{sec:limits}

\begin{itemize}
  \item \textbf{Scope.} Head dimension 128 and 1--8 query heads per KV head. The kernel runs $G = 4$ or 8; other $G \leq 8$ run by zero-padding q (e.g.\ 7 to 8 for Qwen2.5-7B-Instruct and R1-Distill-Qwen-7B, used for their accuracy rows), wasting 1/8 of the tensor-core work. INT4 has only a dense kernel with $G = 4$. H200's budget of about 7 instructions per byte is the tightest of the GPUs we computed it for; H20, with far less compute per byte of bandwidth, should have a smaller one still, and we did not test it. Not covered: models with multi-head latent attention (MLA; DeepSeek-V3/R1), head dimension 256 (Qwen3.x, MiniMax, which vLLM's UltraQuant targets) or 64, and mostly sliding-window models.
  \item \textbf{Models.} Task accuracy inside vLLM on the seven models of Table~\ref{tab:invllm}, plus in-vLLM perplexity on Llama-3.1-8B, Qwen3-32B, Phi-4, OLMo-2-1124-7B, Qwen3-30B-A3B and Qwen2.5-72B-Instruct (FP8 weights); offline on Llama-3.1-8B, Mistral-7B and Qwen3-8B-Base. On Qwen2.5-72B-Instruct with tensor parallelism (TP = 2), the recipe gives correct answers under prefix caching, cross-request prefix hits and chunked prefill (correctness only; speed and accuracy at TP = 2 not measured).
  \item \textbf{Serving measurements.} One GPU type and random prompts. The rate scans of Table~\ref{tab:serve} (600-s windows) cover Llama-3.1-8B with one to five runs per rate. The brackets of Table~\ref{tab:capacity} come from steady-state runs on three models and one workload, with one to three runs per design endpoint and one per post-hoc rate, two where the extra-seed rule applied; a run of 8K-token outputs holds only 4--5 independent blocks of requests, so each verdict is uncertain (Appendix~\ref{sec:serving}). Near saturation, vLLM's default single API-server process becomes the limit for ours (Appendix~\ref{sec:serving}). The serving gain can come from the larger cache and from the faster decode step; we did not separate the two. Workloads with shared prefixes (a kernel-level result is in Appendix~\ref{sec:fa4vllm}), real traces, other workloads on the larger models, models beyond the three of Table~\ref{tab:capacity}, tensor parallelism and disaggregated prefill were not compared.
  \item \textbf{Long inputs and prefill.} Long-input streams gain nothing from the larger cache, because prefill compute limits them while the cache is far from full. In the 300-s scans of Table~\ref{tab:serve} (not steady-state runs), FP8 misses the 99\% line with peak cache use of 19\% and 57\% and no preemptions on the prefill-heavy stream (0.75 and 1.25 requests/s), and of 66\% on the long-context stream at 1.0 requests/s; ours, at 12--37\% peak cache use, is level or slightly worse (0.94 vs 0.98, 0.77 vs 0.87 and 0.84 vs 0.87, all with overlapping block-bootstrap intervals). Prefill itself is slower with ours: time to first token is 1.5--6\% longer end to end, a prefill step 1.7--4.5\% longer in a profile with both configurations in one job, and prefill alone 7\% longer for 8 prompts of 32,000 tokens (Appendix~\ref{sec:prefill}). That profile places the gap in prefill attention and the copies around it and finds our cache write as fast as FP8's; a further part, matrix multiplies running slower at the GPU's power cap, is inferred, not isolated. Future work: prefix-heavy traffic (multi-turn chat, retrieval-augmented generation, agents that resend the same system prompt and tools), where a cache that holds 1.63$\times$ the tokens could keep more prefixes cached and raise the prefix-cache hit rate; we have not tested this.
  \item \textbf{Small batch.} At $B = 1$ the kernel and its split merge reach only 18--38\% of bandwidth (16K--64K context) and the whole step is on par with FP8: the decode kernel is faster than FP8's, but weights dominate the step at batch 1. Removing the sink merge and the q rotation altogether, an upper bound for fusing them into the decode kernel, shortens the step by only 0.7--1.1\% at batch 1 (5.3\% at $8 \times 16$K), so we did not fuse them (Appendix~\ref{sec:prefill}).
  \item \textbf{Long-context accuracy.} The kernel's margin is largest at long contexts, where accuracy is weakest. On Llama-3.1-8B (multi-key retrieval at 120K, the same 100 trials offline and inside vLLM), the recipe loses about 10 points against BF16 KV offline (simulated quantization) and 15 inside vLLM (29\% against 44\%; 16 trials worse, 1 better; sign test $p = 0.0003$), while vLLM's default FP8 path (FlashAttention-3, which also quantizes the query to FP8) loses 13 inside vLLM (31\%, $p = 0.001$) and the recipe is level with it (9 worse, 7 better, $p = 0.80$). The loss is therefore not an artifact of simulated quantization. FP8 storage alone does not share it: offline, with one scale per layer and a BF16 query, FP8 KV loses 3 points on the same trials (5 worse, 2 better, $p = 0.45$) where the recipe loses 10 (11 and 1, $p = 0.006$); how much of the default FP8 path's loss comes from its FP8 query or its fixed scale of 1.0 we have not separated. On Llama-3-8B-ProLong-512k-Instruct, trained for long context, the recipe stays within 1.5 points of BF16 at 120K inside vLLM (50 trials per task) and within 3 offline (Appendix~\ref{sec:accdetail}), so what separates the two results is the model, not the framework. One run per configuration; other models at long context are untested.
  \item \textbf{Kernel headroom.} At batch 8 and above the paged NVFP4 kernel reaches 68--74\% of peak bandwidth, against 93--95\% for a read-only kernel and 79--89\% for C2; no pipe is near its limit, which points to instruction latency and dependencies (Section~\ref{sec:kernel}); we have not reduced them further.
  \item \textbf{Blackwell.} We did not run on Blackwell GPUs, which convert FP4 in hardware and whose tensor cores take block-scaled FP4 operands directly. Comparing ours on H200 with native NVFP4 on B200 would mostly measure the difference in memory bandwidth (about 4.8 against 8~TB/s), not the 4-bit method. On one Blackwell GPU, a port of our kernel and the native NVFP4 path would read about the same bytes per decode step (both store NVFP4), so we expect the decode gap between software unpacking and native FP4 to be small; the native path should win at large batch, where the unpacking work grows, and in prefill. This is a conjecture; we have not measured it.
  \item \textbf{FP8 baselines.} We compared with FlashInfer's FA2 and XQA FP8 kernels, vLLM's FlashAttention-3 FP8 path and FlashAttention-4's native FP8 (Table~\ref{tab:fa4}), not with TensorRT-LLM itself; where XQA runs near peak bandwidth our kernel-level gain is below the byte ratio (Appendix~\ref{sec:fa4vllm}).
  \item \textbf{Statistics.} The 95\% intervals in Table~\ref{tab:invllm} span about $\pm$2.5--3 points: with 200--500 problems and three to five runs per configuration, the smallest difference these comparisons would detect (2.8 bootstrap standard errors, $\alpha = 0.05$, power 0.8) is 3.3--4.6 points. They exclude accuracy losses larger than about 3.5 points, not smaller ones. Most of the interval comes from sampling problems, not runs (single-run intervals are almost as wide), so narrowing it needs more problems rather than more runs.
\end{itemize}

\section{Conclusion and lessons for kernel writers}
\label{sec:conclusion}

In our measurements on Hopper, a 4-bit KV cache decoded faster than FP8 only where software decoding fits the GPU's instruction-issue budget and does not serialize with loading and matrix multiplies. Our measurements suggest five lessons for low-bit kernels on Hopper-class GPUs. (1) Count instructions per byte against the issue budget (about 7 thread instructions per byte at full bandwidth on H200) before tuning anything else; the Triton and FlashInfer 4-bit kernels we counted spend two to six times that. (2) Do layout work when the cache is written, not when it is read: a write-time layout lets each thread load its fragment contiguously and feed decoded values to the tensor cores from registers. (3) Keep loading, decoding and matrix multiplies on separate warps (a TMA producer warp and two compute warpgroups) so that none waits for another. (4) For small query groups, swap the operands so the token dimension fills the tensor core's 64 rows and the query group needs at most an 8-wide N. (5) Check the compiled code: in a K-side prototype, three compiler-level wastes cost about a third of the instructions per value until NVBit exposed them (Appendix~\ref{sec:kerneldetail}). For baselines, log which kernel a framework actually resolved to; vLLM's FlashInfer FP8 path silently falls back from XQA when a network check fails. Beyond these lessons, a 4-bit format that embeds exactly in FP8 can use FP8 tensor cores without FP4 hardware: placing E2M1 bits into E4M3 bytes moves K (C1) and then V (C2) onto them and widens the kernel-level margin over the fastest FP8 kernel we measured from 1.10--1.48$\times$ to 1.22--1.76$\times$. \anonelse{We plan to propose the kernel to vLLM, for its UltraQuant format, and to FlashInfer, for NVFP4.}{We opened a tracking issue in FlashInfer for a Hopper NVFP4 paged decode kernel with a write-time page layout (\#10098) and plan to propose the kernel to vLLM for its UltraQuant format.}

\paragraph{Code availability.} \ifanon Code, result files and a map from every table to its jobs are provided as anonymized supplementary material.\else Code is available at \url{https://github.com/thomaszhang2661/kv4-hopper}.\fi

\section*{Broader Impact Statement}
More efficient LLM serving on existing Hopper GPUs reduces the hardware and energy needed per request. We see no specific risk of misuse beyond that of LLM serving in general.

\ifanon\else
\section*{Acknowledgments}
The author thanks Jiaji Huang for comments on this work and Ilmi Yoon for her support. This work was completed in part using the Explorer Cluster, supported by Northeastern University's Research Computing team.

\fi

\FloatBarrier

\bibliography{refs}
\bibliographystyle{tmlr}

\appendix
\section{Related work in detail}
\label{sec:relwork}

On A100, QServe~\citep{lin2025qserve} found a naive INT4 KV attention kernel slower than an INT8 one, because dequantization on CUDA cores exceeds the roofline in operations per byte; cheaper bit manipulation, FP16 instead of FP32 arithmetic and prefetched scales made it up to 1.5$\times$ faster than INT8. It reports no Hopper results. QServe, Marlin~\citep{frantar2024marlin} and our INT4 path convert integers to FP16 with magic numbers~\citep{kim2022elephants}. On H100, PyTorch's INT4 GQA decode kernels reach 17--30\% of bandwidth after extensive tuning, against 44--64\% for BF16~\citep[our arithmetic from the reported GB/s;][]{pytorch2024int4decoding}. FOSCAR~\citep{zhuang2026foscar} introduced affine folding for INT4 KV decoding and reports near-peak bandwidth on H100. BitDecoding~\citep{du2025bitdecoding} decodes INT4 KV on Hopper with TMA, warp specialization and wgmma, writing the dequantized values to shared memory for wgmma. BitDecoding-FP4~\citep{du2026bitdecodingfp4} targets Blackwell's native FP4 tensor cores (B300 and RTX PRO 5000 in its experiments), and NVFP4-MLA~\citep{ni2026nvfp4mla} stores and computes multi-head latent attention in NVFP4 on Rubin GPUs. LiquidGEMM~\citep[a W4A8 GEMM on Hopper;][]{hu2025liquidgemm} uses one load warpgroup and two compute warpgroups that dequantize in registers and feed wgmma directly; our kernel applies this structure to attention, where a softmax sits between the two matrix multiplies. FuTransformer~\citep{wang2026futransformer} fuses a Llama decode block on H100 with an INT4 copy of a BF16 KV cache, K stored in mma.sync fragment order and filled into registers through a shared-memory lookup table, per-token scales applied after the QK dot product and V transposed through shared memory; it is compared with PyTorch baselines, not with production FP8 KV attention kernels. FServe~\citep{baek2026fserve}, which follows FireQ~\citep{baek2025fireq}, is integrated into vLLM and measured on H100; it stores INT4 KV with per-token group scales, quantizes q to FP8 and dequantizes K and V to FP8 in registers, so that its FlashAttention-3-based prefill and decode kernels run on FP8 wgmma. Its decode kernel's effective (BF16-equivalent) bandwidth is competitive with, or above, FlashAttention-3 with an FP8 cache (its Figure 7), both take 1.52 ms in a profile of 132 variable-length requests (its Table 4), and it is not compared with XQA. We found no public implementation of FServe as of 2026-11-23 (we searched the paper, its ACL Anthology page and GitHub), so we compare only with its reported H100 numbers. TensorBridge~\citep{zou2026tensorbridge} runs NVFP4 weights on Hopper's FP8 wgmma as the register operand ($Y^\top = W X^\top$); it is a weight GEMM, not attention. On AMD MI355X, the UltraQuant paper~\citep{chakrabarti2026ultraquant} finds that 4-bit KV ``buys capacity, not kernel speed'': its decode attention takes 2.45$\times$ FP8's time with software dequantization and 1.18$\times$ with native FP4 (32K context, batch 8). An open vLLM PR (\#53060, \citealp{vllm53060}) adds a CUDA 4-bit decode path on Hopper for TurboQuant, vLLM's other family of low-bit KV formats, compared with BF16 FlashAttention and vLLM's existing TurboQuant path but not with an FP8 KV cache. LMDeploy's TurboMind~\citep{lmdeploy} has an INT4 KV cache (Appendix~\ref{sec:other4bit}). Section~\ref{sec:background} summarizes which of these techniques our kernel combines and how it differs from FServe.

\FloatBarrier
\section{Layer 0 in Qwen2-family models}
\label{sec:layer0}

This appendix gives the full results behind the paragraph ``Reasoning and layer 0'' of Section~\ref{sec:accuracy}, the prior work in detail, and the fixed centers we tested inside vLLM.

\paragraph{Results and prior work.} On DeepSeek-R1-Distill-Qwen-7B (offline, MATH-500, first 200 problems, 4,096 generated tokens), BF16 scores 73.5\%, FP8 KV 0.5\% and every 4-bit format 0--22.5\%. Keeping only layer 0 in BF16 restores FP8 to 73.0\% and the recipe to 73.0\%. Inside vLLM (first 200 problems, one run), FP8 KV collapses R1-Distill-Qwen-7B, Qwen2.5-7B-Instruct and R1-Distill-Qwen-1.5B and skipping layer 0 restores them (numbers in Section~\ref{sec:accuracy}); R1-Distill-Qwen-14B is unaffected (70.0\% BF16, 73.0\% FP8). The same R1 distillation on a Llama-3.1 base, which has no key bias, does not collapse: BF16 68.4\%, FP8 67.0\% ($-$1.4), FP8 with layer 0 in BF16 68.8\% (all 500 problems, one run each; within run-to-run noise). On R1-Distill-Qwen-7B, the per-channel mean makes up 98\% of layer 0's K norm (40--83\% in other layers), so FP8's step size is set by the mean and the token-dependent part is lost; plausibly, layer 0's error then passes through every later layer. Offline, perplexity shows the failure too (Qwen2.5-7B, the base model: 6.31 to 78.2 under FP8 KV, 6.59 with layer 0 in BF16). The collapse itself was reported before: an empirical study of reasoning-model serving with vLLM 0.8.1 and SGLang on RTX 4090, A6000 and A100 servers~\citep{li2025rllmserving} found R1-Distill-Qwen-7B at 4.33\% on MATH-500 with FP8-E4M3 KV and 0.33\% with E5M2 (4,096-token budget), gave no cause, and cautioned that KV quantization may degrade small reasoning models. For 4-bit caches, KVTuner~\citep{li2025kvtuner} reports uniform per-token asymmetric KV4 and KIVI-4 collapsing Qwen2.5-7B-Instruct on GSM8K (0--0.15\% and 0.3--1.8\% over 0--16 shots, against 72--83\% with BF16 KV) while Qwen2.5-14B and 32B are unaffected, and attributes it to channel-wise key outliers, without the key bias or layer 0. Protecting the first layer is also known practice (besides Easy-Kuant, below, \prbias~\citepbias{} recommends skipping layer 0, and \citet{xu2026alignmentcollapse} keep layers 0--1 of Qwen2.5-7B-Instruct in FP16 to restore its safety refusals under simulated 4-bit KV, attributing layer 0's sensitivity to outlier channels that inflate a per-tensor scale; they also report refusals falling from 99\% to 69\% with vLLM's FP8 E5M2 KV cache on that model). A calibrated per-tensor scale does not repair the FP8 collapse: in llama.cpp, R1-Distill-Qwen-7B with an \texttt{f8\_e4m3} K cache and a calibrated per-tensor K scale solves 0 of 200 problems (MATH-500, L40S, one run). Layer 0's largest key magnitude (about 186) is already below E4M3's maximum of 448, so a scale moves the representable range but does not change how much of each key the mean takes up. In llama.cpp's own KL check (wikitext-2 test, $20 \times 2048$ tokens, mean KL against an FP16 K cache, L40S), keeping only layer 0's K in FP16 lowers the FP8 KL from 3.16 to 0.19 on Qwen2.5-7B-Instruct and from 1.63 to 0.021 on R1-Distill-Qwen-7B, and the 4-bit \texttt{q4\_0} KL (rotated) from 5.75 to 0.15 and from 4.50 to 0.062, all far above llama.cpp's 8-bit \texttt{q8\_0} (0.0019 and 0.0013); on Llama-3.1-8B, which has no key bias, the layer-0 rule changes nothing (FP8 0.0049 and 0.0048, \texttt{q4\_0} 0.0135 and 0.0129). So in a second engine, too, layer 0 removes most of the damage but not all of it on Qwen2.5-7B-Instruct, consistent with the recipe's need for K mean removal at 4 bits.
The mechanism is not new either: \citet[Section 3.3, Appendix C.2]{liu2025quantreasoning} traced the K outliers of R1-Distill-Qwen-1.5B and 7B to the large key-projection bias inherited from their Qwen base models and showed layer 0's K before and after RoPE, for 3- and 4-bit KV with research quantizers rather than a serving engine's KV-cache kernels, and rtp-llm PR \#1583~\citep{rtpllm1583} subtracts a fixed center derived from the key bias, after RoPE, for per-token FP8 KV. That the bias survives RoPE in the slowly rotating dimensions and so reaches every key is noted in \prbias~\citepbias{}, which warns about it at startup. As noted in Section~\ref{sec:background}, \mbox{Easy-Kuant}~\citep{easykuant2026} reports the same collapse of the Qwen2.5 family inside vLLM under FP8 and NVFP4 KV caches (GSM8K 90.2\% to 11.0\% and 2.2\%), likewise finds that the error enters through keys, and attributes it to a handful of post-RoPE key channels that are large at every token, which a block-shared scale crushes; a per-layer statistic selects the one to four layers responsible (layer 0 among them on both checkpoints) and keeps them in BF16, at 85\% of NVFP4's cache capacity. Both Easy-Kuant and we find that keeping the first few tokens in BF16 does not repair the collapse (in our ladder the sinks help only once the mean is removed). Our additions are listed in Section~\ref{sec:accuracy}; the next paragraph gives the measurement behind the first.

\paragraph{Where layer 0's mean comes from.} In Qwen2-family models, layer 0's K mean is almost entirely the key-projection bias $b$ seen through RoPE: averaging $\mathrm{RoPE}(b)$ over positions removes 99.7--100\% of layer 0's mean energy on Qwen2.5-7B-Instruct, R1-Distill-Qwen-7B and Qwen2.5-72B-Instruct, computed from the weights and \texttt{rope\_theta} alone, without calibration. Subtracting $b$ per position before RoPE and adding it back at scoring time removes it exactly; that method is not ours (\anonelse{a vLLM PR~\citep{vllm59674anon}}{vLLM PR \#59674~\citep{vllm59674}} for INT4 KV and scratchy PR \#134~\citep{scratchy134}; \citet{liu2025quantreasoning}, Appendix C.2.2, quantize K before the bias is added). We only report measurements. Offline (simulated quantization in FP32, KL against BF16, not in our kernel or in vLLM, no task accuracy or speed), with the per-position subtraction every layer in 4-bit has 8--40\% lower mean KL than our recipe with layer 0 in BF16 (five Qwen2-family models, 2,048 tokens), and applying it to layer 0 alone is worse than keeping layer 0 in BF16. A static calibrated mean also degrades with context: on R1-Distill-Qwen-7B at 8,192 tokens, the static-center recipe's KL is 2.6$\times$ the per-position version's. Inside vLLM we tested only the simpler fixed center: one vector per layer subtracted from every key after RoPE, either $b$ itself (the center of rtp-llm \#1583) or $\mathrm{RoPE}(b)$ averaged over positions, with FP8 KV in all layers (H200 NVL, MATH-500 first 200 problems, 4,096 tokens, one run; perplexity on wikitext-2, $16 \times 8192$ tokens). On Qwen2.5-7B-Instruct it matches BF16 KV within single-run noise (73.0 and 71.5\% against 72.5\% with BF16 KV; perplexity 6.31 and 6.30 against 6.27, and 7.33 with layer 0 in BF16), but only partly restores the R1 distillations: 69.0 and 57.0\% on R1-Distill-Qwen-7B (BF16 KV 74.0\%, layer 0 in BF16 68.5\%, perplexity 32.1 and 21.5 against 19.3) and 48.0 and 50.0\% on R1-Distill-Qwen-1.5B (58.0 and 61.5\%; perplexity 52.7 and 31.4 against 26.0). None of the three is close to BF16 KV on all three models in these single runs: the fixed centers fall short on the R1 distillations, and layer 0 in BF16 raises Qwen2.5-7B-Instruct's perplexity to 7.33 and scores 68.5\% on R1-Distill-Qwen-7B (71.5\% in the run of Section~\ref{sec:accuracy}, an example of run-to-run spread). The main recipe keeps layer 0 in BF16 because, together with mean removal, it shows no significant accuracy difference from BF16 KV on the models of Table~\ref{tab:invllm}, whereas the fixed centers alone, with FP8 in every layer, fall short on the R1 distillations.
\paragraph{The per-position correction in task accuracy.} We also simulated it during decoding, in the harness of Section~\ref{sec:accuracy} (MATH-500 first 200 problems, greedy, 4,096 generated tokens, L40S, one run per configuration, predictions written before the runs): every prompt key and every generated key at position $p$ is stored as $Q(k - \mathrm{RoPE}(b, p) - \mu) + \mu + \mathrm{RoPE}(b, p)$, with $Q$ NVFP4 after the Hadamard rotation (as in the recipe) and $\mu$ the per-channel mean of the row's centred prompt keys; V and the four BF16 sinks are as in the recipe. With the correction in every layer, layer 0 included, NVFP4 K scored 73.0, 62.0 and 72.5\% on R1-Distill-Qwen-7B, R1-Distill-Qwen-1.5B and Qwen2.5-7B-Instruct, against 71.5, 56.0 and 75.0\% with BF16 KV (+1.5 [$-$3.5, +6.5], +6.0 [+0.5, +12.5] and $-$2.5 [$-$7.0, +2.0] points), and within 1.5 points of the recipe with layer 0 in BF16 on each model (73.5, 60.5 and 74.0\%; none of the differences significant). The control on R1-Distill-Qwen-1.5B, layer 0 in 4 bits with the static mean only, collapses to 8.0\%; in this simulation it is the per-position term that makes layer 0 quantizable. R1-Distill-Qwen-1.5B finishes only 117 of 200 BF16 answers within 4,096 tokens, and we do not read its higher scores under quantization as a gain. The correction is not in our kernel; its cost, one more dot product $q \cdot \mathrm{RoPE}(b, p)$ per key (or about 32 multiply-adds if only the 16 slowest rotary pairs are kept), is an estimate, not a measurement.

\paragraph{Which models collapse.} The bias explains where layer 0's mean comes from, but not which models collapse. Layer 0's input is the token embedding, so its keys and queries follow from the weights and the tokens alone; on 4,096 wikitext-2 tokens we computed two layer-0 statistics after the collapse outcomes were known: the mean-energy share of the post-RoPE K, and the signal-to-noise ratio of the attention scores under FP8 K (the variance of $q \cdot k$ over keys against the mean square of the FP8 error in $q \cdot k$; the constant $q \cdot \bar{k}$ cancels in the softmax). Qwen2.5-1.5B and Qwen2.5-7B (base and Instruct), which collapse, have a share of 0.997--1.000 and a score SNR of $-3$ to 0~dB in their worst KV head, so FP8's error exceeds the token-dependent part of the scores. But R1-Distill-Qwen-1.5B and 7B also collapse, with a share of 0.96 and an SNR of 17--19~dB, while Qwen2.5-14B, R1-Distill-Qwen-14B and Qwen2.5-14B- and 32B-Instruct, which do not (the last two as reported by KVTuner), have 0.90--0.91 and 15--17~dB; Llama-3.1-8B and R1-Distill-Llama-8B have 0.48--0.50 and 28--29~dB. The score SNR therefore does not separate the two groups, against our prediction; the share does in these eleven models, but by a margin of only 0.02--0.05 and only after the fact. A twelfth model, computed after we wrote down the prediction that its share would fall below 0.957, breaks it: Qwen3-8B, which has no key bias but normalizes each key head before RoPE, has a median share of 0.963 (worst-head SNR 26~dB) and does not collapse under FP8 keys in llama.cpp (62.0\% with an FP16 K cache, 60.0\% with \texttt{f8\_e4m3}; MATH-500 first 200 problems, one run). One model cannot tell whether the share does not apply to models with this key normalization or fails as a separator; either way we do not claim that it separates the two groups.

\paragraph{Layer 0 on another GPU with vLLM's stock NVFP4 cache.} On L40S (Ada), vLLM 0.33 serves \texttt{-{}-kv-cache-dtype nvfp4} with FlashInfer's kernel (Section~\ref{sec:stall}), and \texttt{-{}-kv-cache-dtype-skip-layers 0} keeps layer 0 in BF16; this is not our kernel and not the full recipe (no sinks; K rotation and the static K mean added through our plugin's hooks). On MATH-500 (first 200 problems, greedy, 4,096 tokens, one run each, paired against BF16 KV), Qwen2.5-7B-Instruct scores 72.5\% with BF16 KV and 2.0\% with FP8, so the collapse also occurs on Ada; NVFP4 with layer 0 in BF16 scores 69.0\% ($-3.5$ [$-8.5$, +1.5]), with K rotation 69.5\% and with rotation and the K mean 70.5\% ($-2.0$ [$-6.0$, +1.5]). On Llama-3.1-8B-Instruct, BF16 KV gives 44.5\%, FP8 45.0\%, NVFP4 with layer 0 in BF16 45.5\% and with rotation 41.5\% ($-3.0$ [$-9.0$, +3.0]). No difference from BF16 KV is significant except FP8 on Qwen2.5-7B-Instruct; with 200 problems the paired intervals span about $\pm$5--6 points, so these runs show that the layer-0 rule carries to a second GPU and a stock 4-bit kernel, not that the loss is below a few points.
Layer 0's key error alone does not decide the outcome: the R1 distillations collapse with less layer-0 damage than the 14B models tolerate, so the later layers or the long generations of a reasoning model also matter, which we did not measure. Two more Qwen2-family models collapse under FP8 keys, in a second engine: in llama.cpp (\texttt{f8\_e4m3} K cache, MATH-500 first 200 problems, L40S, one run each) Qwen2.5-1.5B-Instruct falls from 51.5\% to 3.5\% and Qwen2-7B-Instruct from 41.0\% to 0.0\%. Inside vLLM 0.33, Qwen2.5-1.5B-Instruct falls from 60.0\% to 1.2\% on 250 GSM8K problems (5-shot, H200 NVL, one run), and keeping layer 0 in BF16 restores it only to 52.8\%: on this model, unlike the three of Section~\ref{sec:accuracy}, the one-layer repair recovers most but not all of the loss.

\FloatBarrier
\section{Serving details}
\label{sec:serving}

This appendix has two parts: rate scans with 600-s arrival windows on Llama-3.1-8B (Table~\ref{tab:serve}, the earlier design) and the steady-state runs behind Table~\ref{tab:capacity} (from ``Steady-state runs'' below). In the scans we ran \texttt{vllm serve} with Poisson arrivals at fixed request rates (Llama-3.1-8B, H200 SXM; random prompts with input and output lengths uniform within $\pm$50\% of the nominal values; arrivals for 600 s, or 300 s for the 16K-input workload; both configurations one after the other in the same job, FP8 with XQA decode). The SLO is that of Section~\ref{sec:e2e}, and in the scans a rate counts as sustainable if at least 99\% of its requests meet it. Intervals are circular block-bootstrap intervals over arrival order (blocks of one p99 request latency, 2000 draws); they do not include run-to-run variation. Table~\ref{tab:serve} lists the runs.

\begin{table}[tbp]
\centering
\caption{Serving under load (Llama-3.1-8B, H200 SXM, FP8 with XQA decode). Share of requests meeting the SLO [95\% block-bootstrap interval over arrival order, blocks of one p99 request lifetime; requests of one run are not independent, and with fewer than about four blocks per run such intervals are still too narrow (Appendix~\ref{sec:serving})], peak KV-cache use, and FP8's preemptions (ours: none in the reasoning-like runs listed; 13 in the long-context run at 1.5). One vLLM API-server process (the default) and seed 0 unless noted. At 1.25, comma-separated values are seed 0 and seed 1 for both. At 1.35 and 1.5, comma-separated values are separate runs: FP8 seed 0 and seed 1; ours seed 0, a repeat of seed 0 with the same arrivals (on the same node at 1.35, another H200 SXM node at 1.5) and seed 1. ``4 API'': seeds 0 and 1 with \texttt{-{}-api-server-count 4}. Mean TPOT is time per output token; p99 TTFT is the 99th-percentile time to first token.}
\label{tab:serve}
\small\setlength{\tabcolsep}{2.5pt}
\begin{tabular}{@{}lrlrlrl@{}}
\toprule
Workload & Rate & FP8: SLO share & KV & Ours: SLO share & KV & Notes \\
\midrule
Reasoning-like  & 0.75 & 1.00 & 32\% & 1.00 & 16\% & mean TPOT 13.3 vs 10.3 ms \\
(1K in, 8K out) & 1.0  & 1.00 & 59\% & 1.00 & 29\% & mean TPOT 19.5 vs 14.3 ms \\
                & 1.25 & 1.00, 1.00 & 93--97\% & 1.00, 1.00 & 46\% & mean TPOT 28.2 vs 20.6 ms \\
                & 1.35 & 0.89, 0.94 & 100\% & 0.82, 1.00, 0.86 & 55--59\% & FP8: 628, 656 preemptions \\
                &      & 0.88, 0.94 & 100\% & 1.00, 1.00 & 58--61\% & 4 API; FP8: 683, 651 \\
                & 1.4  & 0.82 [0.66, 0.98] & 100\% & 1.00 & 62\% & FP8: 750 preemptions \\
                & 1.5  & 0.69, 0.75 & 100\% & 0.81, 1.00, 1.00 & 78--79\% & FP8: 1513, 1375 preemptions \\
                &      & 0.69, 0.75 & 100\% & 1.00, 1.00 & 79--81\% & 4 API; FP8: 1494, 1374 \\
\midrule
Long context    & 0.5  & 0.99 & 19\% & 0.99 & 11\% & \\
(16K in, 1K out) & 1.0 & 0.87 [0.75, 0.97] & 66\% & 0.84 [0.71, 0.95] & 35\% & \\
                & 1.5  & 0.14 [0.00, 0.29] & 100\% & 0.04 [0.00, 0.08] & 100\% & both prefill-bound \\
\midrule
Prefill-heavy   & 0.75 & 0.98 [0.95, 1.00] & 19\% & 0.94 [0.86, 0.99] & 12\% & p99 TTFT 2.2 vs 2.5 s \\
(16K in, 512 out) & 1.25 & 0.87 [0.76, 0.94] & 57\% & 0.77 [0.67, 0.86] & 37\% & p99 TTFT 4.0 vs 4.4 s \\
\bottomrule
\end{tabular}
\end{table}

\begin{itemize}
  \item \textbf{Reasoning-like} (1K input, 8K output). Up to 1.25 requests/s both meet the SLO for every request in these runs, and also with four API-server processes at 0.25 and 0.5 and, with seed 2, at 1.25 (where FP8's cache is full, with 190 preemptions); ours has 23--27\% lower mean time per output token, 12\% more output tokens per second at 1.25, and half the KV-cache use (FP8 is at 97\% at 1.25). From 1.35 on, FP8's cache is full in every run: requests wait and are preempted (628--1513 preemptions), and 6--31\% miss the time-to-first-token target. Our cache never exceeds 81\%, and at 1.4 every request meets the SLO (goodput, the requests that meet the SLO per second of the whole run including the generation tail: 1.01 against 0.71 requests/s, as FP8's run lasts 974 s against 830 s).
  \item \textbf{Our misses near saturation come from vLLM's API server, not from the cache.} With one API-server process, vLLM's default, ours met the SLO for 82\%, 100\% and 86\% of requests in three runs at 1.35, and for 81\%, 100\% and 100\% at 1.5. Every miss was a first-token delay of 2--6 s while no request was waiting in vLLM's scheduler and none was preempted, and the client sent every request on schedule. In the seed-1 run at 1.35, the API-server process used 92--97\% of one CPU core for about eight minutes (our server streams about 8,200 output tokens per second), and vLLM's own metrics place the delay in that process: 92 of the 814 requests the server counted (including the benchmark's 5 warm-up requests) had a time to first token above 2.5 s as the server measured it, while the scheduler's queue time summed to 0.009 s over all 814 and prefill took 69 ms on average. With four API-server processes (\texttt{-{}-api-server-count 4}, seeds 0 and 1), ours met the SLO for every request at both rates in both seeds (p99 time to first token at most 0.16 s at 1.35 and 0.69 s at 1.5), and FP8 did not change (0.88 and 0.94 at 1.35, 0.69 and 0.75 at 1.5, against 0.89--0.94 and 0.69--0.75 with one process; 651--1494 preemptions): its misses come with a full cache in every run. So at these rates a single API-server process limits vLLM before our cache does, and FP8's cache limits it before that.
  \item \textbf{Two earlier runs of ours that did not reproduce (history; Table~\ref{tab:capacity} does not use them).} Neither is excluded; both stay in the result files. (i) Seed 2 at 1.25 and 1.5 requests/s with one API-server process met the SLO for 5\% and 3\% of requests (1002 and 1202 preemptions). The GPU was slow for both configurations: with five requests running and no prefill, a decode step took 13.6 ms for ours and 13.0 ms for FP8, in the same job on the same GPU, against 5.4--6.0 ms in the 19 other one-process servers of both configurations; that FP8 server reached the time limit before writing a result. We could not find the cause (the logs record only the GPU and driver). A rerun of seed 2 at 1.25 (four API-server processes, another node) met the SLO for every request. (ii) Seed 0 at 0.25 and 0.5 requests/s with one API-server process met it for 97\% and 87\%, all misses being first-token delays of 2--3 s at 5--10\% cache use; its decode steps were normal and we found no host problem; a rerun with the same settings and one with four API-server processes met it for every request. A rule we wrote after these runs, which flags a run whose median time per output token over its first 20 requests exceeds $k$ times the median of all runs of the same model, workload and configuration (a statistic measured at low load, apart from the outcome), flags at $k = 5$ only the two runs of (i) and no FP8 run (82 ms against a median of 8.3 ms); at $k = 3$ it also flags FP8 and ours at the long-context workload's 1.5 requests/s, where prefill load rather than the host slows the first requests (we had predicted that $k = 3$ would flag only (i)); at $k = 2$ it adds ours at the prefill-heavy 1.5. Run (ii) is flagged at no threshold.
  \item \textbf{Long context} (16K input, 1K output). Both are sustainable at 0.5 requests/s and level at 1.0 (0.87 vs 0.84, intervals overlapping), although ours uses half the cache. At 1.5, 24K prompt tokens arrive per second, close to what either server can prefill (Appendix~\ref{sec:prefill}), both caches fill with partly prefilled requests, and both fail (0.04 vs 0.14, intervals overlapping; p99 time per output token 149 vs 94 ms, consistent with our 7\% slower prefill). A larger cache does not help a prefill-bound stream.
  \item \textbf{Prefill-heavy} (16K input, 512 output). FP8 keeps more requests within the SLO at both rates we ran: 0.98 vs 0.94 at 0.75 requests/s and 0.87 vs 0.77 at 1.25 (one run each; with a block bootstrap over arrival order the 1.25 intervals overlap, [0.76, 0.94] against [0.67, 0.86]), with neither cache near full and no preemptions. The 99th-percentile time per output token stays under the 100 ms target for both (89 vs 94 ms at 1.25), so almost all misses are first-token delays, consistent with our 7\% slower prefill; with short outputs there is no decode advantage to offset it. Ours alone at 0.5, 1.0 and 1.5 requests/s: 0.85, 0.65, 0.16 (no FP8 run at those rates).
\end{itemize}

\paragraph{XQA in the serving runs.} The serving runs set \texttt{VLLM\_HAS\_FLASHINFER\_CUBIN=1}, vLLM's switch for locally available kernels, because the serving script bypasses the cluster's proxy and vLLM's network check would otherwise fall back to FlashInfer's native decode (an earlier set of serving runs did, and is not used here).

\paragraph{Notes on Table~\ref{tab:capacity}.} KV tokens are vLLM's reported cache size at \texttt{-{}-gpu-memory-utilization 0.9} in the earlier serving runs, with the engine settings of the steady-state runs (\texttt{-{}-max-model-len 14336}, \texttt{-{}-max-num-seqs 512}); the latency and throughput runs of Section~\ref{sec:e2e} use other engine settings and report 2.87M and 1.76M on Llama-3.1-8B, the same 1.63$\times$ ratio. The per-step speed-ups use \texttt{vllm bench forward} with Llama-3.1-8B and Qwen3-32B in BF16 weights and Qwen2.5-72B in FP8 weights, its shapes up to $32 \times 8$K and $8 \times 31$K; the Qwen3-32B per-step run predates the logging of FlashInfer's resolved FP8 kernel (Section~\ref{sec:setup}).

\paragraph{Earlier 600-s brackets (history).} \anonelse{Before the steady-state runs, we measured}{Version 1 of this paper reported} brackets from 600-s arrival windows with $\max(150, 600\,\mathrm{s} \times \mathrm{rate})$ requests per point (seed 0, seed-1 repeats at some boundary points): FP8 $[1.25, 1.3)$ and ours $[1.6, 1.65)$ requests/s on Llama-3.1-8B, $[0.25, 0.275)$ and $[0.4, 0.45)$ on Qwen3-32B, $[0.125, 0.15)$ and $[0.25, 0.275)$ on Qwen2.5-72B, that is 1.23--1.32$\times$, 1.45--1.80$\times$ and 1.67--2.20$\times$. Their boundary rates are the rates of the steady-state runs. We replaced them for three reasons: cache use was still rising when arrivals stopped (Section~\ref{sec:e2e}), so they describe a 600-s burst; on the larger models a bracket end could turn on one request (150 requests at most of the 72B points); and the 99\% line was applied to a single run without regard to how few independent blocks such a run holds (below). In those runs FP8's cache was almost full at its highest sustainable rate in seed 0 (99.9\% on Qwen3-32B at 0.25, 95\% on Qwen2.5-72B at 0.125 requests/s; the seed-1 repeats peaked at 68\%), and at the next rate requests were preempted and missed the time-to-first-token target; ours on Qwen2.5-72B at 0.25 requests/s in seed 0 filled its cache, 57 requests were preempted and 149 of 150 met the SLO (seed 1: 80\% peak, none preempted). Llama-3.1-8B's FP8 point at 1.25 requests/s also held with seed 2. As measures of the sustained rate, the steady-state runs supersede these brackets; as measures of burst tolerance (600 s of arrivals from an empty cache) they stand, with the limits stated above and in Section~\ref{sec:e2e}. Of the twelve endpoints that have a steady-state result, five disagree with the 600-s verdict: ours fails 1.6 on Llama-3.1-8B, 0.4 on Qwen3-32B and 0.25 on Qwen2.5-72B, and FP8 is no longer failed at 0.15 on Qwen2.5-72B (uncertain: one seed held, one failed narrowly), four changes against ours; FP8 fails 1.25 on Llama-3.1-8B, one in our favour.

\paragraph{Steady-state runs.} Each boundary rate of the earlier brackets was rerun with arrivals for a warm-up $W$ and a measured window $M$ ($W = 600$ s and $M = 1800$ s on Llama-3.1-8B and Qwen3-32B, 900 s and 2700 s on Qwen2.5-72B; $W$ is about the longest request lifetime at these rates, estimated from the earlier runs). Requests that arrive during the warm-up are served but not counted, and the client waits for every request of the window to finish; failures count as misses. FP8 and ours at a pair of rates run in one job on one H200 SXM, with the engine settings of the earlier runs and four API-server processes. Seed 0 covers every boundary rate; FP8 on Qwen3-32B and Qwen2.5-72B, where one seed gives fewer than 600 measured requests, also has seed 1. The run list, the window lengths and the predictions were written before the runs; Table~\ref{tab:steady} lists every run.

\paragraph{When a window counts as steady, and how that rule changed.} Before the runs we required the KV-cache use and the number of running requests to change by less than 5 points (running requests: 5\%) over $M$ and between the two halves of $M$ (v1). The first run (Qwen3-32B, FP8, 0.25 requests/s, about 73 requests in flight) showed this to be below the fluctuation of a stationary system: with Poisson arrivals the number in flight is itself Poisson-like (standard deviation near $\sqrt{73} \approx 8.5$, 12\%), correlated over a request lifetime (p99 latency 445 s). The same run exposed a bug: the window was aligned with the start of the benchmark client, whose warm-up requests delay the first arrival by 261 s, so it included the ramp; it is now aligned with the first arrival. We then replaced the fixed 5\% by a test against the noise expected for that run (v2), a change of a pre-registered rule after seeing data, decided after this first run; both verdicts are reported for every run. Treating each request as a contribution over its lifetime (1 while in flight for the running count; its input plus the output tokens generated so far for the cache), the stationary autocovariance of both series follows from the measured requests alone, not from the window under test, which gives standard deviations of the drift and of the half difference over $M$. A window is steady if all four $|z| < 2$ and the KV trend, continued one more window past its end, stays below 100\% (headroom rule); a run whose cache is pinned at 100\% with growing waiting and preemptions is saturated, a steady failure. With four correlated tests, a truly steady run is flagged with a probability of roughly 5--15\%, and the model ignores the finite cache and load-dependent step times, so it errs towards ``not steady''. Before the v2 verdicts are used, the noise model is checked on runs far from their limit (peak KV below 85\%, no waiting): the observed spread of 300-s window means after removing the linear trend, against the model's, must lie within a factor of 2 for the median run of each model, or the v2 verdicts are reported as unvalidated and the comparison rests on the run-level shares only. Result: on Qwen2.5-72B three runs qualify (FP8 at 0.125, seed 1, and the post-hoc runs of ours at 0.175 and at 0.2, seed 2), with ratios of 0.88, 0.94 and 0.95 for the running count and 1.05, 0.99 and 0.89 for the cache use (medians 0.94 and 0.99), so the model passes there. On Llama-3.1-8B and Qwen3-32B no design run qualifies; counting the post-hoc runs, Llama-3.1-8B has four, with median ratios of 1.97 and 2.00, at the edge of the band (the observed spread is 2.3$\times$ the model's for ours at 1.0 and 1.1), and Qwen3-32B has one, ours at 0.30, with 0.46 for both series, below the band. Under the rule the v2 verdicts on Qwen3-32B are therefore unvalidated, and there the comparison rests on the run-level shares; with six to nine window means per run, a single run's ratio has a 95\% sampling range of roughly 0.4--1.6 times its true value (chi-squared, after the trend is removed). Table~\ref{tab:steady} gives both verdicts for every run; of the 18 design runs, 13 are not steady under v1, four steady and one saturated, while under v2 11 are steady, six fail the headroom rule and one is saturated; the long-window run is not steady under v1 and steady under v2. As a further check that does not depend on the plateau test, we report the SLO share of the requests arriving in the second half of $M$, with a one-sided 95\% Clopper--Pearson lower bound (Table~\ref{tab:steady}).

\paragraph{Verdicts, extra seeds and brackets (fixed before the remaining results).} A run holds if its SLO share is at least 0.99 and fails otherwise. An endpoint (model, configuration, rate) is held if every run holds, failed if every run fails, and uncertain if they disagree; an endpoint with one run takes that run's verdict, marked as single-run, and an endpoint without a result counts as uncertain. An endpoint gets extra seeds (seed 2, then seed 3 only if still triggered; same $W$, $M$ and job layout) if the two-sided 95\% Clopper--Pearson interval of its pooled share (base-design runs, requests as units; a trigger rule, not the claim) contains 0.99, or if any of its runs is not steady or fails the headroom rule. No run is dropped. Our sustainable rate is our highest held rate below our lowest failed-or-uncertain rate; FP8's limit treats FP8's uncertain endpoints as held, which is conservative for our claim; the ratio range is [our held rate / FP8's first rate not held, our first rate not held / FP8's held rate]. A model counts as showing a gain only if our lower endpoint is held and FP8's upper endpoint is failed, so that the lower end of the ratio range exceeds 1. The rule added two seeds to design endpoints. The first was for FP8 on Qwen3-32B at 0.275 requests/s, whose seed-1 run failed the headroom rule. Its seed 2 met the SLO for 129 of 475 requests (27.2\%; headroom rule failed); with both runs pooled the interval excludes 0.99, so there is no seed 3. We first applied only the interval clause to this endpoint and added the seed when we found the error; the rule itself did not change. The second was for ours on Qwen3-32B at 0.4, whose seed-0 run failed the headroom rule; seed 2 met the SLO for 160 of 708 requests (22.6\%; headroom rule failed), and the pooled interval excludes 0.99. The rule also called for seed 2 of FP8 on Qwen2.5-72B at 0.15 (pooled interval containing 0.99) and seed 1 of ours at 0.25 (headroom rule failed). Neither could change a verdict, because the FP8 endpoint is already uncertain and an uncertain endpoint of ours counts as failed, so neither was run; they are the two design seeds counted below.

\paragraph{What ``sustainable'' means here.} This is an operational definition, in the spirit of fixed-percentile latency targets such as MLPerf's, but at a much smaller scale: MLPerf requires far more queries and longer runs per result, so each of our verdicts is less certain. Requests in one run share a queue and a cache, so they are not independent. A circular block bootstrap over arrival order gives zero-width intervals at a share of 100\%, and with the effective sample size of a run (the number of non-overlapping blocks of one p99 request latency, 4--5 per steady-state run so far) the one-sided 95\% lower bound of a run with every request met is only 0.47--0.55 (Table~\ref{tab:steady}). We had first defined the effective sample size from the design effect of a block bootstrap with blocks of two p99 latencies; when one such block spans a whole run its variance is zero and that rule returned the full request count, the least conservative value exactly where the data are weakest. We now use the smaller of the two estimates (both are in the supplementary result files). A single run therefore cannot show 99\%; the comparison rests on applying the same definition, hardware and statistics to both configurations, and on agreement across seeds. Table~\ref{tab:steady} gives every run's share and verdicts, the lower bounds with the nominal and the effective sample size, and block-bootstrap intervals for blocks of 1, 2 and 4 p99 latencies; these are supplementary columns, not the verdict. The rate steps between the design rates are 3--4\% of the rate on Llama-3.1-8B, 10--12.5\% on the larger models and 20\% for FP8 on Qwen2.5-72B, so a bracket cannot resolve smaller changes.

\paragraph{Results reported as they came.} Before the runs we predicted that FP8 would stay below 99\% at its earlier failed end on all three models. On Qwen3-32B it did in all three seeds at 0.275 requests/s (122 of 495, 388 of 489 and 129 of 475; cache at 100\%, headroom rule failed): past the limit the share varies widely between seeds. On Qwen2.5-72B it did not: at 0.15 requests/s seed 1 met the SLO for 400 of 402 requests (99.5\%; v2 steady; peak cache 100\%, mean 73\%; 29 preemptions; nominal lower bound 0.984, effective-$n$ bound 0.54). By the rules above, FP8's endpoint at 0.15 is then held or uncertain, and either way it counts as held, so FP8's limit on Qwen2.5-72B is at least 0.15 requests/s and its earlier bracket $[0.125, 0.15)$ does not stand. The runs planned before the results include no FP8 rate above 0.15 on this model; the FP8 rate added later (0.175) is post-hoc. Our design rates failed on all three models: on Llama-3.1-8B at 1.6 and 1.65 (1.2\% and 0\% of requests met the SLO), on Qwen3-32B at 0.4 in two seeds (20.0\% and 22.6\%) and at 0.45 (0.4\%), and on Qwen2.5-72B at 0.25 and 0.275 (28.3\% and 0.8\%). We report these results whether or not they favour our configuration.

\paragraph{Results at the design rates.} No model met the gain criterion, and on none did the design rates give a valid verdict (Table~\ref{tab:capacity}). On Llama-3.1-8B both configurations failed at both of their design rates: FP8 at 1.25 and 1.3 requests/s (none of 2,232 and none of 2,319 measured requests met the SLO), ours at 1.6 and 1.65 (33 of 2,860 and none of 2,960), all with the cache full or nearly so (mean use 96--100\%); both sustainable rates lie below the design, so neither bracket was measured. On Qwen3-32B FP8 holds 0.25 (seeds 0 and 1: 447 of 449 and 443 of 443) and fails 0.275 in all three seeds (24.7\%, 79.4\% and 27.2\% of requests met the SLO), a bracket of $[0.25, 0.275)$; ours fails 0.4 in two seeds (20.0\% and 22.6\%) and 0.45 (0.4\%), so all our design rates lay above our sustainable rate. On Qwen2.5-72B FP8 holds 0.125 in both seeds (338 of 338 and 333 of 333); at 0.15 seed 1 holds (400 of 402) and seed 0 fails narrowly (394 of 400, 98.5\%), so that endpoint is uncertain and counts as held for FP8; no higher FP8 rate was in the design, so its upper end was not measured. Ours fails 0.25 (190 of 672) and 0.275 (6 of 733). FP8's first seed-0 runs on this model were lost (0.15: the server did not start in the time allowed) or anomalous (0.125: decode 3--5$\times$ slower from the first request), and our first run at 0.25 was cut by the job's time limit; all three were rerun, on another node for FP8 (``Lost and anomalous runs''). The pre-set test thus neither shows nor rules out a gain; it shows that the design rates, placed at the boundaries of our 600-s scans, were too high for ours on every model.

\paragraph{Post-hoc runs at lower rates.} After these results we added runs below the design rates for both configurations, with rates and predictions fixed before they ran: on Llama-3.1-8B FP8 at 0.75 and 0.85 (then 1.0 and 1.1, added after these two held) and ours at 1.0 and 1.1 (then 1.25 and 1.4, added after 1.1 held); on Qwen3-32B ours at 0.30, 0.33 and 0.36; on Qwen2.5-72B FP8 at 0.175 and ours at 0.175, 0.2 and 0.225. They are labelled post-hoc throughout, have one seed each unless the extra-seed rule adds one, and do not change the pre-set verdicts; we extended the extra-seed rule to post-hoc runs after seeing the result at 0.36 below (``Extra seeds for post-hoc runs''). On Qwen3-32B ours holds 0.30, 0.33 and 0.36 (542 of 542, 586 of 586 and 643 of 643 requests; mean cache use 55\%, 69\% and 80\%; no preemptions). At 0.36 cache use was still rising at the end of the window, so that run fails the headroom test; by the extended rule it got a second seed, which held as well (633 of 633 requests; mean cache use 77\%; no preemptions) with a steady window, so 0.36 is held on two seeds and lies near our limit, as the Little's-law estimate made before the runs (0.35--0.37 requests/s) expected. Against FP8's three failed seeds at 0.275, ours thus sustains at least $0.36/0.275 = 1.31\times$ FP8's lowest failed rate and $0.36/0.25 = 1.44\times$ its highest held one ($1.20\times$ and $1.32\times$ at 0.33, held on one seed with a steady window); all of these are post-hoc.

On Qwen2.5-72B ours holds 0.175 and 0.2 (470 of 470 and 542 of 542 requests; mean cache use 51\% and 66\%; no preemptions) and fails 0.225 (574 of 600, 95.7\%; 187 preemptions), a bracket of $[0.2, 0.225)$; the Little's-law estimate made before the runs had put our limit above 0.225, too high here. The run at 0.2 is not steady under v2 (cache use fell during the window), so the rule added a second seed, which held as well (521 of 521 requests; mean cache use 64\%; no preemptions) with a steady window, so 0.2 is held on two seeds. FP8 fails 0.175 by a wide margin (38 of 470, 8.1\%; cache full, 2,151 preemptions), so its sustainable rate on this model lies between 0.15 (uncertain) and 0.175. Ours thus sustains $0.2/0.175 = 1.14\times$ FP8's lowest failed rate and $0.2/0.15 = 1.33\times$ its highest rate counted as held ($1.60\times$ against 0.125, held in both seeds); post-hoc, with two seeds of ours at 0.2.

On Llama-3.1-8B the first post-hoc rates were too low for both configurations. FP8 holds 0.75 and 0.85 (1,330 of 1,330 and 1,513 of 1,513 requests; mean cache use 26\% and 35\%; no preemptions) and ours holds 1.1 (1,971 of 1,972; mean cache use 35\%). We had chosen these rates with the same Little's-law estimate, which takes the decode step at the batch that fills the cache; on this model the step grows strongly with the batch, and at these rates both configurations settled at about a third of their cache, so the estimate gave a lower bound rather than the limit. We then added FP8 at 1.0 and 1.1 and ours at 1.25 and 1.4, again with predictions fixed before they ran. Both FP8 runs at 0.75 and 0.85 are not steady under v2 (cache use fell during the window); the extra seeds the rule asks for were not run (``Extra seeds for post-hoc runs''). FP8 holds 1.0 (1,787 of 1,787; mean cache use 72\%) and fails 1.1 (397 of 1,972, 20.1\%; cache full, 8,526 preemptions), a bracket of $[1.0, 1.1)$. Ours holds 1.0 and 1.1 (1,787 of 1,787 and 1,971 of 1,972) and fails 1.4 (777 of 2,496, 31.1\%); at 1.25 its first seed held (2,232 of 2,232) with cache use still rising, and the extra seed the rule then added failed (1,983 of 2,236, 88.7\%), so 1.25 is uncertain and our sustainable rate is 1.1, a bracket of $[1.1, 1.25)$. The ratio range is $[1.1/1.1, 1.25/1.0] = [1.00, 1.25]$: on Llama-3.1-8B the post-hoc runs neither show nor rule out a gain, and bound it by $1.25\times$; post-hoc, one seed per rate except 1.25.

\paragraph{Extra seeds for post-hoc runs (a rule added after a result).} The post-hoc runs were planned with one seed each. Our run at 0.36 requests/s on Qwen3-32B held (643 of 643 requests) but failed the headroom test, the case in which the design rule adds a seed. The post-hoc plan did not say whether that rule applies to post-hoc runs; we decided after seeing this result that it does, for every post-hoc run from then on (seed 2, then seed 3 only if still triggered; an endpoint is held if all its runs hold, failed if none does, and uncertain otherwise, which counts as not held for ours). The second seed at 0.36 held (633 of 633) with a steady window and the pooled interval excludes 0.99, so the rule added no third seed. We ran further seeds only where the result could change a verdict or a ratio we report; nine runs were not run: two design and five post-hoc extra seeds, and FP8 on Qwen2.5-72B at 0.2 and 0.275 after it failed at 0.175. None of them could change a verdict or a ratio, except the second seed of ours at 1.0 on Llama-3.1-8B, which could have lowered our bracket end only by failing where the first seed met the SLO for 1,787 of 1,787 requests at 23\% mean cache use. On Qwen2.5-72B our run at 0.2 was not steady under v2, and its second seed held (521 of 521 requests) with a steady window.

\paragraph{Lost and anomalous runs.} Two runs of the design were lost and one is labelled anomalous; all three are reported, and their files are kept. (i) FP8 at 0.15 requests/s on Qwen2.5-72B, seed 0: the server was still loading the 72B weights from network storage when the 10-minute start-up limit expired (7 of 16 shards loaded), so the run has no result; later jobs wait at most 12 minutes, give up early when the server log has stopped growing for 5 minutes, and retry the start once. (ii) Ours at 0.25 on Qwen2.5-72B, seed 0: it shared a job with FP8 at 0.125, whose overloaded run drained until near the job's time limit, so our run was cut; by a rule fixed before the runs, a configuration lost this way is rerun alone, and the rerun failed (190 of 672). (iii) FP8 at 0.125 on Qwen2.5-72B, seed 0 (that same overloaded run): 0 of 333 requests met the SLO, against 338 of 338 in seed 1, because decode was 3--5$\times$ slower from the first requests on: mean inter-token latency per 25 requests by start time 104, 131, 145, 233~ms, against 33, 37, 37~ms in seed 1, and a mean time per output token of 184.6~ms against 37.0 (seed 1), 37.6 (FP8 at 0.15) and 35.7~ms (FP8 on Qwen3-32B). The step was already 3$\times$ slower with fewer than 25 requests running, so load does not explain it; the server started with the same engine settings, decode kernel, CUDA graphs and compilation cache as seed 1, and a slow node is the likely cause, which we could not check afterwards. No rule covered such a run, so the decision to label it anomalous and rerun it on another node was taken after seeing the data; it does not change the 72B verdict either way (with the run kept, FP8's endpoint at 0.125 is uncertain, which counts as held for FP8). From then on we fixed a check for later runs: a run whose mean inter-token latency over its first 25 requests exceeds twice that of the same configuration's other seeds at the same rate is labelled anomalous, reported and rerun once on another node. The two lost or anomalous FP8 runs were rerun on other nodes: 0.125 held (333 of 333; mean inter-token latency 37.1~ms, as in seed 1) and 0.15 failed narrowly (394 of 400).

\paragraph{A longer window.} One run of FP8 on Qwen3-32B at 0.25 requests/s (seed 2) used $M = 3600$ s instead of 1800 s, with the same $W$, to see whether the 1800-s verdicts depend on the window. It met the SLO for 827 of 877 requests (94.3\%, interval [0.928, 0.958]) and for 412 of 457 in its second half (90.2\%); it is steady under v2 (all $|z| < 1$; on Qwen3-32B the v2 verdicts are unvalidated, see above), with mean cache use 81\% and a peak of 100\%, and the mean number of running requests (74.6) matches the model's prediction from the measured requests (75.2) within 1\%. Its first 1800 s alone give 98.8\% (415 of 420), also below the line. So with seed 2 FP8 does not hold 0.25 even in an 1800-s window, and the longer window makes the verdict stricter still. The run is reported in Table~\ref{tab:steady} and is not pooled into the design endpoint. A post-hoc run of ours at 0.33 with the same layout (seed 2) met the SLO for all 1,166 measured requests and for all 582 in its second half (one-sided 95\% lower bound 0.995); it is steady under v2 (all $|z| < 1.5$), with mean cache use 69\%, a peak of 93\%, no waiting requests and no preemptions. So the longer window made FP8's verdict at 0.25 stricter but left ours at 0.33 unchanged; with one run each, this does not show that the window affects the two configurations differently in general.

\paragraph{Page size.} Ours serves with 64-token pages and FP8 with vLLM's default of 16. In \texttt{vllm bench forward} (FP8 KV, FlashInfer with XQA decode, one job per model, H200 SXM), the default and explicit 16-token pages give the same step time within 0.02\%, so the FP8 serving runs used 16-token pages. With 64-token pages FP8's step is 0.1--1.2\% shorter on Qwen3-32B and 0.9\% shorter to 0.3\% longer on Qwen2.5-72B, at batch $\times$ context $64 \times 5$K, $80 \times 4$K, $32 \times 8$K and $8 \times 8$K (32B) and $48 \times 5$K, $64 \times 4$K, $32 \times 8$K and $8 \times 8$K (72B); the spread over repeated samples is at most 0.12\%. Page size changes FP8's step time by at most 1.2\%. Larger pages waste more of each request's last page: on average 31.5 tokens with 64-token pages against 7.5 with 16, out of about 5,100 tokens of context per request in flight (1,024 input tokens plus half of an 8,240-token output), about 0.6\% against 0.15\% of the cache; no prefix is shared in these workloads. So the 64-token pages cost ours about 0.5\% of its capacity.

\paragraph{The throughput split under preemption.} For each run, the output token rate over $M$ should equal the mean number of running requests divided by the step time. With the mean inter-token latency seen by the client as the step, this holds within 0.3\% in every run without preemptions (e.g.\ FP8 on Qwen3-32B at 0.25: 1990.8 against 1991.6 tokens/s) and fails with them (FP8 on Qwen3-32B at 0.275: $-$1.5\%, $-$8\% and $-$10\%; ours at 0.4: $-$10\%; ours on Qwen2.5-72B at 0.275: $-$14\%). Output tokens are counted the same way on both sides and running requests are sampled every 2 s, so the definitions agree; a preempted request leaves the running batch, but its next client gap contains its wait. With gaps longer than 1 s removed (about 20 steps; the cut was fixed before this check), the check holds within 1\% in 11 of the 12 overloaded or near-limit runs, the removed gaps making up at most 11\% of the gap time; it fails on ours on Qwen2.5-72B at 0.275 (+8.7\%, with 21\% of the gap time removed). The saved server counters cannot test whether recomputation after preemption explains this run: they count each request's prompt once and do not record resumed requests.

\begin{table}[tbp]
\centering
\caption{Steady-state runs, one row per run (H200 SXM, settings of Table~\ref{tab:capacity}); $^{\dagger}$post-hoc rate (added after the design results). Met: measured requests meeting the SLO. Second half: requests arriving in the second half of $M$ that meet it, with the one-sided 95\% Clopper--Pearson lower bound. v1 and v2: steadiness under the criterion fixed before the runs and under the revised one (st: steady, hf: headroom rule fails, ns: not steady); $|z|$: the largest of v2's four test statistics. KV: mean and maximum cache use over $M$. p99: 99th-percentile request latency (arrival to last token). Lower bounds on the share: Clopper--Pearson with the nominal $n$ and with the effective $n_{\mathrm{eff}}$, and the lower ends of block-bootstrap intervals with blocks of $k = 1, 2, 4$ p99 latencies. Supplementary columns; the verdicts of Table~\ref{tab:capacity} use only the share.}
\label{tab:steady}
\scriptsize\setlength{\tabcolsep}{2.2pt}
\resizebox{\linewidth}{!}{\begin{tabular}{@{}lllrrllrlrrrrl@{}}
\toprule
Model, config. & Rate & Seed (job) & Met & Share & Second half & v1 / v2 & $|z|$ & KV (\%) & p99 (s) & CP nom. & $n_{\mathrm{eff}}$ & CP eff. & Block $k = 1, 2, 4$ \\
\midrule
Llama-3.1-8B, FP8 & 0.75$^{\dagger}$ & 0 (phLf75) & 1330 / 1330 & 1.000 & 644 / 644 (0.995) & ns / ns & 2.48 & 26 / 34  & 175 & 0.998 & 10 & 0.741 & 1.000, 1.000, 1.000 \\
Llama-3.1-8B, FP8 & 0.85$^{\dagger}$ & 0 (phLf85) & 1513 / 1513 & 1.000 & 749 / 749 (0.996) & ns / ns & 4.56 & 35 / 46  & 225 & 0.998 & 8 & 0.688 & 1.000, 1.000, 1.000 \\
Llama-3.1-8B, FP8 & 1.0$^{\dagger}$ & 0 (phLf10) & 1787 / 1787 & 1.000 & 899 / 899 (0.997) & st / st & 1.16 & 72 / 85  & 384 & 0.998 & 4 & 0.473 & 1.000, 1.000, 1.000 \\
Llama-3.1-8B, FP8 & 1.1$^{\dagger}$ & 0 (phLf11) & 397 / 1972 & 0.201 & 2 / 1012 (0.000) & sa / sa & 1.25 & 98 / 100 & 846 & 0.187 & 2 & 0.000 & 0.002, 0.141, 0.201 \\
Llama-3.1-8B, FP8 & 1.25 & 0 (L1) & 0 / 2232 & 0.000 & 0 / 1137 (0.000) & sa / sa & 0.24 & 100 / 100 & 1280 & 0.000 & 1 & 0.000 & 0.000, 0.000, 0.000 \\
Llama-3.1-8B, FP8 & 1.3 & 0 (L2) & 0 / 2319 & 0.000 & 0 / 1179 (0.000) & st / st & 0.26 & 100 / 100 & 1346 & 0.000 & 1 & 0.000 & 0.000, 0.000, 0.000 \\
Llama-3.1-8B, ours & 1.0$^{\dagger}$ & 0 (phLr10) & 1787 / 1787 & 1.000 & 899 / 899 (0.997) & ns / ns & 3.39 & 23 / 32  & 210 & 0.998 & 8 & 0.688 & 1.000, 1.000, 1.000 \\
Llama-3.1-8B, ours & 1.1$^{\dagger}$ & 0 (phLr11) & 1971 / 1972 & 0.999 & 1011 / 1012 (0.995) & ns / st & 0.91 & 35 / 45  & 281 & 0.998 & 6 & 0.606 & 0.998, 0.998, 0.999 \\
Llama-3.1-8B, ours & 1.25$^{\dagger}$ & 0 (phLr125) & 2232 / 2232 & 1.000 & 1137 / 1137 (0.997) & ns / hf & 8.15 & 66 / 89  & 543 & 0.999 & 3 & 0.368 & 1.000, 1.000, 1.000 \\
Llama-3.1-8B, ours & 1.25$^{\dagger}$ & 2 (phLr125s2) & 1983 / 2236 & 0.887 & 861 / 1114 (0.751) & ns / hf & 10.92 & 74 / 100 & 633 & 0.875 & 2 & 0.159 & 0.702, 0.774, 0.887 \\
Llama-3.1-8B, ours & 1.4$^{\dagger}$ & 0 (phLr14) & 777 / 2496 & 0.311 & 0 / 1256 (0.000) & ns / hf & 3.55 & 92 / 100 & 803 & 0.296 & 2 & 0.003 & 0.000, 0.203, 0.311 \\
Llama-3.1-8B, ours & 1.6 & 0 (L1r) & 33 / 2860 & 0.012 & 0 / 1467 (0.000) & st / st & 0.09 & 96 / 100 & 1147 & 0.008 & 1 & 0.000 & 0.000, 0.012, 0.012 \\
Llama-3.1-8B, ours & 1.65 & 0 (L2r) & 0 / 2960 & 0.000 & 0 / 1527 (0.000) & st / st & 0.21 & 96 / 100 & 1172 & 0.000 & 1 & 0.000 & 0.000, 0.000, 0.000 \\
Qwen3-32B, FP8 & 0.25 & 0 (Q01) & 447 / 449 & 0.996 & 214 / 214 (0.986) & ns / st & 1.08 & 82 / 100 & 459 & 0.986 & 3 & 0.364 & 0.987, 0.991, 0.996 \\
Qwen3-32B, FP8 & 0.25 & 1 (Q11) & 443 / 443 & 1.000 & 234 / 234 (0.987) & ns / st & 0.96 & 78 / 97  & 445 & 0.993 & 4 & 0.473 & 1.000, 1.000, 1.000 \\
Qwen3-32B, FP8 & 0.275 & 0 (Q02r) & 122 / 495 & 0.246 & 8 / 229 (0.017) & ns / hf & 0.53 & 98 / 100 & 724 & 0.215 & 2 & 0.001 & 0.042, 0.123, 0.246 \\
Qwen3-32B, FP8 & 0.275 & 1 (Q12) & 388 / 489 & 0.793 & 171 / 260 (0.606) & ns / hf & 1.15 & 93 / 100 & 512 & 0.761 & 3 & 0.209 & 0.665, 0.681, 0.793 \\
Qwen3-32B, FP8 & 0.275 & 2 (Q22) & 129 / 475 & 0.272 & 44 / 233 (0.148) & st / hf & 0.19 & 98 / 100 & 675 & 0.238 & 2 & 0.001 & 0.166, 0.196, 0.272 \\
Qwen3-32B, ours & 0.3$^{\dagger}$ & 0 (phQ30) & 542 / 542 & 1.000 & 269 / 269 (0.989) & ns / st & 1.47 & 55 / 65  & 424 & 0.994 & 4 & 0.473 & 1.000, 1.000, 1.000 \\
Qwen3-32B, ours & 0.33$^{\dagger}$ & 0 (phQ33) & 586 / 586 & 1.000 & 301 / 301 (0.990) & ns / st & 0.62 & 69 / 92  & 479 & 0.995 & 3 & 0.368 & 1.000, 1.000, 1.000 \\
Qwen3-32B, ours & 0.36$^{\dagger}$ & 0 (phQ36) & 643 / 643 & 1.000 & 320 / 320 (0.991) & ns / hf & 1.44 & 80 / 94  & 518 & 0.995 & 3 & 0.368 & 1.000, 1.000, 1.000 \\
Qwen3-32B, ours & 0.36$^{\dagger}$ & 2 (phQ36s2) & 633 / 633 & 1.000 & 321 / 321 (0.991) & ns / st & 0.98 & 77 / 95  & 508 & 0.995 & 3 & 0.368 & 1.000, 1.000, 1.000 \\
Qwen3-32B, ours & 0.4 & 0 (Q01) & 142 / 711 & 0.200 & 0 / 365 (0.000) & ns / hf & 0.94 & 98 / 100 & 1060 & 0.175 & 1 & 0.000 & 0.000, 0.200, 0.200 \\
Qwen3-32B, ours & 0.4 & 2 (Q21) & 160 / 708 & 0.226 & 0 / 367 (0.000) & ns / hf & 0.54 & 99 / 100 & 1026 & 0.200 & 1 & 0.000 & 0.010, 0.226, 0.226 \\
Qwen3-32B, ours & 0.45 & 0 (Q02) & 3 / 807 & 0.004 & 0 / 421 (0.000) & ns / st & 0.36 & 100 / 100 & 1576 & 0.001 & 1 & 0.000 & 0.000, 0.004, 0.004 \\
Qwen2.5-72B, FP8 & 0.125 & 0 (X0fa) & 333 / 333 & 1.000 & 171 / 171 (0.983) & ns / st & 0.74 & 64 / 91  & 459 & 0.991 & 5 & 0.549 & 1.000, 1.000, 1.000 \\
Qwen2.5-72B, FP8 & 0.125 & 1 (X11) & 338 / 338 & 1.000 & 172 / 172 (0.983) & ns / st & 0.53 & 59 / 83  & 459 & 0.991 & 5 & 0.549 & 1.000, 1.000, 1.000 \\
Qwen2.5-72B, FP8 & 0.15 & 0 (X0fb) & 394 / 400 & 0.985 & 187 / 192 (0.946) & ns / st & 0.68 & 74 / 100 & 465 & 0.971 & 5 & 0.531 & 0.955, 0.958, 0.970 \\
Qwen2.5-72B, FP8 & 0.15 & 1 (X12) & 400 / 402 & 0.995 & 197 / 199 (0.969) & ns / st & 1.09 & 73 / 100 & 463 & 0.984 & 5 & 0.543 & 0.985, 0.985, 0.990 \\
Qwen2.5-72B, FP8 & 0.175$^{\dagger}$ & 0 (phF175) & 38 / 470 & 0.081 & 0 / 219 (0.000) & ns / hf & 0.25 & 99 / 100 & 961 & 0.061 & 2 & 0.000 & 0.000, 0.006, 0.081 \\
Qwen2.5-72B, ours & 0.175$^{\dagger}$ & 0 (phX17) & 470 / 470 & 1.000 & 219 / 219 (0.986) & ns / st & 1.78 & 51 / 78  & 454 & 0.994 & 5 & 0.549 & 1.000, 1.000, 1.000 \\
Qwen2.5-72B, ours & 0.2$^{\dagger}$ & 0 (phX20) & 542 / 542 & 1.000 & 269 / 269 (0.989) & ns / ns & 2.46 & 66 / 87  & 528 & 0.994 & 5 & 0.549 & 1.000, 1.000, 1.000 \\
Qwen2.5-72B, ours & 0.2$^{\dagger}$ & 2 (phX20s2) & 521 / 521 & 1.000 & 260 / 260 (0.989) & ns / st & 0.35 & 64 / 83  & 499 & 0.994 & 5 & 0.549 & 1.000, 1.000, 1.000 \\
Qwen2.5-72B, ours & 0.225$^{\dagger}$ & 0 (phX22) & 574 / 600 & 0.957 & 278 / 304 (0.883) & ns / st & 0.59 & 84 / 100 & 598 & 0.940 & 4 & 0.427 & 0.872, 0.900, 0.915 \\
Qwen2.5-72B, ours & 0.25 & 0 (X01r) & 190 / 672 & 0.283 & 0 / 335 (0.000) & ns / hf & 0.52 & 98 / 100 & 1063 & 0.254 & 2 & 0.002 & 0.042, 0.107, 0.283 \\
Qwen2.5-72B, ours & 0.275 & 0 (X02) & 6 / 733 & 0.008 & 0 / 371 (0.000) & ns / st & 0.37 & 100 / 100 & 1681 & 0.004 & 1 & 0.000 & 0.000, 0.008, 0.008 \\
\midrule
\multicolumn{14}{@{}l@{}}{Long window (reported separately; not used in Table~\ref{tab:capacity}):} \\
Qwen3-32B, FP8 & 0.25 & 2 (Qlw, $M = 3600$ s) & 827 / 877 & 0.943 & 412 / 457 (0.876) & ns / st & 0.97 & 81 / 100 & 485 & 0.928 & 7 & 0.577 & 0.854, 0.851, 0.886 \\
Qwen3-32B, ours & 0.33$^{\dagger}$ & 2 (phQlw, $M = 3600$ s) & 1166 / 1166 & 1.000 & 582 / 582 (0.995) & ns / st & 1.42 & 69 / 93  & 496 & 0.997 & 7 & 0.652 & 1.000, 1.000, 1.000 \\
\midrule
\multicolumn{14}{@{}p{0.95\linewidth}@{}}{sa: saturated.} \\
\bottomrule
\end{tabular}}
\end{table}

\paragraph{End-to-end latency and throughput.} End to end (\texttt{vllm bench latency}, recipe vs FlashInfer FP8, two consecutive jobs on one H200 SXM node): time per output token 5.30 vs 5.08 ms at $1 \times 16$K (0.96$\times$), 5.55 vs 5.72 at $1 \times 32$K (1.03$\times$), 6.95 vs 7.77 at $8 \times 16$K (1.12$\times$), 8.60 vs 10.94 at $8 \times 32$K (1.27$\times$) and 12.77 vs 13.94 at $32 \times 16$K (1.09$\times$). Offline throughput (8K and 16K input, 512 output tokens, 256 requests) is 1.06--1.10$\times$ in the same jobs. The KV cache holds 2,867,200 tokens against 1,761,600 for FP8 at the same memory setting (1.63$\times$). The bytes per token alone would give 1.65$\times$ (39,808 against 65,536 bytes with layer 0 in BF16); the rest is memory: after its profiling run vLLM gave our cache 114.1 GB against 115.5 GB for FP8 (1.1\% less). Decode-led throughput (16K in, 2,048 out, 256 requests) is 1.17$\times$ against FlashInfer FP8 and 1.40$\times$ against the default FlashAttention-3 FP8 path; these three runs are separate jobs on H200 SXM, and the FP8 runs predate a plugin change that does not affect them.

\paragraph{A batch FP8's cache cannot hold.} At $64 \times 32$K (\texttt{vllm bench latency}, both configurations in one job, H200 SXM), where FP8's cache cannot hold all 64 sequences (2.1M tokens against 1.76M) and ours can, decode time per output token is 38.7 vs 52.2 ms (1.35$\times$), and FP8's figure includes the time its sequences wait for cache space, so this is a capacity effect rather than decode speed; the whole request with 256 output tokens still takes 86.8 vs 85.0 s (2\% longer), because prefill of the 2.1M prompt tokens is 7\% slower (77.0 vs 71.7 s).

\paragraph{Runs per rate.} The rate scans of Table~\ref{tab:serve} have one run per rate at 0.75 and 1.0 requests/s, two (FP8) and three (ours) at 0.25 and 0.5, two at 1.25 (three with the seed-2 rerun), and four (FP8) and five (ours) at 1.35 and 1.5, including the runs with four API-server processes.

\paragraph{K on FP8 tensor cores (C1) under load.} These runs use the earlier 600-s design and are compared with the 600-s brackets above, not with Table~\ref{tab:capacity}. Faster steps also raise the request rate one GPU sustains, although the cache holds the same number of tokens: requests finish sooner, so fewer are resident at once. Under the reasoning-like stream (Section~\ref{sec:c1}; Llama-3.1-8B, four API-server processes, seed 0), at 1.65 requests/s our kernel without C1 meets the SLO for 94.2\% of requests (92.4\% in a second run; 96\% peak cache use, 30 requests waiting) and with it for 100\% (two runs, 90\% peak cache use, none waiting); with it, 1.7 requests/s fails (95.9\%). The sustainable rate thus moves from $[1.6, 1.65)$ to $[1.65, 1.7)$, 1.27--1.36$\times$ FP8's $[1.25, 1.3)$ in the same design. On the larger models, Qwen3-32B stays at $[0.4, 0.45)$ with C1 (at 0.45 requests/s 93.0\% of requests meet the SLO with it, 87.4\% without; 116 against 194 preemptions). On Qwen2.5-72B the bracket moves down: with 150 requests per point the 99\% line allows one miss, and with C1 seed 0 misses it at 0.25 requests/s by one request (148 of 150, 98.7\%, against 149 without C1) while seed 1 meets it (all 150), so by our rule its bracket lies below 0.25 requests/s. On the larger models C1 lowers peak cache use at the rates that hold (87--89\% against 91--92\%). One request decides the 72B bracket, and the rate steps of our brackets on the larger models are 10--12.5\% of the rate (20\% for FP8 on Qwen2.5-72B; 3--4\% on Llama-3.1-8B), so these brackets cannot resolve a change of a few percent.

\FloatBarrier
\section{\anonelse{Findings about vLLM's KV-cache paths}{Findings reported to vLLM}}
\label{sec:vllm}

\anonelse{We list four findings about vLLM's KV-cache paths, each with a minimal reproducer (Table~\ref{tab:vllm}).}{We reported our findings to vLLM, each with a minimal reproducer (Table~\ref{tab:vllm}). At the time of submission, all of the issues and PRs below are open; the authors of two PRs have acted on our input, and a third PR cites our measurements.}

\ifanon
\begin{table}[!htb]
\centering
\caption{Findings about vLLM's KV-cache paths, each with a minimal reproducer.}
\label{tab:vllm}
\small
\begin{tabularx}{\linewidth}{@{}P{3.0cm}L@{}}
\toprule
Where & Finding \\
\midrule
A vLLM issue & UltraQuant's scale constant $c = 0.156$ is fit for a continuous scale but applied with a power-of-two snap. The UltraQuant paper names 0.195 the power-of-two-optimal value but measured it 4.4 points less accurate in one run~\citep[its Table 5;][]{chakrabarti2026ultraquant}; we find that 0.20 lowers KL offline and the perplexity increase over BF16 inside vLLM (Llama-3.1-8B, Mistral-7B-v0.3, Qwen3-8B-Base), at the same bytes \\
\addlinespace
A vLLM issue and a PR & FP8 KV silently breaks Qwen2-family reasoning models at layer 0; skipping layer 0 restores accuracy, but crashes FlashAttention-3 for some models (\#65821, \citealp{vllm65821}; proposed fix in open PR \#65891, \citealp{vllm65891}) \\
\addlinespace
Two vLLM PRs & TurboQuant and UltraQuant skip cached KV when a request with cached context shares a step with a longer new prompt, giving wrong answers with no error; the first PR fixed only TurboQuant, the second fixes both, and we confirmed on H200 that it fixes UltraQuant end to end \\
\addlinespace
A vLLM RFC and a PR & A prompt-logprob check, of the kind the RFC proposes for numerics PRs, catches the layer-0 FP8 failure but not the cached-KV bug above, which shows only when a cached prefix shares a step with a longer new prompt; a targeted request of that shape catches it \\
\bottomrule
\end{tabularx}
\end{table}
\else
\begin{table}[!htb]
\centering
\caption{Findings reported to vLLM, each with a minimal reproducer (status at the time of submission).}
\label{tab:vllm}
\small
\begin{tabularx}{\linewidth}{@{}P{3.9cm}LP{3.8cm}@{}}
\toprule
Where & Finding & Status \\
\midrule
\href{https://github.com/vllm-project/vllm/issues/67176}{Issue \#67176}~\citep{vllm67176} & UltraQuant's scale constant $c = 0.156$ is fit for a continuous scale but applied with a power-of-two snap. The UltraQuant paper names 0.195 the power-of-two-optimal value but measured it 4.4 points less accurate in one run~\citep[its Table 5;][]{chakrabarti2026ultraquant}; we find that 0.20 cuts KL by 40--56\% offline and the perplexity increase over BF16 by 66--79\% inside vLLM (Llama-3.1-8B, Mistral-7B-v0.3, Qwen3-8B-Base), at the same bytes & Opened by us; we posted two fixes (a constant of 0.20, or the better of two power-of-two exponents per group) with unit tests and perplexity on three models, and asked the maintainers to choose \\
\addlinespace
\anonelse{An existing issue and a PR (links withheld for review)}{\href{https://github.com/vllm-project/vllm/issues/41343}{Issue \#41343} \citep{vllm41343}, \href{https://github.com/vllm-project/vllm/pull/63081}{PR \#63081} \citep{vllm63081}} & FP8 KV silently breaks Qwen2-family reasoning models at layer 0; skipping layer 0 restores accuracy, but crashes FlashAttention-3 for some models (\#65821, \citealp{vllm65821}; proposed fix in open PR \#65891, \citealp{vllm65891}) & \anonelse{We posted text-model data in the issue (an existing issue about Qwen-VL with \texttt{fp8\_e5m2}) and in the PR; the PR author added a note on the crash to its docs}{We posted text-model data in \#41343 (an existing issue about Qwen-VL with \texttt{fp8\_e5m2}) and \#63081; the \#63081 author added a note on the crash to its docs} \\
\addlinespace
\anonelse{A PR and a follow-up PR (links withheld for review)}{\href{https://github.com/vllm-project/vllm/pull/46461}{PR \#46461} \citep{vllm46461} $\rightarrow$ \href{https://github.com/vllm-project/vllm/pull/67789}{PR \#67789} \citep{vllm67789}} & TurboQuant and UltraQuant skip cached KV when a request with cached context shares a step with a longer new prompt, giving wrong answers with no error; \anonelse{the first PR}{\#46461} fixed only TurboQuant & We reported that UltraQuant has the same bug, with a reproducer; \anonelse{the follow-up PR}{\#67789} fixes both and shows that default chunked prefill also triggers it; we confirmed on H200 that it fixes UltraQuant end to end (0/4 $\rightarrow$ 4/4 correct on both triggers); its author added our results to the PR description; we then opened a draft PR with a regression test for both triggers that depends on it, and another contributor reproduced its results on an RTX 5090 \\
\addlinespace
\anonelse{An RFC and a PR (links withheld for review)}{\href{https://github.com/vllm-project/vllm/issues/61698}{RFC \#61698} \citep{vllm61698}, \href{https://github.com/vllm-project/vllm/pull/69042}{PR \#69042} \citep{vllm69042}} & A prompt-logprob check, of the kind the RFC proposes for numerics PRs, catches the layer-0 FP8 failure but not the cached-KV bug above, which shows only when a cached prefix shares a step with a longer new prompt; a targeted request of that shape catches it & \anonelse{We posted H200 measurements in the RFC; the PR, a logprob drift eval for that RFC, cites them}{We posted H200 measurements in \#61698; \#69042, a logprob drift eval for that RFC, cites them}, leaves cache-reading paths out of its scope and proposes such a request as a follow-up test, which we wrote (the draft PR in the row above) \\
\bottomrule
\end{tabularx}
\end{table}
\fi

We also found that vLLM 0.33 cannot enable UltraQuant on NVIDIA GPUs together with its default boundary layers, because UltraQuant is missing from the CUDA backend priority list; our plugin adds it.\anonelse{}{ We noted this in the \#46461 and \#67789 threads but have not filed a separate issue.}

\FloatBarrier
\section{Further comparisons and breakdowns}
\label{sec:limdetail}

This appendix collects the comparisons and breakdowns behind the kernel results (Section~\ref{sec:kernel}), the end-to-end results (Section~\ref{sec:e2e}) and the limitations (Section~\ref{sec:limits}). \textbf{Measurement.} Every speed comparison has both configurations on the same GPU type (H200 SXM unless marked). The kernel tables and the pure-decode table each come from one job; the end-to-end latency pairs come from two consecutive jobs on one node, and the decode-led throughput from separate jobs. Reasoning accuracy needs repeated, shuffled runs (Section~\ref{sec:accuracy}).

\subsection{Other 4-bit decode paths}
\label{sec:other4bit}

\begin{itemize}
  \item \textbf{Against FOSCAR} (its released code, same process and K/V, its cold-L2 timer): on H200 our INT4 is 1.08--1.61$\times$ faster at equal bytes (68 per row, 136 per token and head), and our NVFP4 kernel is 0.99--1.40$\times$ FOSCAR INT4's speed. With the recipe, our NVFP4 path is slower than FOSCAR at $1 \times 16$K (1.22$\times$ vs 1.46$\times$ FlashInfer's FA2 FP8 kernel), $4 \times 32$K and $32 \times 4$K (0.84--0.97$\times$ FOSCAR's speed) and faster at the other 12 shapes; this H200 run used an earlier, unfused q preparation. FOSCAR's H200 times match its published H100 numbers, so it leaves H200's extra bandwidth unused. On H100, the GPU FOSCAR was tuned for, our INT4 is 1.02--1.31$\times$ faster and our NVFP4 0.94--1.21$\times$, so the margin there is small.
  \item \textbf{Against BitDecoding.} BitDecoding's Hopper INT4 kernel (\texttt{hopper} branch, 584d0d3, the latest commit) is the closest earlier design: a TMA producer warp, consumer warpgroups that unpack in registers, and wgmma, with per-token asymmetric INT4 (68 bytes per row, 136 per token and KV head). We timed it at the 11 shapes in the same job as our paged kernels, XQA and FlashAttention-4's FP8 (H200 SXM, CUDA-graph replays, the best of its own split heuristic and 2--64 splits; Table~\ref{tab:bitdecoding}). It reaches 22--26\% of peak bandwidth at batch 4 and above (14--24\% at batch 1); our paged NVFP4 kernel is 2.10--2.62$\times$ faster at batch 4 and above and 1.26--2.04$\times$ at batch 1, and BitDecoding is 1.12--1.83$\times$ slower than the faster FP8 kernel at every shape. In our correctness check (batch 2, 4K tokens, four query heads per KV head) its output differs from attention on the original K and V by $1.6 \times 10^{-2}$ on average, against $2.9 \times 10^{-3}$ for our INT4 kernel on the same data; with K set to zero the error vanishes, so its K path, not its V path or head mapping, computes different scores. That does not change the bytes it reads or its instruction mix, so we take the times as representative. From its source, it writes the unpacked values back to shared memory and runs wgmma from there, after the unpacking on the same warps, which fits its place next to FlashInfer's NVFP4 kernel at about 25\% of bandwidth (Table~\ref{tab:nvbit}).
  \item \textbf{Other 4-bit paths on Hopper.} Other 4-bit KV decodes on Hopper also beat an 8-bit baseline (FOSCAR's paper compares with TensorRT-LLM's FP8 decoding on H100, the library XQA comes from; the others were not compared with XQA): FOSCAR's INT4 (SGLang, \citealp{zheng2023sglang}) is 1.39--2.26$\times$ FlashInfer's FA2 FP8 kernel at the kernel level on H200, FServe's INT4 decode is competitive with, or above, FlashAttention-3's FP8 decode on H100 in its measurements~\citep{baek2026fserve}, and LMDeploy's TurboMind INT4 KV is 1.04--1.32$\times$ faster than its own INT8 KV and 1.06--1.70$\times$ faster than its BF16 KV per decode step (Llama-3.1-8B, H200, $B = 1$--32, 16K--32K; BF16 does not fit at $32 \times 32$K). Open SGLang PRs decode a packed E2M1 cache inside FlashMLA's SM90 sparse decode kernel for DeepSeek-V4.1's multi-head latent attention~\citep{sglang43852}; the measurements quoted there (H20, before a rebase) are level with the existing path at batch 32 and 0.3--4.7\% slower at batch 2. FlagRot's INT4 cache~\citep{shao2026flagrot}, inside vLLM 0.30 on H200, serves agent traffic at 0.92--0.94$\times$ the throughput of FlashInfer FP8 with BF16 sink and recent tokens and at 1.00--1.09$\times$ without them (our arithmetic from its table). KV4-RL~\citep{park2026kv4rl} uses INT4 KV with sinks and recent tokens in INT8 in its own engine for RL rollouts on H200 and reports decode forward passes 1.44--1.97$\times$ faster than FlashInfer BF16 and a 1.80$\times$ rollout speed-up over BF16 across 400 GRPO iterations. TurboMind has no FP8 KV, and these whole-step numbers come from a different engine, so they are not comparable with our vLLM numbers. Every other 4-bit path we measured for vLLM was slower than FlashInfer FP8: vLLM 0.33's stock UltraQuant backend (1.8--4.4$\times$ slower per step, H200 SXM, one job), and FlashInfer NVFP4 and the open PR \#53060 at the kernel level (H200 SXM; \#53060 at 0.70--0.90$\times$ FlashInfer's FA2 FP8 kernel at all 11 shapes, CUDA graphs).
\end{itemize}

\begin{table}[tbp]
\centering
\caption{BitDecoding's Hopper INT4 kernel against our paged NVFP4 kernel and the faster FP8 kernel (XQA or FlashAttention-4) at each shape: ms per decode call, CUDA-graph replays, one job on one H200 SXM (Llama-3.1-8B head shape). BitDecoding: best of its split heuristic and 2--64 splits, and its heuristic alone. Ratios above 1: BitDecoding slower.}
\label{tab:bitdecoding}
\small
\begin{tabular}{@{}lrrrrrr@{}}
\toprule
$B \times$ context & BitDecoding & its heuristic & Ours (NVFP4) & Best FP8 & vs ours & vs best FP8 \\
\midrule
$1 \times 16$K   & 0.0270 & 0.0697 & 0.0215 & 0.0241 & 1.26 & 1.12 \\
$1 \times 64$K   & 0.0713 & 0.2412 & 0.0409 & 0.0528 & 1.74 & 1.35 \\
$1 \times 128$K  & 0.1262 & 0.4716 & 0.0619 & 0.0839 & 2.04 & 1.50 \\
$4 \times 32$K   & 0.1263 & 0.2381 & 0.0601 & 0.0823 & 2.10 & 1.53 \\
$8 \times 64$K   & 0.4606 & 0.4634 & 0.1859 & 0.2591 & 2.48 & 1.78 \\
$16 \times 32$K  & 0.4603 & 0.4610 & 0.1864 & 0.2591 & 2.47 & 1.78 \\
$16 \times 128$K & 1.7981 & 1.8055 & 0.6869 & 0.9923 & 2.62 & 1.81 \\
$32 \times 4$K   & 0.1327 & 0.1338 & 0.0603 & 0.0777 & 2.20 & 1.71 \\
$32 \times 16$K  & 0.4647 & 0.4649 & 0.1870 & 0.2553 & 2.49 & 1.82 \\
$64 \times 16$K  & 0.9156 & 0.9186 & 0.3514 & 0.4990 & 2.61 & 1.83 \\
$128 \times 16$K & 1.8161 & 1.8186 & 0.6945 & 0.9968 & 2.61 & 1.82 \\
\bottomrule
\end{tabular}
\end{table}

\subsection{Ablation details}
\label{sec:ablationdetail}
\begin{table}[tbp]
\centering
\caption{Removal ablation (H200 SXM, one job, Llama-3.1-8B head shape, split count fixed at the full NVFP4 kernel's best): time relative to the full kernel (lower is faster) and fraction of peak bandwidth. Each variant drops one part of the work and reads the same bytes, except MXFP4 and INT4 (68 bytes per row, 6\% fewer); ``copy'' is a device read and write of the same bytes, so it is not a strict ceiling for a read-only kernel.}
\label{tab:ablation}
\small
\begin{tabular}{@{}lrrrr@{}}
\toprule
Variant & $1 \times 64$K & $8 \times 64$K & $32 \times 16$K & $128 \times 16$K \\
\midrule
Full NVFP4 kernel (ms)            & 0.042 & 0.188 & 0.188 & 0.675 \\
No E2M1 reconstruction            & 0.98 & 0.87 & 0.87 & 0.87 \\
No matrix multiplies              & 0.97 & 0.92 & 0.92 & 0.94 \\
Interleaved decoding (ILP)        & 1.08 & 1.10 & 1.11 & 1.18 \\
MXFP4                             & 0.98 & 1.01 & 0.99 & 1.01 \\
INT4                              & 0.93 & 0.87 & 0.87 & 0.85 \\
\midrule
Full kernel, fraction of peak     & 38\% & 67\% & 67\% & 75\% \\
No reconstruction, fraction of peak & 39\% & 77\% & 77\% & 85\% \\
Copy of the same bytes            & 76\% & 87\% & 87\% & 89\% \\
\bottomrule
\end{tabular}
\end{table}

\begin{itemize}
  \item \textbf{Where our kernel's time goes (removal ablation).} In one job on H200 SXM, with the split count fixed at the full kernel's best, we time variants that each drop one part of the work while reading the same bytes (Table~\ref{tab:ablation}). Without E2M1 reconstruction the kernel takes 0.87$\times$ the time at $B \geq 8$ (77--85\% of peak), and without the two matrix multiplies 0.92--0.94$\times$; a device copy of the same bytes reaches 87--89\% (counting read and write bytes; a large read-only kernel reaches 93--95\%). Decoding is therefore a real but minority cost for our kernel, about 13\% of its time at $B \geq 8$; the rest of the gap to the read-only ceiling is in the pipeline, the matrix multiplies and the softmax. A variant that interleaves the decoding of two rows for more instruction-level parallelism is 8--18\% slower, so we keep the simpler one. The instruction-budget diagnosis of Section~\ref{sec:stall} concerns the Triton and FlashInfer kernels, which execute two to six times the budget; ours executes about 92\% of the budget's instruction count per byte in FP4 (6.4 of about 7, at an implied issue utilization of about 62\%; next item), and removing its decoding saves only 13\%, so for our kernel the remaining limit is not decoding instructions alone.
  \item \textbf{Cumulative ablation (Table~\ref{tab:cumul}), details.} The write-time layout pays even while the decoded operands still pass through shared memory (1.45$\times$ and 1.31$\times$); we have not diagnosed why (candidates: shared-memory bank conflicts of the natural row reads and wgmma's transposed-A reads). In the bit-level step, the largest at $32 \times 16$K (Section~\ref{sec:kernel}), the generic table decode costs about one more instruction per value in NVBit (the decode routine itself accounts for 0.5, 15 against 11 instructions per 8 values; we did not isolate the rest); the shipped kernel's implied issue utilization is $6.44 \times 0.67 \div 7 \approx 62\%$. With staged operands, the bit-level decode alone buys 1.23$\times$. The generic decode is one choice (a 16-entry PRMT table), and a cheaper one would shrink the last step. The order matters too: the natural layout with the register path was not built.
  \item \textbf{FP4 on H200: the instruction stream, not bytes.} Besides the MXFP4 comparison of Section~\ref{sec:kernel}: an earlier build executed 6.7 instructions per byte, 96\% of the budget's instruction count (at 68\% of peak bandwidth, an implied issue utilization of 65\%), but later removing spin and address instructions gained at most 2\%, so the instruction total alone does not explain the limit; per-warp dependency chains are a candidate we could not measure without stall counters, and no single pipe is near its limit (Appendix~\ref{sec:kerneldetail}). On H100, with less bandwidth per instruction, both formats run near the memory limit.
\end{itemize}

\subsection{Prefill gap and per-kernel breakdowns}
\label{sec:prefill}

\begin{itemize}
  \item \textbf{Why the per-step gain is smaller than at kernel level.} Reading the weights (Section~\ref{sec:e2e}) takes at least 3.3 ms per step at H200's peak bandwidth, against 6.69 ms for our whole step at $8 \times 16$K (Table~\ref{tab:e2e}). In a per-kernel profile at $8 \times 16$K the attention-related kernels are about a quarter of our step (next item).
  \item \textbf{Decode step at $8 \times 16$K.} A per-kernel profile at $8 \times 16$K (both configurations in one job, H200 SXM, 32 decode steps) shows the split: our decode kernel takes 1.50 ms per step against FlashInfer's 2.88 ms (1.92$\times$), and with the sink merge, the cache write, the q rotation and layer 0 in BF16, the attention-related kernels total 2.00 against 3.02 ms, about a quarter of our step (7.55 vs 8.43 ms under the profiler); the GEMMs take about 4.0 ms in both.
  \item \textbf{Prefill.} vLLM's 4-bit backends run prefill attention on FlashAttention-2; we switch to FlashAttention-3 in BF16 (also proposed in vLLM PR \#47531). Prefill alone is still 7\% slower than FlashInfer FP8 (8 prompts of 32,000 tokens: 9.29 vs 8.66 s, same job), and time to first token is 1.5--6\% slower end to end. In a per-kernel profile of the current code (both configurations in one job, H200 SXM), a prefill step takes 3.6\%, 1.7\% and 4.5\% longer in wall time at $1 \times 16$K, $8 \times 4$K and 8K new tokens on 8K cached ones; the profile attributes the gap to prefill attention (6--7\% slower per layer than FlashInfer's FP8 prefill attention), copies around it (the output copy, contiguous K and V, the sink capture) and dequantization for continued prefill; our cache-write kernel takes the same time as FP8's. In the same profile the model's matrix multiplies, identical in both configurations, run 0.1--5.2\% slower with ours while the GPU sits at its power cap; we infer a shared power budget but have not isolated the effect. An FP8 prefill path does not close it: FlashAttention-3's FP8 prefill kernel is itself 3--12\% slower than its BF16 one in our runs, and quantizing q, K and V in front of it costs more than it saves, so our FP8 prefill option is 5--7\% slower than BF16 overall (and, on an earlier all-layer MXFP4 configuration, raised perplexity from +0.68\% to +1.51\%). FlashInfer's FP8 prefill kernel, not FP8 itself, is what is faster here. With prefill and decode on separate GPUs, the decode pool does not run prefill.
  \item \textbf{Small batch.} At $B = 1$ the kernel and its split merge reach only about 38\% of bandwidth at 64K context (18\% at 16K, for an isolated call of the kernel and its merge; in a kernel-level replay of the plugin's call sequence over 32 separate page pools, with the serial sink merge, the decode kernel alone takes 11.7~\textmu s per layer at 16 splits, about 34\% of peak and close to the 12.3~\textmu s per layer (382~\textmu s over 31 layers) of the serial-merge vLLM profile in the next item, and the sink merge 7.5~\textmu s; the whole call sequence, q rotation included, takes 22.8~\textmu s per layer, against 29.0~\textmu s for the same calls timed in isolation), and the whole-step result is on par with FP8: two measurements at 16K, each with both configurations on one node (end-to-end time per output token on one node, a per-step profile on another), gave 0.96$\times$ and 1.05$\times$, and FlashInfer's own time varied by 6\% across them (5.08--5.40 ms).
  \item \textbf{Small batch, per-kernel profiles.} With the old serial sink merge and 16 splits at $B = 1$, a per-kernel profile at $1 \times 16$K gave our decode kernel 382~\textmu s per token against FlashInfer's 702~\textmu s, but the sink merge (269~\textmu s, 8.7~\textmu s per layer), the cache write and the q rotation brought the 31 4-bit layers to 848~\textmu s (plus about 30~\textmu s for layer 0 in BF16) against 847~\textmu s. With the current defaults (parallel sink merge, 33 splits at $B = 1$; both configurations profiled in one job, H200 SXM), the decode kernel takes 12.4~\textmu s per layer, the sink merge 4.0, the q rotation 1.1 and the cache write 5.1, so the 31 4-bit layers take about 700~\textmu s per token against 836~\textmu s for FP8 (XQA and its cache write over 32 layers), while the step stays on par because weights dominate it at batch 1.
  \item \textbf{Fusion upper bound.} To bound what fusing the sink merge and the q rotation into the decode kernel could gain, we ran the plugin with both steps skipped (outputs invalid, timing only; \texttt{vllm bench forward}, one job, H200 SXM): the step shortens by 0.7\% at $1 \times 16$K, 1.1\% at $1 \times 32$K, 5.3\% at $8 \times 16$K and 2.1\% at $16 \times 64$K (skipping the merge alone: 0.2, 0.4, 3.3 and 1.4\%). A real fusion would add part of this back (the reduction moves into the decode kernel's tail), so we did not fuse them. This bounds only the removal of these two steps, not every possible change to the attention path.
  \item \textbf{Write path.} After optimization the cache-write kernel is 3.1$\times$ faster than its previous version, and in prefill as fast as FP8's (98.6 vs 97.8 ms for 8 prompts of 32,000 tokens, same job); in decode it is still slower (161.8 vs 144.5~\textmu s per token at $1 \times 16$K). The remaining prefill overhead is listed in the prefill item above.
\end{itemize}

\subsection{FP8 baselines and FlashAttention-4 inside vLLM}
\label{sec:fa4vllm}

We ran our plugin unchanged on a vLLM development build that contains the two FlashAttention-4 pull requests~\citep{fa206,vllm63139}. In it, plain \texttt{-{}-kv-cache-dtype fp8} selects FlashAttention-4 for all 32 layers (logged); FlashInfer's XQA path is selected explicitly for comparison. In-vLLM perplexity matches vLLM 0.33 (wikitext-2, $16 \times 8192$ tokens: 5.4831 against 5.4841 for the recipe, 5.4858 for both with C1). This build runs Llama through the Transformers modeling backend and is 0.65--1.4\,ms per step slower than vLLM 0.33 outside attention for every configuration, so ratios here are slightly diluted and are not comparable with Table~\ref{tab:e2e} across builds; within the build they are. The other per-step gains we report are against vLLM 0.33's FP8 path.

\begin{itemize}
  \item \textbf{FP8 baselines.} On that build (Table~\ref{tab:fa4vllm} below), per decode step our recipe is 1.04--1.24$\times$ faster than FlashAttention-4's FP8 at batch 8 and above and level at batch 1 (0.99--1.00$\times$), against 1.12--1.59$\times$ over XQA at batch 8 and above in the same job. XQA runs at 88--90\% of H200's peak at $32 \times 16$K and $128 \times 16$K (Table~\ref{tab:xqa}), and there our kernel-level gain (1.37--1.47$\times$; 1.37--1.48$\times$ in Table~\ref{tab:fa4}) is below the byte ratio of 1.78$\times$; raising our own bandwidth fraction at large batch is the main kernel headroom. NVBit counts at $32 \times 16$K (Section~\ref{sec:c1}) show why XQA is close there. Borrowing XQA's FP8 tensor-core path for K is our C1 path; its kernel, in-vLLM and capacity results are in Section~\ref{sec:c1}.
  \item \textbf{Why FlashInfer FP8 is the baseline.} vLLM's default FP8 path (FlashAttention-3 with an FP8 query) was slower than FlashInfer at large batch in our runs ($32 \times 32$K on H200 NVL: 30.2 vs 21.6 ms per step, same job), so all in-vLLM ratios are against FlashInfer; against the default they are larger (decode-led throughput 1.40$\times$ vs 1.17$\times$). FlashInfer's FP8 decode inside vLLM is the XQA kernel: forcing it with \texttt{VLLM\_HAS\_FLASHINFER\_CUBIN=1} changes the step time by at most 0.3\% (same job as Table~\ref{tab:e2e}).
  \item \textbf{Shared prefixes.} When requests share most of their KV pages, XQA gains more from L2 reuse than our kernel: with 90\% of pages shared, XQA's time over our paged NVFP4 kernel's falls from 1.38--1.86$\times$ without sharing to 1.14--1.18$\times$ at batch 32 and above (kernel level, one job on H200 with its own shapes and page sizes). Once bytes come from L2 our kernel is instruction-bound; one pass over the shared pages for all requests that share them (as in cascade or Hydragen-style attention) is not implemented.
\end{itemize}

\begin{table}[tbp]
\centering
\caption{Pure decode time per step against FlashAttention-4's native FP8 (\texttt{vllm bench forward}, Llama-3.1-8B, H200 SXM, all configurations in one job on the development build; median of 100 CUDA-graph replays; the shapes of Table~\ref{tab:e2e}). Ours: the recipe; ours + C1: with K on FP8 tensor cores (Section~\ref{sec:c1}). Speed-ups are over FlashAttention-4.}
\label{tab:fa4vllm}
\small
\begin{tabular}{@{}lrrrrrr@{}}
\toprule
$B \times$ context & FA4 FP8 (ms) & XQA FP8 (ms) & Ours (ms) & Ours + C1 (ms) & Ours & Ours + C1 \\
\midrule
$1 \times 16$K   & 6.01  & 6.06  & 6.07  & 6.08  & 0.99$\times$ & 0.99$\times$ \\
$1 \times 32$K   & 6.27  & 6.47  & 6.30  & 6.27  & 1.00$\times$ & 1.00$\times$ \\
$8 \times 16$K   & 7.88  & 8.52  & 7.60  & 7.46  & 1.04$\times$ & 1.06$\times$ \\
$8 \times 32$K   & 9.79  & 11.19 & 8.97  & 8.70  & 1.09$\times$ & 1.13$\times$ \\
$32 \times 16$K  & 13.74 & 13.78 & 11.95 & 11.45 & 1.15$\times$ & 1.20$\times$ \\
$32 \times 32$K  & 21.43 & 21.44 & 17.43 & 16.04 & 1.23$\times$ & 1.34$\times$ \\
$64 \times 16$K  & 21.61 & 25.51 & 17.82 & 16.81 & 1.21$\times$ & 1.29$\times$ \\
$16 \times 64$K  & 21.37 & 27.48 & 17.29 & 16.21 & 1.24$\times$ & 1.32$\times$ \\
$128 \times 8$K  & 22.13 & 22.95 & 18.67 & 17.48 & 1.18$\times$ & 1.27$\times$ \\
\bottomrule
\end{tabular}
\end{table}

FlashAttention-4 takes 0.78--1.00 of XQA's step time, and our per-step gain shrinks most where it beats XQA most ($16 \times 64$K: 1.24$\times$ over FlashAttention-4 against 1.59$\times$ over XQA in the same job). Serving capacity does not shrink in 600-s runs (the earlier design; not rerun in the steady-state design of Table~\ref{tab:capacity}). Under the reasoning-like stream (Llama-3.1-8B, four API servers, seed 0, both configurations in one job on this build except the two FP8 points noted below), FP8 with FlashAttention-4 sustains $[1.25, 1.275)$ requests/s, inside the $[1.25, 1.3)$ bracket of FP8 with XQA in vLLM 0.33 in 600-s runs: its cache is full at 1.25 (100\% peak, 132 preemptions, 99.2\% of requests within the SLO), and 85.8\% meet it at 1.275 and 89.9\% at 1.3 (the 1.2 and 1.275 points, FP8 only, ran in a second job on another H200 SXM node). Ours with C1 sustains $[1.65, 1.7)$ (100\% and 94.3\%; 90\% peak cache, no preemptions), 1.29--1.36$\times$ FP8's rate (highest held over lowest failed and the reverse, as in Table~\ref{tab:capacity}). In these runs FP8's cache is full at its limit, so there the cache, not attention speed, sets FP8's limit.

\FloatBarrier
\subsection{Per-step speed in vLLM: recipe costs, UltraQuant's format and other models}
\label{sec:otherspeed}

On Qwen3-32B, Qwen3-30B-A3B (MoE) and Qwen2.5-72B with FP8 weights the per-step gain at $B \geq 8$ is 1.04--1.32$\times$ (0.94--1.02$\times$ at $B = 1$), smaller than on Llama-3.1-8B because attention is a smaller share of the step (details below).

\begin{itemize}
  \item \textbf{Rerun of Table~\ref{tab:e2e} with the current defaults.} A rerun with the current defaults (parallel sink merge; FP8 and the recipe in one job, H200 SXM) gives the same speed-ups within 0.013 (1.01--1.05$\times$ at batch 1, 1.12--1.64$\times$ at batch 8 and above), with every time, FP8's included, within $-$0.6 to +2.9\% of the table's.
  \item \textbf{Llama-3.1-8B with FP8 weights.} With vLLM's online FP8 weight quantization (dummy weights; FP8 KV and the recipe in one job, H200 SXM, the shapes of Table~\ref{tab:e2e}), the recipe is 1.15--1.70$\times$ faster per step than FlashInfer FP8 at batch 8 and above.
  \item \textbf{Cost of mean removal and sinks.} Mean removal and sinks cost 1.2--3.2\% against NVFP4 with only layer 0 in BF16 (both in one earlier job on an H200 SXM node; the job of Table~\ref{tab:e2e} does not include that configuration).
  \item \textbf{UltraQuant's own format.} On UltraQuant's own format (MXFP4 with K rotation, the same bytes per token, vLLM's default four BF16 boundary layers; one job, H200 SXM, the shapes of Table~\ref{tab:e2e}), our kernel decodes 1.9--5.1$\times$ faster per step than vLLM's stock UltraQuant backend, which falls back to Triton on NVIDIA, and is 1.03--1.06$\times$ FlashInfer FP8 at batch 1 and 1.08--1.48$\times$ at batch 8 and above; the stock backend is 1.8--4.4$\times$ slower than FP8.
  \item \textbf{Zero-padded query groups.} In vLLM (two jobs, H200 SXM, FP8 with XQA decode) Qwen2.5-7B-Instruct ($G = 7$, zero-padded to 8) decodes at 0.97$\times$ FP8's speed at batch 1 and 1.06--1.28$\times$ at batch 8 and above, and Phi-4 ($G = 4$) at 0.99--1.00$\times$ and 1.01--1.28$\times$; at kernel level $G = 7$ takes the same time as $G = 8$ (within 0.7\%) and ties XQA at batch 1, so padding is not the cause of Qwen2.5-7B-Instruct's batch-1 gap, which we have not located.
  \item \textbf{Speed on other models} (pure decode, \texttt{vllm bench forward}, H200 SXM, vs FlashInfer FP8; Qwen3-32B and Qwen3-30B-A3B one job each, contexts up to 32K): Qwen3-32B (8 query heads per KV head, 64 layers) 1.00--1.02$\times$ at $B = 1$ and 1.05--1.18$\times$ at $B \geq 8$; Qwen3-30B-A3B (MoE, 4 KV heads) 0.94--0.95$\times$ at $B = 1$ and 1.11--1.32$\times$ at $B \geq 8$; Qwen2.5-72B-Instruct with FP8 weights 1.00--1.02$\times$ at $B = 1$ and 1.04--1.18$\times$ at $B \geq 8$ (contexts up to 31K, batch up to 32). The gain is smaller than on Llama-3.1-8B where attention is a small part of the step (short contexts, large weights).
\end{itemize}

\FloatBarrier
\section{Kernel details}
\label{sec:kerneldetail}

\paragraph{Drivers and resolved FP8 kernels.} vLLM's CUDA 13 wheel runs on the 570 driver through NVIDIA's CUDA 13 forward-compatibility package (\texttt{cuda-compat} 13.0.2), loaded only on drivers older than 580; our CUDA 12.8 kernels are compiled ahead of time for \texttt{sm\_90a} and run on the H100's 545 driver under CUDA 12 minor-version compatibility. This held for our kernels (the H100 numbers were measured on that node) but is not general: \texttt{sm\_90a} cubins of a FlashInfer build made with the same CUDA 12.8 were rejected there (\emph{device kernel image is invalid}). A few runs predate the logging of FlashInfer's resolved decode kernel: the end-to-end latency, throughput and prefill comparisons of Section~\ref{sec:e2e}, the $32 \times 32$K FlashAttention-3 comparison on H200 NVL and the per-step rows for Qwen3-32B and Qwen3-30B-A3B. They used the same scripts with the cluster's proxy in place, under which the logged runs resolve to XQA, and kernel profiles from that period show XQA's kernel.

\paragraph{Bandwidth ceiling and per-pipe rates.} A microbenchmark on H200 SXM (\anonelse{in the supplementary material}{\nolinkurl{bench/pipe_microbench.cu}}; SASS checked) reads 8 GB with 128-bit loads at 4.47--4.54 TB/s (93.1--94.5\% of 4.8 TB/s); a device-to-device copy moves 4.30 TB/s counting read and write bytes (89.5\%), so the copy figures elsewhere in the paper count both. Measured thread instructions per SM per clock: LOP3, PRMT, SHF and IMAD 63--64, the E4M3-to-FP16x2 unpack (F2FP) about 57--64, FFMA 126 and MUFU.EX2 16, with dependent-issue latencies of about 4 cycles (EX2 17). Applied to the shipped NVFP4 kernel's NVBit opcode counts at $32 \times 16$K (0.188 ms), the integer and logic instructions (LOP3, SHF, FMNMX, LEA and similar) keep their pipe 35\% busy, 41\% if F2FP runs on it; the IMAD instructions (mostly IMAD.SHL) 24\%, at most 41\% even if the HMUL2 scale multiplies share their pipe; MUFU 3\%; and total issue is 62\% of the scheduler limit. About 9\% of the instructions are spin-waits (branches, NANOSLEEP and barrier polls). No pipe is near its limit (Section~\ref{sec:kernel} discusses what this leaves); we could not measure stall reasons, as performance counters were not available to us.

\paragraph{How much of the ratio is the baseline.} FlashInfer's FA2 FP8 kernel reaches 57\% of H200's peak at $32 \times 16$K (0.396 ms for 1.07 GB, Table~\ref{tab:xqa}); its XQA kernel reaches 88\% there and 90\% at $128 \times 16$K, close to the 93--95\% that a large read-only kernel reaches. Against XQA our margin is therefore below the byte ratio of 128/72 = 1.78$\times$ wherever XQA runs near bandwidth (1.31--1.47$\times$ at $32 \times 4$K, $32 \times 16$K and $128 \times 16$K, where we reach 51--75\%), and near or above it where XQA does not (1.75--2.05$\times$ at batch 4--64 with long contexts or many sequences, where XQA reaches 45--73\%). At batch 1, XQA gains little over FA2 at 64K and 128K, and ours is 1.24--1.93$\times$ faster. The times reported for FlashAttention-4's native FP8 on H200~\citep{fa206} imply a higher bandwidth fraction than FA2's (our arithmetic from its table, a different job); we measure it at 85--92\% of peak at batch 8 and above, except $32 \times 4$K (Table~\ref{tab:fa4}).

\begin{table}[!htb]
\centering
\caption{Speed-up over FlashInfer's FA2 FP8 kernel (H200, Llama-3.1-8B head shape, one job and GPU, our times include the split merge; selected shapes). Against its XQA kernel see Table~\ref{tab:xqa}.}
\label{tab:kernel}
\small
\begin{tabular}{@{}lrrrrr@{}}
\toprule
$B \times S$ & FlashInfer FA2 FP8 (ms) & FlashInfer NVFP4 & Ours INT4 & Ours NVFP4 & Ours MXFP4 \\
\midrule
$1 \times 16$K & 0.033 & 0.83$\times$ & 1.54$\times$ & 1.43$\times$ & 1.46$\times$ \\
$1 \times 128$K & 0.116 & 0.62$\times$ & 1.95$\times$ & 1.81$\times$ & 1.83$\times$ \\
$8 \times 64$K & 0.400 & 0.59$\times$ & 2.48$\times$ & 2.14$\times$ & 2.14$\times$ \\
$32 \times 16$K & 0.395 & 0.59$\times$ & 2.44$\times$ & 2.11$\times$ & 2.11$\times$ \\
$16 \times 128$K & 1.557 & 0.59$\times$ & 2.72$\times$ & 2.31$\times$ & 2.26$\times$ \\
$128 \times 16$K & 1.542 & 0.59$\times$ & 2.70$\times$ & 2.29$\times$ & 2.27$\times$ \\
\bottomrule
\end{tabular}
\end{table}

\begin{table}[!t]
\centering
\caption{Against FlashInfer's XQA FP8 decode kernel (H200 SXM, one job, all 11 shapes; FlashInfer 0.7.3). XQA is called as vLLM calls it (HND page layout, i.e.\ heads, tokens, head dimension; BF16 query); the best of page sizes 16, 32, 64 and 128 is shown, with its bandwidth fraction on FP8 bytes. Ours: NVFP4, best split; the earlier fixed split rule (cap 16 at $B = 1$) is within 4\% of it except at $1 \times 128$K, where it gives 1.79$\times$ over XQA instead of 1.93$\times$. XQA's output differs from FA2's on the same FP8 cache by at most $4.8\times10^{-3}$.}
\label{tab:xqa}
\small
\begin{tabular}{@{}lrrrrr@{}}
\toprule
$B \times S$ & FA2 FP8 (ms) & XQA FP8 (ms) & XQA bandwidth & Ours NVFP4 (ms) & Ours vs XQA \\
\midrule
$1 \times 16$K   & 0.033 & 0.027 & 26\% & 0.022 & 1.24$\times$ \\
$1 \times 64$K   & 0.070 & 0.069 & 40\% & 0.042 & 1.66$\times$ \\
$1 \times 128$K  & 0.119 & 0.121 & 46\% & 0.063 & 1.93$\times$ \\
$4 \times 32$K   & 0.118 & 0.125 & 45\% & 0.061 & 2.05$\times$ \\
$8 \times 64$K   & 0.399 & 0.343 & 65\% & 0.187 & 1.83$\times$ \\
$16 \times 32$K  & 0.402 & 0.342 & 66\% & 0.187 & 1.83$\times$ \\
$16 \times 128$K & 1.552 & 1.314 & 68\% & 0.678 & 1.94$\times$ \\
$32 \times 4$K   & 0.118 & 0.080 & 70\% & 0.061 & 1.31$\times$ \\
$32 \times 16$K  & 0.396 & 0.255 & 88\% & 0.187 & 1.37$\times$ \\
$64 \times 16$K  & 0.794 & 0.615 & 73\% & 0.352 & 1.75$\times$ \\
$128 \times 16$K & 1.552 & 0.993 & 90\% & 0.675 & 1.47$\times$ \\
\bottomrule
\end{tabular}
\end{table}

\paragraph{Notes on Table~\ref{tab:fa4}.} The contiguous NVFP4 kernel, timed in the same job, is within 0.995--1.03$\times$ of the paged one except at $16 \times 128$K, where its graph time is an outlier (0.7213 ms against 0.6821 ms eager). FlashInfer's FA2 FP8 kernel, the third FP8 kernel we measured, is not in this job; in the job of Table~\ref{tab:xqa} it is at most 6\% faster than XQA, at two shapes where FlashAttention-4 takes 0.67--0.68 of XQA's time in Table~\ref{tab:fa4}, so it is not the fastest at any shape.

\paragraph{Reproducibility of the FlashAttention-4 times.} We measured FlashAttention-4 from the open pull requests (FlashAttention PR files at \texttt{8ef2c611} and vLLM PR files at \texttt{0d99666a} on a vLLM 0.34.0rc1 nightly build, 2026-11-22). Table~\ref{tab:fa4} times CUDA-graph replays, as vLLM decodes. With the eager timing of Tables~\ref{tab:xqa} and~\ref{tab:kernel}, FlashAttention-4 also pays 25--30~\textmu s of host overhead per call at batch 1--4 on this node (eager minus graph time, same job), so graph replays are the conservative choice for comparisons with it. The FP8 times match those of \anonelse{an earlier job on the same node}{the job behind this table in version 1 of this paper, on the same node} within 2.4\% at every shape but one: at $16 \times 128$K, FlashAttention-4 is 5.2\% faster in the job of Table~\ref{tab:fa4} (0.9757 against 1.0291 ms, with the same configuration, 16-token pages and its own split count; its eager time there is 1.0169 ms), at 92\% of peak it is just below the 93--95\% that a large read-only kernel reaches; we did not find why the two jobs differ there. A third job on the same node (the FlashInfer NVFP4 job of Figure~\ref{fig:kernel}) gives 1.0323 ms there, close to the earlier job; Table~\ref{tab:fa4} keeps the faster time, the conservative choice for our margin. It lowers our NVFP4 margin there to 1.42$\times$, against 1.49$\times$ with the earlier FlashAttention-4 time; at the other ten shapes the NVFP4 margin moved by at most 2\%.

\paragraph{With the recipe.} With K mean removal, rotation, four BF16 sinks and CUDA-graph replay, NVFP4 stays at 1.77--2.17$\times$ of FlashInfer's FA2 FP8 kernel (the slower of FlashInfer's two FP8 kernels) for $B \geq 32$; the sink merge and q rotation add 0--7\% at $B \geq 4$, except at $16 \times 128$K (+9--13\% under CUDA graphs in two jobs, about 1\% without), and 12--31\% at $B = 1$. Against FOSCAR with the recipe, see Appendix~\ref{sec:other4bit}.

\paragraph{Split count.} With its earlier cap of 16 at $B = 1$, the plugin's fixed split rule (Section~\ref{sec:setup}) picked the same split count as the per-shape search at 9 of the 11 shapes of the main kernel table, including the shapes with the lowest (1.43$\times$, $1 \times 16$K) and highest (2.31$\times$, $16 \times 128$K) speed-up; it differed only at $1 \times 64$K and $1 \times 128$K (16 instead of 32 splits). The current default caps $B = 1$ at 33 for the parallel sink merge, which is within one of the search's 32 at those two shapes; at $1 \times 16$K, where the search picks 16, it also gives 33.

\paragraph{How the instruction count came down.} The first K-side prototype reached 60\% of bandwidth at 3.0 instructions per value. NVBit then showed three compiler-level wastes (a LOP3 split into two instructions, wgmma descriptors recomputed per instruction, idle lanes doing softmax work); fixing them gave 1.95 instructions per value and 72\% (87\% in steady state). That prototype computed only $K q^\top$ (no softmax, no V); the complete kernel reaches 67--75\% (Table~\ref{tab:ablation}).

\paragraph{Register spills of the C1 kernel with eight query heads.} With eight query heads per KV head, the first version of the C1 kernel (Section~\ref{sec:c1}) was up to 2.5\% slower than our NVFP4 kernel: the extra registers pushed the eight output accumulators over the 96-register limit (with 288 threads and two CTAs, i.e.\ thread blocks, per SM, two of the four sub-partitions hold five warps, and 16,384 registers $\div$ 160 threads, rounded down to the allocation granularity, is 96), and they spilled to local memory around every $PV$ step. Accumulating the $PV$ wgmmas directly into the rescaled output, as FlashAttention-3 does, and issuing the $qK^\top$ steps in two rounds removes the spills; it is then 1.05--1.07$\times$ faster at batch 1--4 and 1.07--1.135$\times$ at batch 8 and above. The same restructuring also removes the smaller spills of the four-head kernel and is the default for both; Table~\ref{tab:fa4} uses it.

\FloatBarrier
\subsection{C2 implementation details}
\label{sec:c2detail}

\paragraph{MATH-500 runs.} On R1-Distill-Llama-8B (MATH-500, all 500 problems, three shuffled runs per configuration, H200 NVL), C1 scores 66.2\% and C2 67.8\% (BF16 and the recipe score 67.7\% in Table~\ref{tab:invllm}; all within run-to-run noise), a difference of +1.60 points ($p = 0.244$, 95\% CI [$-$1.67, +4.87]; method of Section~\ref{sec:accuracy}). These runs used C2's earlier, serial sink merge (and C1's parallel one).

\paragraph{Cache write.} Decode appends one token at a time, but a group's scale depends on all 32 of its tokens. C2's cache-write kernel keeps a running maximum: when a new token raises a group's scale, the group's earlier codes are re-encoded at the new scale, a second rounding (groups written in one call, as in prefill, are rounded once). On synthetic V, writing in decode order makes V's RMS error 1.05--1.11$\times$ that of writing in one call. A BF16 tail of up to 31 tokens per sequence, like KIVI's full-precision residual~\citep{liu2024kivi}, would avoid the second rounding but needs a side buffer and its own merge in vLLM. Our first version re-encoded one position at a time and took 34.8~\textmu s per call in vLLM's CUDA graph, against 5.5~\textmu s for C1's cache write (H200 NVL, same job), because with real data some channel's maximum grows at almost every step; it made C2's decode step slower than C1's at small batch (C1 took 0.93--0.97 of C2's step time at batch 1 and 8). The final version re-encodes each 32-bit word of eight codes at once, with a byte-permute table lookup (\texttt{\_\_byte\_perm}) on the 3-bit magnitudes, and writes the same bits in 8.7~\textmu s per call (same node).

\paragraph{Sinks and correctness.} C1 joins the BF16 sinks in the split merge by subtracting the 4-bit copies of the sink tokens; with an E4M3 P that is not exact, because the kernel rounds P against a running maximum the merge does not know. C2 instead masks the sink tokens inside the decode kernel and adds the BF16 sinks as one more partial in its own merge, which now spreads the splits over warps as C1's merge does and is the default\anonelse{}{ (\texttt{KV4\_C2\_MERGE=v2})}; it reorders a floating-point sum (FP32 difference from the serial version at most $7.8\times10^{-7}$) and passes our prefix-caching and chunked-prefill checks in vLLM. Against an FP32 emulation that uses the kernel's own operands, C2's output is within 0.53--0.92 of a BF16 unit in the last place on random data without a dominant K channel, including the sinks and the path through the cache write (max relative error $2.0$--$4.7\times10^{-3}$ with a strong K channel).

\paragraph{Instructions and bandwidth.} In the NVBit counts of Section~\ref{sec:c12results} ($32 \times 16$K, C2 against paged C1), the half-precision multiplies of the V scales disappear (0.25 per value to 0), integer multiply-adds fall from 0.54 to 0.26 and bit operations from 0.58 to 0.44, while the FP32 scaling after each matrix multiply adds 0.10 fused multiply-adds per value. The bandwidth fractions of Section~\ref{sec:c12results} count the 144 bytes per token and head that all three paged kernels copy; C1 and C2 keep NVFP4's page size and leave 4 and 8 of those bytes unused, so on the bytes they use, C2 reaches 75--84\% and C1 74--83\%.

\paragraph{Step time inside vLLM.} Per decode step (Llama-3.1-8B, \texttt{vllm bench forward}, the shapes of Table~\ref{tab:e2e}; FlashInfer FP8 with XQA decode, C1 and C2 in one job on H200 SXM, with the final cache write), C1 takes 0.997--1.002 of C2's step time at batch 1, 1.001--1.009 at $8 \times 16$K, $8 \times 32$K and $32 \times 16$K, and 1.014--1.032 at $32 \times 32$K, $64 \times 16$K, $16 \times 64$K and $128 \times 8$K; against FP8 in the same job C2 is 1.00--1.05$\times$ at batch 1 and 1.15--1.82$\times$ at batch 8 and above. C2's 4--5\% kernel gain over C1 at batch 8 and above thus shrinks to at most 3.2\% per step, and only at large batch with long contexts, because attention is a minority of the step (Section~\ref{sec:e2e}). That job ran the serial sink merges for both C1 and C2. With the parallel merge, now the default for all our kernels, the picture is the same (a later job on H200 SXM, FP8 with XQA decode, C1 and C2 in one job): the parallel merge makes C2's step 1.1--1.4\% faster at batch 1 and at most 0.5\% faster at batch 8 and above than its serial merge; C1 then takes 0.996--1.000 of C2's step time at batch 1, 1.000--1.008 at $8 \times 16$K, $8 \times 32$K and $32 \times 16$K, and 1.014--1.029 at $32 \times 32$K, $64 \times 16$K, $16 \times 64$K and $128 \times 8$K, and C2 is 1.03--1.08$\times$ FP8 at batch 1 and 1.15--1.83$\times$ at batch 8 and above.

\section{Accuracy details}
\label{sec:accdetail}

\paragraph{Summary.} With simulated quantization after RoPE (wikitext-2), NVFP4 with K rotation and K mean removal reaches a KL of 0.006--0.012 against BF16 on all three models tested; mean removal matters most on Qwen3-8B-Base, whose KL drops from 0.210 (rotation only) to 0.0114 (Table~\ref{tab:offline}). Table~\ref{tab:baselines} sets the recipe beside each KV-cache option in vLLM, including vLLM's NVFP4, which collapses on Qwen2.5-7B-Instruct and decodes more slowly than FP8 (Table~\ref{tab:e2e}); the paragraph below explains each step of the ladder of Table~\ref{tab:ladder}.

\providecommand{\pend}{\textcolor{gray}{pend.}}
\begin{table}[tbp]
\centering
\caption{What each KV-cache option gives on H200 (Llama-3.1-8B for speed, perplexity and KL). Bytes per 128-value
row of K or V, scales included. Speed: \texttt{vllm bench forward} decode step, all rows in one job, as a speed-up
over vLLM's FlashInfer FP8 path (XQA decode; higher is faster). MATH-500 (Qwen2.5-7B-Instruct, R1-Distill-Llama-8B, Phi-4) inside vLLM, mean over runs; perplexity increase over BF16 inside vLLM
(wikitext-2, $16\times8192$; Qwen2.5-7B-Instruct and Llama-3.1-8B); KL offline (FP32, $8\times2048$). One scale per tensor has no decode kernel in vLLM or
FlashInfer, so it has only offline numbers. KL: no sinks except ours (four); ours with the window K mean. vLLM NVFP4 is \texttt{-{}-kv-cache-dtype nvfp4} with vLLM's defaults
(every layer, global scale 1.0); $\times$246: perplexity 1542 against 6.27 (FP8: 231). A $\times$ gives the perplexity as a multiple of BF16's, used where the increase exceeds 100\%; smaller increases are in percent. MATH-500 of vLLM NVFP4: three runs each; FP8: one run each. $^*$BF16 at the three shapes whose cache fits (8$\times$16K, 8$\times$32K, 32$\times$16K tokens; at 64$\times$16K and above it does not), from a second job that reran FP8 and ours (ours there 1.12--1.63$\times$).}
\label{tab:baselines}
\small\setlength{\tabcolsep}{2.9pt}
\begin{tabular}{@{}lrrrrrrrrr@{}}
\toprule
& & \multicolumn{2}{c}{Speed-up vs FP8} & \multicolumn{3}{c}{MATH-500 (\%)} & \multicolumn{2}{c}{$\Delta$ppl} & Llama \\
\cmidrule(lr){3-4}\cmidrule(lr){5-7}\cmidrule(lr){8-9}
KV cache & Bytes & $B\geq8$ & $B=1$ & Qwen2.5 & R1-Llama & Phi-4 & Qwen2.5 & Llama & KL \\
\midrule
BF16                                & 256 & 0.63--0.86$^*$ & 0.96--0.97 & 71.2 & 67.7 & 75.3 & -- & -- & -- \\
FP8 (FlashInfer XQA)                & 128 & 1.00  & 1.00  & 1.6 & 67.0 & 76.0 & $\times$36.8 & +0.94\% & 0.0037 \\
4-bit, one scale/tensor & 64 & -- & -- & -- & -- & -- & -- & -- & 0.297 \\
vLLM NVFP4                          & 72  & 0.43--0.72 & 0.87--0.90 & 0.0 & 68.4 & 74.9 & $\times$246 & +2.62\% & 0.0282 \\
\quad layer 0 in BF16            & 72  & --    & --    & 65.9 & --    & --    & +78\% & +2.56\% & 0.0262 \\
Ours (recipe)                  & 72  & 1.12--1.64 & 1.01--1.05 & 72.0 & 67.7 & 74.1 & +0.05\% & +0.06\% & 0.0103 \\
\bottomrule
\end{tabular}
\end{table}

\paragraph{What each step fixes.}
Steps (a)--(c) run without sinks, (d)--(f) with four; FP8 is as vLLM ships it (every layer, no sinks).
\textbf{(a)$\to$(b) Block scales.} With one scale per tensor, the largest value in the layer sets the step size;
E2M1's smallest non-zero value is $0.5$ of that scale, so where K has a channel or token tens of times larger than
the typical value, most entries round to zero. A scale per 16 values confines an outlier to its own block: KL falls
11$\times$ on Llama and 103$\times$ on Qwen3-8B-Base. (INT4 with one scale per tensor is worse still: 1.74 and 4.26.)
Fixing NVFP4's global scale at 1.0, as vLLM stores it, changes the KL by at most 1.3\% on these two models.
\textbf{(b)$\to$(c) Rotation.} Within a block an outlier channel still sets the scale for the other 15 values; a
Hadamard rotation spreads each channel over all 128, and lowers Llama's KL by 30\%. On Qwen3-8B-Base it does the
opposite (0.029 to 0.281): K there has a large per-channel mean, and the rotation spreads it into every block,
where it takes over the scale (KV4-RL, \citealp[Appendix D.4]{park2026kv4rl}, also finds that K rotation raises NVFP4 KV mismatch on Qwen3 models without mean removal). Once the mean is removed (four sinks, layer 0 in BF16, one job per model), rotation
still lowers the KL, but only by 12\% on Llama, 14\% on Qwen2.5-7B-Instruct and 8\% on Qwen3-8B-Base.
\textbf{(c)$\to$(d) Sinks.} The first tokens receive a large share of the attention mass in most layers, so their
error weighs more than other tokens'. Before the mean is removed, sinks help little (Llama $-10\%$). With the mean
removed (recipe without the static mean, sink 0, 1, 2, 4, 8, 16), one sink already gives 98\% of the gain of four on
Llama, 101\% on Qwen3-8B-Base and 80\% on Qwen2.5-7B-Instruct (two give 97\%), and 8 or 16 add nothing we can
resolve: the gain is the first token, and four leave a margin.
\textbf{(d)$\to$(e) Mean removal.} The per-channel K mean is shared by all tokens and cancels in the softmax
($q\cdot\mu$ is the same for every key), so subtracting it before quantizing removes energy the scales would
otherwise spend on it, at no read-time cost. It is the step that rescues the Qwen models: Qwen2.5-7B-Instruct
goes from 7.05 (collapsed, as FP8 is at 3.86) to 0.0167, Qwen3-8B-Base from 0.249 to 0.0125, both close to Llama's
0.0111. It needs the sinks: at sink 0 it makes Llama worse, not better (0.0233 against 0.0187).
\textbf{(e)$\to$(f) Layer 0.} In Qwen2-family models the key-projection bias makes layer 0's K almost entirely its
mean (98\% of the norm on R1-Distill-Qwen-7B); keeping that layer in BF16, which takes 9--12\% of the cache bytes (one of 28--36 layers, at 3.6$\times$ the bytes per token), lowers the
KL by a further 12\% on Qwen2.5-7B-Instruct, 22\% on Qwen3-8B-Base and 7\% on Llama (one job each).
On its own it is not enough for 4-bit: on Qwen2.5-7B-Instruct, NVFP4 with four sinks and only layer 0 in BF16 has a
KL of 0.250, 17$\times$ the recipe's 0.0146, while the same protection takes FP8 from 3.79 to 0.033 (one job, four BF16 sinks in every configuration; Table~\ref{tab:ladder}, from other jobs, gives 0.0147 for the recipe and 3.86 for FP8 without sinks).
Inside vLLM the order is the same (Table~\ref{tab:baselines}): on Llama-3.1-8B, perplexity rises by 2.62\% with vLLM's NVFP4, 2.56\% with its layer 0 in BF16, 0.94\% with FP8 and 0.06\% with the recipe; on Qwen2.5-7B-Instruct, vLLM's NVFP4 with only layer 0 in BF16 still raises perplexity by 78\% and scores 65.9\% on MATH-500 against BF16's 71.2\% ($-$5.3 points, 95\% CI [$-$8.4, $-$2.4], $p < 0.001$, three runs against five), while the recipe scores 72.0\%.
\textbf{vLLM's NVFP4 as shipped} corresponds to (b) with the global scale fixed at 1.0 and no layer protection:
offline it is (b) within 1.3\% on Llama and Qwen3 in one job (the 0.0282 of Table~\ref{tab:baselines} comes from another job, which measured both vLLM NVFP4 rows; for the same configuration the two jobs give 0.0282 and 0.0271, 4\% apart) and collapsed on Qwen2.5-7B-Instruct (4.91).
Inside vLLM it is accurate on the two models without a key bias, R1-Distill-Llama-8B ($+0.7$ points against BF16, 95\% CI [$-$2.4, +3.9], $p = 0.60$) and Phi-4 ($-$0.33 between unrounded three-run means, [$-$2.6, +1.9], $p = 0.80$; three runs each), and collapses on Qwen2.5-7B-Instruct (0.0\% on MATH-500 in all three runs, perplexity 1542 against 6.27), as FP8 does (1.6\%). Away from the Qwen2 family its cost is speed (Table~\ref{tab:e2e}).
\textbf{Context length.} On Llama with four sinks, the recipe's KL (window mean, layer 0 in BF16) moves from
0.0093 at 2K tokens to 0.0113 at 16K, and its ratio to FP8 from 3.0 to 2.7 (32K tokens per point, a different sample from Table~\ref{tab:ladder}'s $8 \times 2048$ tokens, hence 0.0093 against its 0.0103).

\paragraph{Offline perplexity and KL} (simulated quantization after RoPE, wikitext-2~\citep{merity2016pointer}, $16 \times 4096$ tokens, first 4 tokens in BF16; Table~\ref{tab:offline}).

\begin{table}[tbp]
\centering
\caption{Offline KL against BF16 and perplexity increase (in parentheses); simulated quantization after RoPE, wikitext-2, $16 \times 4096$ tokens, first 4 tokens in BF16. The K mean here is taken over each window; a mean from the first 512 tokens is 8--14\% worse, and inside vLLM the mean is a static calibration.}
\label{tab:offline}
\begin{tabular}{@{}llll@{}}
\toprule
Model & NVFP4, K rotated & NVFP4, K mean-removed + rotated & FP8 \\
\midrule
Llama-3.1-8B & KL 0.0165 (+0.097) & KL 0.0115 (+0.065) & +0.015 \\
Mistral-7B-v0.3 & KL 0.0071 (+0.030) & KL 0.0060 (+0.025) & +0.007 \\
Qwen3-8B-Base & KL 0.210 (+1.38) & KL 0.0114 (+0.041) & $\leq$ +0.015 \\
\bottomrule
\end{tabular}
\end{table}

In this setting ($16 \times 4096$ tokens, unlike Table~\ref{tab:ladder}), rotation lowers each 4-bit format's error 1.6--2.9$\times$ on Llama and Mistral with sinks (up to 3.3$\times$ without); the order is NVFP4, then INT4, then MXFP4. On Qwen3, rotation alone raises NVFP4's KL to 0.210, 18 times that of the mean-removed version: the per-channel mean of K, shared by all tokens, holds about half of K's energy and in some Qwen3 layers sits in one channel, and rotation spreads it into every 16-value group (FlagRot reports the same harm for whole-head rotation on Qwen3-8B). Removing the mean first (SageAttention's smooth K, also part of BitDecoding-FP4's recipe) fixes it at no read-time cost, since $q \cdot \mu$ is the same for every token. Mean removal needs the sink: without it, Llama is 25\% worse. These KL values were computed in BF16; recomputing Llama-3.1-8B's in FP32 lowers FP8's KL by 17--18\% and the 4-bit formats' by 2--6\%, and leaves the order of formats unchanged.

\paragraph{Perplexity inside vLLM.} Wikitext-2, $16 \times 8192$ tokens (OLMo-2 $16 \times 4000$), increase over BF16, ours vs FP8: Llama-3.1-8B +0.06\% vs +0.94\%; Qwen3-32B +0.07\% vs +0.59\%; Phi-4 +0.15\% vs +0.30\%; OLMo-2-1124-7B +0.06\% vs +0.34\%; Qwen3-30B-A3B +0.07\% vs +0.04\%; Qwen2.5-72B-Instruct with FP8 weights +0.03\% vs +0.95\%. The static K mean is calibrated on the first $8 \times 2048$ tokens of the wikitext-2 test split, which the first two of the 16 windows contain. Calibrating on the train split instead gives +0.060\% on Llama-3.1-8B and +0.066\% on Qwen2.5-7B-Instruct, against +0.059\% and +0.047\% with the test split, and leaving out the two overlapping windows gives +0.069\% and +0.051\%, so the overlap does not make these rows look better.

\paragraph{Long context, offline.} On LongBench~\citep{bai2024longbench}, with Llama-3-8B-ProLong-512k-Instruct (12 tasks, 600 questions), no format differs from BF16 by more than 0.6 points; the recipe is +0.01. On RULER-style tasks at 32K--64K (600 paired trials), NVFP4 with rotation alone loses 2.4 points ($p = 0.001$); the recipe loses 0.5 ($p = 0.55$). At 120K (multi-key, 100 trials, Llama-3.1-8B) every 4-bit format loses 6--12 points, the recipe about 10, and MXFP4 45--50. Inside vLLM on the same 100 trials (H200 NVL, one run each), BF16 scores 44\% (42\% offline), FP8 on vLLM's default FlashAttention-3 path (FP8 query) 31\% and the recipe 29\%: the recipe against BF16 16 trials worse and 1 better (sign test $p = 0.0003$), FP8 against BF16 14 and 1 ($p = 0.001$), the recipe against FP8 9 and 7 ($p = 0.80$). Offline, FP8 storage (one scale per layer, BF16 query) scores 39\% against BF16's 42\% (5 trials worse, 2 better, $p = 0.45$) and the recipe 32\% (11 and 1, $p = 0.006$). On ProLong, trained for long context, the recipe is within 1.5 points of BF16 at 120K inside vLLM (50 trials per task) and within 3 points offline.

\FloatBarrier
\section{Practical notes for vLLM users on Hopper}
\label{sec:practical}

From our measurements (vLLM 0.33, Llama-3.1-8B unless noted, H200):
\begin{itemize}
  \item vLLM's NVFP4 KV cache (\texttt{-{}-kv-cache-dtype nvfp4}) saves memory but decodes 1.4--2.3$\times$ slower than FP8 at batch 8 and above and about 1.1$\times$ slower at batch 1 (Section~\ref{sec:e2e}). Its UltraQuant format falls back to Triton decode kernels on NVIDIA and is 1.8--4.4$\times$ slower than FP8 per step.
  \item With FlashInfer and FP8 KV, vLLM uses the faster XQA decode kernel only if its network check to NVIDIA's artifact server succeeds; when the check fails (on our cluster, behind a proxy), it falls back to FlashInfer's own decode kernel without an error. Check the log for \texttt{decode\_backend=xqa} (Section~\ref{sec:background}).
  \item Qwen2-family reasoning models can collapse under FP8 KV without any error. Keeping layer 0 in BF16 (\texttt{-{}-kv-cache-dtype-skip-layers 0}, 1/28 more KV memory on a 28-layer model) restored them in our runs (Section~\ref{sec:accuracy}), but skipping layers crashes FlashAttention-3 for some models (\#65821, \citealp{vllm65821}; fix proposed in \#65891, \citealp{vllm65891}).
  \item Our plugin helps most when decoding dominates and KV memory is the limit (batch 8 and above, long outputs, large models on one GPU; Section~\ref{sec:e2e}). At batch 1 it is on par with FP8, and prefill-heavy streams are slower (Section~\ref{sec:e2e}). It supports head dimension 128 with up to 8 query heads per KV head (Section~\ref{sec:limits}).
\end{itemize}

\end{document}